\documentclass[11pt,eqsecnum,nofootinbib]{revtex4}

\usepackage{amsmath,amsfonts,bm}
\usepackage[hyperfootnotes=false, linktoc=section]{hyperref}
\usepackage{url}
\usepackage{epsfig}

\usepackage{epigraph}
\usepackage{titlesec}
\titleformat{\section}{\normalfont\huge\bfseries}{\thesection.}{1em}{}
\renewcommand{\thesection}{\Roman{section}} 
\titleformat{\subsection}{\normalfont\Large\bfseries}{\thesubsection.}{1em}{}
\renewcommand{\thesubsection}{\arabic{subsection}} 
\titleformat{\subsubsection}{\normalfont\normalsize\bfseries}{\thesubsubsection.}{1em}{}
\renewcommand{\thesubsubsection}{\Alph{subsubsection}} 
\usepackage{xcolor}         % colors
\definecolor{darkgreen}{rgb}{0,0.8,0}

\newcommand{\be}{\begin{equation}}
\newcommand{\ee}{\end{equation}}
\newcommand{\bi}{\begin{itemize}}
\newcommand{\ei}{\end{itemize}}
\renewcommand{\le}{\left}
\newcommand{\ri}{\right}

\usepackage{amssymb}
\newcommand{\sbreak}{
    \begin{center}
        $\blacklozenge$$\blacklozenge$$\blacklozenge$$\blacklozenge$
    \end{center}
}

\newcommand{\E}[1]{\mathbb{E}\left[#1\right]} %expectation at initialization
\newcommand{\loss}{\mathcal{L}}

\newcommand{\A}{\mathcal{A}}

\newcommand{\fanin}{\texttt{fan\_in}}

\newcommand{\lrt}{\texttt{lr}}%{\eta_{\text{og}}}
\newcommand{\wdt}{\texttt{wd}}%{\lambda_{\text{wd}}}
\newcommand{\ws}{\Omega}%{\omega}%{\varpi}
\newcommand{\wst}{\widetilde{\ws}}
\newcommand{\WE}{W^{\text{WE}}}
\newcommand{\patch}{W^{\text{patch}}}
\newcommand{\head}{W^{\text{head}}}
\newcommand{\PE}{b^{\text{PE}}}
\newcommand{\headb}{b^{\text{head}}}
\newcommand{\sa}{\alpha}%{\delta}% sample index
\newcommand{\nin}{n_{\text{in}}}%input dimension
\newcommand{\nnout}{n_{\text{out}}}%output dimension
\newcommand{\hid}{n}%hidden dimension (embedding dimension)
\newcommand{\npatch}{n_{\text{patch}}}%{16\times16\times3}
\newcommand{\nvocab}{n_{\text{vocab}}}
\newcommand{\rescale}{\mathcal{N}_{\text{rescale}}}
\newcommand{\xerox}{\xi}%{\chi}
\newcommand{\kernelLN}{F}%{\Sigma}
\newcommand{\NTK}{\widehat{H}}
\newcommand{\NTKM}{H}
\newcommand{\LambdaSuper}{\lambda}%{\Lambda}
\newcommand{\lambdaSub}{\Lambda}%{\lambda}
\newcommand{\LambdaG}{\LambdaSuper_{G}}
\newcommand{\LambdaGmu}{\LambdaSuper_{G(\mu)}}
\newcommand{\LambdaSuperA}{\widetilde{\LambdaSuper}}
\newcommand{\LambdaGA}{\LambdaSuperA_{G}}
\newcommand{\LambdaGAmu}{\LambdaSuperA_{G(\mu)}}
\newcommand{\lambdaSubA}{\widetilde{\lambdaSub}}
\newcommand{\blocki}{\ell}%block index
\newcommand{\LNopp}{\texttt{LN}}
\newcommand{\Resopp}{\texttt{R}}
\newcommand{\kernelA}{A}

\usepackage{dcolumn}

\begin{document}

%\title{Effective Theory of Transformers at Initialization and Beyond}
%\title{Effective Theory of Transformers at Leading Order}
\title{Effective Theory of Transformers at Initialization}

\author{Emily Dinan}
\email{edinan@meta.com}
\affiliation{Meta AI\\
Meta Platforms, Inc.}
\thanks{The author ordering was determined by the hypothetical coin toss that 100\%-respects the alphabetical ordering.}
\author{Sho Yaida}
\email{shoyaida@meta.com}
\affiliation{Meta AI\\
Meta Platforms, Inc.}
\thanks{The author ordering was determined by the hypothetical coin toss that 100\%-respects the alphabetical ordering.}
\author{Susan Zhang}
\email{susanz@meta.com}
\affiliation{Meta AI\\
Meta Platforms, Inc.}
\thanks{The author ordering was determined by the hypothetical coin toss that 100\%-respects the alphabetical ordering.}

\begin{abstract}
We perform an effective-theory analysis of forward--backward signal propagation in wide and deep Transformers, i.e., residual neural networks with multi-head self-attention blocks and multilayer perceptron blocks. This analysis suggests particular width scalings of initialization and training hyperparameters for these models. We then take up such suggestions, training Vision and Language Transformers in practical setups.
\end{abstract}
\maketitle
\newpage
\tableofcontents
\newpage

%\section{Foundations}\label{sec:foundations}
\section{Theoretical Foundations}\label{sec:foundations}

%\epigraph{\textit{Freedom is the right of all sentient beings.}}{Optimus Prime}
\setlength{\epigraphwidth}{0.31\textwidth}
\epigraph{\textit{Autobots, transform and roll out!}}{Optimus Prime}
%\setlength{\epigraphwidth}{0.58\textwidth}
%\epigraph{\textit{There's a thin line between being a hero and being a memory.}}{Optimus Prime}

We live in an exciting time in the history of artificial intelligence science and technology. On the one hand, our state-of-the-art models are getting larger and larger in the blink of an eye~\cite{brown2020language,rae2021scaling,smith2022using,chowdhery2022palm,zhang2022opt}, with the number of their model parameters in the order of trillions at the time of this writing. On the other hand, we have an effective theoretical description that becomes asymptotically more accurate and more relevant as models get wider~\cite{Neal1996,LBNSPS2017,MRHTG2018,jacot2018neural} and deeper~\cite{poole2016exponential,raghu2017expressive,schoenholz2016deep}. Thus, we now have no excuse but to bring the theory and practice closer together.
% -- or else the effective theory will just become a memory.

Transformers~\cite{vaswani2017attention,dosovitskiy2020image} provide an ideal case in point, not only because they are driving numerous technical innovations of this era but also because -- to achieve these innovations -- they are getting so wide and deep that the effective theory should in principle be getting more apt and relevant. With that optimistic conviction, in Part~\ref{sec:foundations} of this note, we'll roll out the general-purpose blueprint laid out in Ref.~\cite{PDLT} and fine-tune it to develop the effective theory of Transformers; in Part~\ref{sec:applications}, we'll then see the impacts of the resulting theoretical suggestions in practice.\footnote{Realistically speaking, Transformers won't forever be the pinnacle of the neural-network evolution but will be replaced by other model families. The broader goal of this note is to provide a \textit{meta-blueprint} to roll out the said effective-theory blueprint -- which was illustrated in Ref.~\cite{PDLT} for the simple case of multilayer perceptrons  -- for more complicated neural-network architectures that come our way. To that end, we'll be as verbose as possible in our presentation so that it should be relatively straightforward to mirror the way we roll out the blueprint.}

The rest of Part~\ref{sec:foundations} is organized as follows. We'll kick off our forward-path analysis in \S\ref{subsec:forward}  by briefly reviewing the building blocks of Transformers while setting up our notations.  Then in \S\ref{subsec:initialization} we'll calculate the statistics of preactivations at initialization, all the way from inputs to outputs: this analysis helps us determine how to scale initialization hyperparameters with width~\eqref{eq:init-preview-stem}--\eqref{eq:init-preview-head}.
Shifting our gear to the backward path, after a brief detour on neural tangent kernels in \S\ref{subsec:NTK}, we'll calculate the statistical means of  squared gradients in \S\ref{subsec:backward}:
this analysis helps us determine how to scale group-wise learning-rate factors with width, both for the stochastic gradient descent optimizer~\eqref{eq:SGDlambdaG-preview-stem}--\eqref{eq:SGDlambdaG-preview-head} and for the AdamW optimizer~\eqref{eq:AdamWlambdaG-preview-stem}--\eqref{eq:AdamWlambdaG-preview-head}.\footnote{For those in the know: our effective-theory analysis stays at a meta level. That is, there are two things we \textit{won't} do: \textit{(a)} the criticality analysis -- because the putative exploding or vanishing gradient problems are mostly taken care of by normalization layers and residual connections --  and \textit{(b)} the finite-width analysis  -- because such a treatment would quadruple the length of our note and such labor is not necessary in reading off the hyperparameter scalings that we'll test in Part~\ref{sec:applications}. That said, hopefully this note lays the foundations for those who dare to push on these extensions, which can be useful, e.g., in investigating how to judiciously scale up architecture hyperparameters.}

\newpage
\setcounter{subsection}{-1}
\subsection{A Crash Course on Transformers}\label{subsec:forward}
Given a dataset $\mathcal{D}$, we denote its tokenized inputs as $x_{\sa;t;i}$ with a sample index $\sa=1,\ldots,\vert\mathcal{D}\vert$, a token index $t=1,\ldots,T$, and a vector index $i=1,\ldots,\nin$ (see \S\ref{subsubsec:forward-stem} for concrete examples). Then a standard Transformer~\cite{vaswani2017attention,dosovitskiy2020image} with an embedding dimension $n$ recursively transforms them as
\begin{align}
z^{(1)}_{\sa;t;i}=&\Resopp^{(1)}_{t;i}\le(x_{\sa}; \theta^{(1)}\ri)\, \ \ \ \ \ \ \ \ \ \ \ \ \ \ \ \ \ \ \text{for}\ \ \ i=1,\ldots,\hid\, ,\label{eq:stem}\\
s^{(\blocki)}_{\sa;t;i}=&\LNopp_i\le(z_{\sa;t}^{(\blocki)}\ri)\, \ \ \ \ \ \ \ \ \ \ \ \ \ \ \ \ \ \ \ \ \ \ \ \text{for}\ \ i=1,\ldots,\hid;\ \ \blocki=1,\ldots,L-1\, ,\label{eq:LN}\\
z^{(\blocki+1)}_{\sa;t;i}=&\Resopp^{(\blocki+1)}_{t;i}\le(s^{(\blocki)}_{\sa}; \theta^{(\blocki+1)}\ri)+z^{(\blocki)}_{\sa;t;i}\, \ \ \text{for}\ \ i=1,\ldots,\hid;\ \ \blocki=1,\ldots,L-2\, ,\label{eq:resrec}\\
z^{(L)}_{\sa;t;i}=&\Resopp^{(L)}_{t;i}\le(s_{\sa}^{(L-1)}; \theta^{(L)}\ri)\, \ \ \ \ \ \ \ \ \ \ \ \ \text{for}\ \ \ i=1,\ldots,\nnout\, ,\label{eq:head}
\end{align}
where $\theta^{(\blocki)}$ are the model parameters for the $\blocki$-th block. Specifically, first, the stem-block operation $\Resopp^{(1)}$~\eqref{eq:stem} converts the inputs $x_{\sa;t;i}$ into the first-block preactivations $z^{(1)}_{\sa;t;i}$, which are $T$-sequenced $\hid$-dimensional vectors; then, these preactivations recursively go through a series of the layer-normalization operations~\eqref{eq:LN}~\cite{ba2016layer} (\S\ref{subsubsec:forward-LN}) and block operations~\eqref{eq:resrec} -- with each residual path $\Resopp^{(\blocki)}$ being either a multi-head self-attention block (\S\ref{subsubsec:forward-MHSA}) or a multilayer perceptron block (\S\ref{subsubsec:forward-MLP}) -- that are sandwiched by skip connections; and, last, the head-block operation $\Resopp^{(L)}$~\eqref{eq:head} converts them into $T$-sequenced $\nnout$-dimensional vectors $z^{(L)}_{\sa;t;i}$ (see \S\ref{subsubsec:forward-head} for concrete examples).\footnote{More generally, each skip connection can come with its own parameters as
\be
z^{(\blocki+1)}_{\sa;t;i}=\Resopp^{(\blocki+1)}_{t;i}\le(s^{(\blocki)}_{\sa}; \theta^{(\blocki+1)}\ri)+\xerox_i^{(\blocki+1)} z^{(\blocki)}_{\sa;t;i}\, ,
\ee
where element-wise affine parameters $\xerox_i^{(\blocki)}$ dictate how much of the signal we copy and paste in each channel. But,  in this note, we'll keep them fixed as $\xerox_i^{(\blocki)}=1$ for simplicity -- as is also standard in practice. Similarly, each layer-normalization operation can come with its own trainable element-wise affine parameters (\S\ref{subsubsec:forward-LN}), but we'll again keep them fixed in our theoretical and practical treatments -- which is less standard but not unheard of:
\url{https://pytorch.org/docs/stable/generated/torch.nn.LayerNorm.html}.\\
Incidentally, normalization layers were originally placed differently~\cite{vaswani2017attention}, but the above positioning is more standard today and, for residual neural networks, such placement ensures that block-to-block signal propagation is \textit{everywhere critical}~\cite{doshi2021critical}, that is,
it makes the success of training less sensitive to the choices of initialization hyperparameters.}

Now that we have an overall schematic, let us define each operation in detail, block by block.

\subsubsection{Stem Block}\label{subsubsec:forward-stem}
\begin{center}\textit{Vision: patchify embedding and positional embedding}\end{center}

For Vision Transformers~\cite{dosovitskiy2020image}, input images are typically tokenized into non-overlapping patches. To be very concrete, in typical ImageNet~\cite{deng2009imagenet} training setups, given a $3$-colored $224$-by-$224$-pixelated image, we can for instance partition it into $(T=14^2=196)$ tokens of non-overlapping patches, each of which can be seen as an $(\nin\equiv\npatch=16\cdot16\cdot3=768)$-dimensional vector.
To implement this, we can use patchify weights -- $16$-by-$16$ convolutional weights with stride $16$, intaking $3$ in-channels and outputting $n$ out-channels -- which act as
\be
\tilde{z}_{\sa;t;i}=\sum_{j=1}^{\npatch}\patch_{ij}x_{\sa;t;j}\, ,
\ee
with the embedding index $i=1,\ldots,\hid$.
Note that, due to convolutional weight tying, these weights act in the same way for all tokens $t=1,\ldots,T$.
Then they are often amended by positional-embedding parameters as
\be
z^{(1)}_{\sa;t;i}=\PE_{t;i}+\tilde{z}_{\sa;t;i}=\PE_{t;i}+\sum_{j=1}^{\npatch}\patch_{ij}x_{\sa;t;j}\, ,\label{eq:PE}
\ee
where the $T$-by-$\hid$-dimensional tensor $\PE_{t;i}$ acts like bias parameters and is folklored to be useful for distinguishing different patches.\footnote{There exist several variants of the stem block in Vision Transformers.\label{foot:class} As one variant, we can attach an $\hid$-dimensional class token $c_i$ at, say, $t=0$ such that we have $(T+1)$-by-$\hid$-dimensional first-block preactivations
\be
z^{(1)}_{\sa;t;i}=\begin{cases}
c_i+\PE_{0;i}\, \ \ \ \ \ \ \ \ \ \ \ \ \ \ \ \ \ \ \ \ \ \ \ \ \ \ \text{for}\ \ t=0\, ,\\
\PE_{t;i}+\sum_{j=1}^{\npatch}\patch_{ij}x_{\sa;t;j}\, \ \ \text{for}\ \ t=1,\ldots,T\, ,
\end{cases}\, 
\ee
for each input $\alpha$~\cite{dosovitskiy2020image} (see also Ref.~\cite{zhai2022scaling}). Another variant is to have a stack of several convolutional layers~\cite{dosovitskiy2020image,xiao2021early}. While we'll focus on the stem block as described in the main text, it is straightforward to account for these variants.}\\

\begin{center}\textit{Language: word embedding and positional embedding}\end{center}

For Language Transformers~\cite{vaswani2017attention}, inputs are typically given in the form of -- or at least can be interpreted as -- one-hot vectors $x_{\sa;t;j}=\delta_{j j_{\star}}$ where the vector index $j=1,\ldots,\nvocab$ runs over all possible tokens in the vocabulary and, for a given pair of sample--token indices $(\sa,t)$, the input vector takes the unit value at a specific index $j_{\star}=j_{\star}(\sa;t)$  and otherwise returns zero. Each word is then embedded into an $\hid$-dimensional space as
\be
\tilde{z}_{\sa;t;i}=\sum_{j=1}^{\nvocab}\WE_{ij}x_{\sa;t;j}=\WE_{i j_{\star}(\sa;t)}\, ,
\ee
where the word-embedding parameters $\WE_{ij}$ act like an $\hid$-by-$\nvocab$ weight matrix. As in the vision case~\eqref{eq:PE}, they are then amended by the positional-embedding parameters as $z^{(1)}_{\sa;t;i}=\tilde{z}_{\sa;t;i}+\PE_{t;i}$.

\subsubsection{Layer Normalization}\label{subsubsec:forward-LN}
In general, a layer-normalization operation~\cite{ba2016layer} is  defined as 
\be
s^{(\blocki)}_{\sa;t;i}=\gamma^{(\blocki)}_i\le[ \frac{z^{(\blocki)}_{\sa;t;i}-\le(\frac{1}{\hid}\sum_{j=1}^{\hid}z^{(\blocki)}_{\sa;t;j}\ri)}{\sqrt{\frac{1}{\hid}\sum_{j=1}^{\hid}\le(z^{(\blocki)}_{\sa;t;j}\ri)^2-\le(\frac{1}{\hid}\sum_{j=1}^{\hid}z^{(\blocki)}_{\sa;t;j}\ri)^2+\epsilon}}\ri]+\beta_i^{(\blocki)}\, ,
\ee
with the regularization parameter $\epsilon$ and element-wise affine parameters $\gamma^{(\blocki)}_i$ and $\beta_i^{(\blocki)}$.  For simplicity, in this note, we'll keep these element-wise affine parameters fixed at $\gamma^{(\blocki)}_i=1$ and $\beta_i^{(\blocki)}=0$, that is,
\be
s^{(\blocki)}_{\sa;t;i}=\LNopp_i\le(z^{(\blocki)}_{\sa;t}\ri)= \frac{z^{(\blocki)}_{\sa;t;i}-\le(\frac{1}{\hid}\sum_{j=1}^{\hid}z^{(\blocki)}_{\sa;t;j}\ri)}{\sqrt{\frac{1}{\hid}\sum_{j=1}^{\hid}\le(z^{(\blocki)}_{\sa;t;j}\ri)^2-\le(\frac{1}{\hid}\sum_{j=1}^{\hid}z^{(\blocki)}_{\sa;t;j}\ri)^2+\epsilon}}\, .\label{eq:LNbare}
\ee
 In particular, when $\epsilon=0$ each signal is normalized exactly to unity as $(1/\hid)\sum_{i=1}^{\hid}\le(s^{(\blocki)}_{\sa;t;i}\ri)^2=1$ and nearly to unity for sufficiently small $\epsilon$.

\subsubsection{Multi-Head Self-Attention Block}\label{subsubsec:forward-MHSA}
A residual path of a multi-head self-attention (MHSA) block takes in a $T$-sequenced $\hid$-dimensional signal $s_{t;i}$ and outputs another $(T,\hid)$-dimensional tensor $r_{t;i}$, acting nontrivially in the sequence direction $t$. Specifically -- suppressing the block index $\blocki$ -- the standard MHSA-residual-block operation with $H$ heads,
\be
r_{\sa;t;i}=\Resopp^{\text{MHSA}}_{t;i}(s_{\sa}; Q,K,V,U)\, ,
\ee
is defined as follows.
\begin{enumerate}
\item Define query, key, and value vectors as
\begin{align}
q_{\sa;t;c}^{h}\equiv&\sum_{i=1}^{\hid} Q_{ci}^{h}s_{\sa;t;i}\, ,\label{eq:q-vector}\\
k_{\sa;t;c}^{h}\equiv&\sum_{i=1}^{\hid} K_{ci}^{h}s_{\sa;t;i}\, ,\label{eq:k-vector}\\
v_{\sa;t;c}^{h}\equiv&\sum_{i=1}^{\hid} V_{ci}^{h}s_{\sa;t;i}\, ,\label{eq:v-vector}
\end{align}
for $t=1,\ldots,T$, $c=1,\ldots,C$, and $h=1,\ldots,H$, where the number of channels per head, $C\equiv \hid/H$, must be an integer.
\item Define the query--key dot product as
\be\label{eq:matrix-product-1}
\wst_{\sa;tt'}^{h}\equiv\frac{1}{\sqrt{C}}\sum_{c=1}^{C}q_{\sa;t;c}^{h}k_{\sa;t';c}^{h}=\frac{1}{\sqrt{C}}\sum_{c=1}^{C}\sum_{i_1,i_2=1}^{\hid}Q_{ci_1}^{h}K_{ci_2}^{h}s_{\sa;t;i_1}s_{\sa;t';i_2}\, .
\ee
(We'll justify the factor of $1/\sqrt{C}$ first in \S\ref{subsubsec:initialization-MHSA} and then again in \S\ref{subsubsec:backward-MHSA}.)
\item Define the self-attention matrix $\ws_{\sa;tt'}^{h}$ as softmax based on the query--key dot product. Specifically, for \textit{bidirectional} MHSA blocks used in \textit{encoders}, we typically set
\be\label{eq:bidirectional-MHSA}
\ws_{\sa;tt'}^{h}\equiv\frac{\exp\le(\wst_{\sa;tt'}^{h}\ri)}{\sum_{t^{\prime\prime}=1}^{T}\exp\le(\wst_{\sa;tt^{\prime\prime}}^{h}\ri)}\, ,
\ee
while, for \textit{masked} MHSA blocks used in \textit{decoders}, we typically set
\be\label{eq:masked-MHSA}
\ws_{\sa;tt'}^{h}\equiv
\begin{cases}
\frac{\exp\le(\wst_{\sa;tt'}^{h}\ri)}{\sum_{t^{\prime\prime}=1}^{t}\exp\le(\wst_{\sa;tt^{\prime\prime}}^{h}\ri)}\, \ \ \text{for}\ \ t'\leq t\, ,\\
0\, \ \ \ \ \ \ \ \ \ \ \ \ \ \ \ \ \ \ \ \ \ \ \text{for}\ \ t'>t\, ,
\end{cases}\, 
\ee
that causally masks the future tokens from queriers.\footnote{In general, the self-attention matrix can be any order-one matrix-valued function of the query--key dot product, $\ws_{\sa;tt'}^{h}=\mathcal{F}_{tt'}\le(\wst_{\sa}^{h}\ri)$, that maps a $T$-by-$T$  matrix to a $T$-by-$T$  matrix. Indeed, in this note, we won't use any particular property of the function $\mathcal{F}$ beyond that it is of order one.}
\item Weigh the value vector with the self-attention matrix in the token direction for each head, and then tie $H$ heads and $C$ channels with the unifying -- a.k.a.~\texttt{out\_proj} -- matrix $U_{ic}^{h}$ as
\be\label{eq:matrix-product-2}
r_{\sa;t;i}\equiv \sum_{h=1}^{H}\sum_{c=1}^{C}U_{ic}^{h} \le(\sum_{t'=1}^{T}\ws_{\sa;tt'}^{h}v_{\sa;t';c}^{h}\ri)=\sum_{h=1}^{H}\sum_{t'=1}^{T} \ws^{h}_{\sa;tt'}\sum_{c=1}^{C}\sum_{j=1}^{\hid}U_{ic}^{h}V_{cj}^{h}s_{\sa;t';j}\, .
\ee
\end{enumerate}

\subsubsection{Multilayer Perceptron Block}\label{subsubsec:forward-MLP}
A residual path of a multilayer perceptron (MLP) block in Transformers often takes the form
\begin{align}
w_{\sa;t;i}=&\sum_{j=1}^{\hid}W_{ij}s_{\sa;t;j}\, \ \ \ \ \ \ \ \ \text{for}\ \ \ i=1,\ldots,M\hid\, , \\
r_{\sa;t;i}=&\sum_{j=1}^{M\hid}X_{ij}\sigma\le(w_{\sa;t;j}\ri)\, \ \ \ \text{for}\ \ \ i=1,\ldots,\hid\, ,
\end{align}
where $\sigma$ is an activation function and $M$ is an MLP multiplier, typically set to $M=4$ for sufficiently-large Transformers.\footnote{For some variant of the MLP block, see, e.g., Ref.~\cite{shazeer2020glu}.}

\subsubsection{Head Block}\label{subsubsec:forward-head}
\begin{center}\textit{Vision: linear classification layer}\end{center}

For  Vision Transformers, the head block often simply consists of a standard linear layer as
\be
z^{(L)}_{\sa;t;i}=\headb_{i}+\sum_{j=1}^{\hid}\head_{ij} s^{(L-1)}_{\sa;t;j}\, \ \ \text{for}\ \ i=1,\ldots,n_{\text{out}}\, ,
\ee
where, e.g., for image classification tasks, $n_{\text{out}}\equiv n_{\text{class}}$ is the number of the classes.\footnote{In one implementation -- which we'll follow in our Vision-Transformer experiments in \S\ref{subsec:ViT} -- the actual outputs are given by mean-pooling in the token direction as
\be
\widetilde{z}^{(L)}_{\sa;i}=\frac{1}{T}\sum_{t=1}^{T}z^{(L)}_{\sa;t;i}=\headb_{i}+\sum_{j=1}^{\hid}\head_{ij} \le(\frac{1}{T}\sum_{t=1}^{T}s^{(L-1)}_{\sa;t;j}\ri)=\headb_{i}+\sum_{j=1}^{\hid}\head_{ij} \widetilde{s}^{(L-1)}_{\sa;j}\, ,\label{eq:mean-pool}
\ee
where $\widetilde{s}^{(L-1)}_{\sa;j}\equiv (1/T)\sum_{t=1}^T s^{(L-1)}_{\sa;t;j}$.
%Practically, it is faster to first pool then act by the linear layer instead of the other way around.
In another implementation -- which we mentioned in footnote~\ref{foot:class} but won't follow  -- the class token is attached at $t=0$ in the stem block and the output would be given by $\widetilde{z}^{(L)}_{\sa;i}=z^{(L)}_{\sa;0;i}$.}

\begin{center}\textit{Language: word embedding, transposed (and rescaled)}\end{center}

For Language Transformers, we typically tie weights between the stem and head blocks~\cite{press2016using}, which means that we multiply the signals by a \textit{transpose} of word-embedding parameters as
\be
z^{(L)}_{\sa;t;i}=\rescale\sum_{j=1}^{\hid}(\WE)^{\!\top}_{\ ij} s^{(L-1)}_{\sa;t;j}=\rescale \sum_{j=1}^{\hid} \WE_{ji}s^{(L-1)}_{\sa;t;j}\, \ \ \text{for}\ \ i=1,\ldots,n_{\text{out}}\, ,
\ee
with $n_{\text{out}}\equiv\nvocab$.
Here, we've also introduced the rescaling factor $\rescale$, with the reason of its existence -- and its width scaling  -- to be elucidated first in \S\ref{subsubsec:initialization-head} and vindicated again in \S\ref{subsubsec:backward-head}.

\newpage
\subsection{Statistics of Preactivations}\label{subsec:initialization}
Before any training happens, the model parameters need to be initialized in some way. Typically, they are initialized by independently and identically drawing them from \textit{mean-zero}  uniform, normal, or truncated-normal distributions. For notational housekeeping, let us stipulate up front that their covariances should be scaled as
\begin{align}
\E{\patch_{i_1j_1}\patch_{i_2j_2}}=&\le(\frac{C_{\text{patch}}}{\npatch}\ri) \delta_{i_1i_2}\delta_{j_1j_2}\, \ \ \ \ \ \ \ \ \ \ \ \ \ \ \ \ \Big\vert\Big\vert\ \ \E{\WE_{i_1j_1}\WE_{i_2j_2}}=\le(C_{\text{WE}}\ri) \delta_{i_1i_2}\delta_{j_1j_2}\, ,\label{eq:init-preview-stem}\\
\E{\PE_{t_1;i_1}\PE_{t_2;i_2}}=&\le(C_{\text{PE}}\ri) \delta_{t_1t_2}\delta_{i_1i_2}\, ,\label{eq:init-preview-PE}\\
\E{Q_{c_1i_1}^{h_1}Q_{c_2i_2}^{h_2}}=&\le(\frac{C_Q}{\hid}\ri)\delta_{c_1c_2}\delta_{i_1i_2}\delta^{h_1h_2}\, ,\label{eq:init-preview-Q}\\
\E{K_{c_1i_1}^{h_1}K_{c_2i_2}^{h_2}}=&\le(\frac{C_K}{\hid}\ri)\delta_{c_1c_2}\delta_{i_1i_2}\delta^{h_1h_2}\, ,\label{eq:init-preview-K}\\
\E{V_{c_1i_1}^{h_1}V_{c_2i_2}^{h_2}}=&\le(\frac{C_V}{\hid}\ri)\delta_{c_1c_2}\delta_{i_1i_2}\delta^{h_1h_2}\, ,\label{eq:init-preview-V}\\
\E{U_{i_1c_1}^{h_1}U_{i_2c_2}^{h_2}}=&\le(\frac{C_U}{\hid}\ri)\delta_{i_1i_2}\delta_{c_1c_2}\delta^{h_1h_2}\, ,\label{eq:init-preview-U}\\
\E{W_{i_1j_1}W_{i_2j_2}}=&\le(\frac{C_W}{\hid}\ri)\delta_{i_1i_2}\delta_{j_1j_2}\, ,\label{eq:init-preview-W}\\
\E{X_{i_1j_1}X_{i_2j_2}}=&\le(\frac{C_X}{M\hid}\ri)\delta_{i_1i_2}\delta_{j_1j_2}\, ,\label{eq:init-preview-X}\\
\E{\head_{i_1j_1}\head_{i_2j_2}}=&\le(\frac{C_{\text{head}}}{\hid}\ri)\delta_{i_1i_2}\delta_{j_1j_2}\, ,\   b_i^{\text{head}}=0\, \ \ \Big\vert\Big\vert\ \ \rescale=\sqrt{\frac{1}{\hid}}\, ,\label{eq:init-preview-head}
\end{align}
where $\E{\cdot}$ denotes an expectation value with respect to the initialization distribution, $\delta_{ij}$ is the Kronecker delta (i.e., $\delta_{ij}=1$ when $i=j$ and $\delta_{ij}=0$ when $i\ne j$), and all the initialization hyperparameters $C_{G}$ are order-one numbers for each group $G$ of model parameters. Here, by ``order-one,''  we mean that these hyperparameters are fixed when we scale up the width $n$, depth $L$, and any other architecture hyperparameters of the networks; we'll also often say ``observables stay of order one'' to mean that the said observable numbers don't blow up to infinity or vanish down to zero as networks are scaled up indefinitely.\footnote{More generally,\label{foot:Meta1} if we'd like to employ generic meta-principled scaling strategies~\cite{yaida2022meta}  that interpolate the neural-tangent scaling strategy~\cite{jacot2018neural} (at $s=0$) and maximal-update scaling strategy~\cite{yang2021tensor} (at $s=1$), then the head scalings~\eqref{eq:init-preview-head} should be modified to
\be
\E{W^{\text{head}}_{i_1j_1}W^{\text{head}}_{i_2j_2}}=\le(\frac{C_{\text{head}}}{\hid^{1+s}}\ri)\delta_{i_1i_2}\delta_{j_1j_2}\, ,\ \  b_i^{\text{head}}=0\, \ \ \Big\vert\Big\vert\ \ \mathcal{N}_{\text{rescale}}=\sqrt{\frac{1}{\hid^{1+s}}}\, ,\label{eq:init-preview-Meta-head}
\ee
In the main text, we'll focus on the neural-tangent scaling strategy with $s=0$, and we'll mention the corresponding meta-change for learning-rate factors only in future footnote~\ref{foot:Meta2}.}

For the rest of this section, we'll see how these width scalings of initialization hyperparameters~\eqref{eq:init-preview-stem}--\eqref{eq:init-preview-head} are chosen to ensure that the preactivations $z^{(\blocki)}_{\sa;t;i}$ stay of order one as we widen Transformers. More specifically, we'll recursively show that their covariances take the form
\be
\E{z^{(\blocki)}_{\sa_1;t_1;i_1}z^{(\blocki)}_{\sa_2;t_2;i_2}}=\delta_{i_1i_2} G^{(\blocki)}_{(\sa_1;t_1)(\sa_2;t_2)}\, ,\label{eq:forward-goal}
\ee
with order-one kernels $G^{(\blocki)}_{(\sa_1;t_1)(\sa_2;t_2)}$ and we'll also show in \S\ref{subsubsec:initialization-LN} that the expected squared norms of signals right after the layer-normalization operation are given by
\be
\kernelLN^{(\blocki)}_{(\sa_1;t_1)(\sa_2;t_2)}\equiv\E{\frac{1}{\hid}\sum_{i=1}^{\hid}s^{(\blocki)}_{\sa_1;t_1;i}s^{(\blocki)}_{\sa_2;t_2;i}}=\frac{G^{(\blocki)}_{(\sa_1;t_1)(\sa_2;t_2)}}{\sqrt{G^{(\blocki)}_{(\sa_1;t_1)(\sa_1;t_1)}+\epsilon}\sqrt{G^{(\blocki)}_{(\sa_2;t_2)(\sa_2;t_2)}+\epsilon}}+O\le(\frac{1}{n}\ri)\, ,\label{eq:LN-kernel-preview}
\ee
at leading order in $1/n$. All in all, we'll show that, with the stipulated initialization hyperparameter scalings~\eqref{eq:init-preview-stem}--\eqref{eq:init-preview-head}, Transformers respect the principle of criticality~\cite{PDLT}, that is, signals stay of order one 
on forward paths.\footnote{That is, at a meta level. If there were no layer normalization, then we would have further followed the non-meta principle of criticality~\cite{poole2016exponential,raghu2017expressive,schoenholz2016deep,PDLT} to fine-tune initialization hyperparameters $C_{G}$'s and (initial) element-wise affine parameters $(\gamma^{(\blocki)}_i, \beta^{(\blocki)}_i,  \xerox^{(\blocki)}_i)$ so as to avoid exponentially exploding or vanishing signal problems; since there typically \textit{are} normalization layers, we'll see in \S\ref{subsubsec:initialization-MHSA} and  \S\ref{subsubsec:initialization-MLP} that the kernel $G^{(\blocki)}$ grows only linearly with depth $\blocki$ -- until it finally gets further normalized right before entering the head block at $\blocki=L$ (\S\ref{subsubsec:initialization-head}).}

With those goals in mind, let's analyze the statistics of preactivations forwardly, block by block.

\subsubsection{Stem Block}\label{subsubsec:initialization-stem}
\begin{center}\textit{Vision: patchify embedding and positional embedding}\end{center}

In the stem block of Vision Transformers, which outputs the first-block preactivations $z^{(1)}_{\sa;t;i}=\PE_{t;i}+\sum_{j=1}^{\npatch}\patch_{ij}x_{\sa;t;j}$, patchify weights and positional-embedding parameters are initialized with mean-zero distributions whose covariances are scaled as
\be
\E{\patch_{i_1j_1}\patch_{i_2j_2}}=\le(\frac{C_{\text{patch}}}{\npatch}\ri) \delta_{i_1i_2}\delta_{j_1j_2}\, ,\ \ \ \E{\PE_{t_1;i_1}\PE_{t_2;i_2}}=\le(C_{\text{PE}}\ri) \delta_{t_1t_2}\delta_{i_1i_2}\, ,
\ee
with order-one initialization hyperparameters $C_{\text{patch}}$ and $C_{\text{PE}}$, respectively. This way, we have the vanishing mean,
\be
\E{z^{(1)}_{\sa;t;i}}=0\, ,
\ee
and order-one covariance,
\begin{align}
\E{z^{(1)}_{\sa_1;t_1;i_1}z^{(1)}_{\sa_2;t_2;i_2}}=&\E{\le(\PE_{t_1;i_1}+\sum_{j_1=1}^{\npatch}\patch_{i_1j_1}x_{\sa_1;t_1;j_1}\ri)\le(\PE_{t_2;i_2}+\sum_{j_2=1}^{\npatch}\patch_{i_2j_2}x_{\sa_2;t_2;j_2}\ri)}\, \\
=&C_{\text{PE}}\delta_{i_1i_2} \delta_{t_1t_2}+\delta_{i_1i_2}C_{\text{patch}} \le(\frac{1}{\npatch}\sum_{j=1}^{\npatch}x_{\sa_1;t_1;j}x_{\sa_2;t_2;j}\ri)\, \nonumber\\
=&\delta_{i_1i_2} \le(C_{\text{PE}}\delta_{t_1t_2}+C_{\text{patch}}G^{(0)}_{(\sa_1;t_1)(\sa_2;t_2)}\ri)\, \nonumber\\
=&\delta_{i_1i_2} G^{(1)}_{(\sa_1;t_1)(\sa_2;t_2)}\, .\nonumber
\end{align}
Here, in the penultimate line, we introduced the input kernel,
\be
G^{(0)}_{(\sa_1;t_1)(\sa_2;t_2)}\equiv\frac{1}{\npatch}\sum_{j=1}^{\npatch}x_{\sa_1;t_1;j}x_{\sa_2;t_2;j}\, ,\label{eq:input-kernel-vision}
\ee
which is of order one since all pixel values are usually preprocessed to be of order one, and in the last line, we introduced the stem-block kernel,
\be
G^{(1)}_{(\sa_1;t_1)(\sa_2;t_2)}\equiv C_{\text{patch}} G^{(0)}_{(\sa_1;t_1)(\sa_2;t_2)}+C_{\text{PE}} \delta_{t_1t_2}\, ,\label{eq:stem-kernel-vision}
\ee
which  is consequently of order one.
%Occasionally, pixel values are in the range $[0,255]$; in that case, to implement our theory in practice, we should first preprocess them to be roughly in the range $[-1,1]$ (we could alternatively adjust scalings of initialization and stem-block learning rates, but that's much more cumbersome).
In particular, each component of the first-block preactivations is expected to be of order one:  $\E{\le(z^{(1)}_{\sa;t;i}\ri)^2}=G^{(1)}_{(\sa;t)(\sa;t)}=C_{\text{patch}} \le(\frac{1}{\npatch}\sum_{j=1}^{\npatch}x_{\sa;t;j}^2\ri)+C_{\text{PE}}=O(1)$.\\

\begin{center}\textit{Language: word embedding and positional embedding}\end{center}

In the stem block of Language Transformers, which outputs the first-block preactivations $z^{(1)}_{\sa;t;i}=\PE_{t;i}+\sum_{j=1}^{\nvocab}\WE_{ij}x_{\sa;t;j}$, 
the positional-embedding parameters are initialized as in the vision case with zero mean and the covariance $\E{\PE_{t_1;i_1}\PE_{t_2;i_2}}=C_{\text{PE}} \delta_{t_1t_2}\delta_{i_1i_2}$, while
the word-embedding parameters are initialized with zero mean and the covariance
\be
\E{\WE_{i_1j_1}\WE_{i_2j_2}}=\le(C_{\text{WE}}\ri) \delta_{i_1i_2}\delta_{j_1j_2}\, ,\label{eq:init-WE}
\ee
where the initialization hyperparameter $C_{\text{WE}}$ is of order one.
Here, note that we did \textit{not} divide the covariance by the vocabulary size $\nvocab$; this way, we have
\begin{align}
\E{z^{(1)}_{\sa_1;t_1;i_1}z^{(1)}_{\sa_2;t_2;i_2}}=&\E{\le(\PE_{t_1;i_1}+\sum_{j_1=1}^{\nvocab}\WE_{i_1j_1}x_{\sa_1;t_1;j_1}\ri)\le(\PE_{t_2;i_2}+\sum_{j_2=1}^{\nvocab}\WE_{i_2j_2}x_{\sa_2;t_2;j_2}\ri)}\, \nonumber\\
=&\delta_{i_1i_2}\le[C_{\text{PE}} \delta_{t_1t_2}+C_{\text{WE}} \le(\sum_{j=1}^{\nvocab}x_{\sa_1;t_1;j}x_{\sa_2;t_2;j}\ri)\ri]\, \nonumber\\
=&\delta_{i_1i_2}\le[C_{\text{PE}} \delta_{t_1t_2}+C_{\text{WE}}G^{(0)}_{(\sa_1;t_1)(\sa_2;t_2)} \ri]\, \nonumber\\
=&\delta_{i_1i_2} G^{(1)}_{(\sa_1;t_1)(\sa_2;t_2)}\, ,
\end{align}
where the input kernel
\be
G^{(0)}_{(\sa_1;t_1)(\sa_2;t_2)}\equiv\sum_{j=1}^{\nvocab}x_{\sa_1;t_1;j}x_{\sa_2;t_2;j}\, ,\label{eq:input-kernel-language}
\ee
is of order one due to the one-hot structure of word inputs, and the stem-block kernel
\be
G^{(1)}_{(\sa_1;t_1)(\sa_2;t_2)}\equiv C_{\text{WE}} G^{(0)}_{(\sa_1;t_1)(\sa_2;t_2)}+C_{\text{PE}} \delta_{t_1t_2}\, ,\label{eq:stem-kernel-language}
\ee
is consequently of order one as well. In particular the input kernel $G^{(0)}_{(\sa_1;t_1)(\sa_2;t_2)}$ is one if and only if $(\sa_1;t_1)$ and $(\sa_2;t_2)$ correspond to the same token in the vocabulary, and zero otherwise.

\subsubsection{Layer Normalization}\label{subsubsec:initialization-LN}
Here, we'll analyze how each layer-normalization operation~\eqref{eq:LNbare} transforms the kernel. 
Before doing so, we note that layer normalization operates only over the embedding direction $i$ and in particular acts independently on each pair of sample--token indices $(\sa;t)$, so, to declutter our analysis,  we'll drop these latter indices along with the block index $\blocki$ for now; we'll add these indices back after all the dust settles. With those in mind, we'll be analyzing the baby version of the object~\eqref{eq:LN-kernel-preview},
\be\label{eq:kernelLN-simple}
\kernelLN\equiv\E{\frac{1}{\hid}\sum_{i=1}^{\hid}s_{i}s_{i}}=\frac{\le(\frac{1}{\hid}\sum_{i=1}^{\hid}z_i^2\ri)-\le(\frac{1}{\hid}\sum_{i=1}^{\hid}z_i\ri)^2}{\le[\sqrt{\le(\frac{1}{\hid}\sum_{i=1}^{\hid}z_i^2\ri)-\le(\frac{1}{\hid}\sum_{i=1}^{\hid}z_i\ri)^2+\epsilon}\ \ri]^2}\, ,
\ee
for a vectorial random variable $z_i$, with zero mean and order-one covariance
\be
\E{z_{i_1}z_{i_2}}=\delta_{i_1i_2} G\, .
\ee

Intuitively speaking, for sufficiently large $\hid$, we should have $\frac{1}{\hid}\sum_{i=1}^{\hid}z_i^2\approx G$ and -- as $z_i$'s are mean-zero order-one random numbers -- $\frac{1}{\hid}\sum_{i=1}^{\hid}z_i\sim \frac{\sqrt{\hid}}{\hid}\approx 0$. To make precise this intuition, let us introduce two types of finite-$\hid$ corrections: the instantiation-to-instantiation magnitude fluctuations,
\be
\widehat{\Delta G} \equiv \le(\frac{1}{\hid}\sum_{i=1}^{\hid}z_i^2\ri)-G\, ,
\ee
and the square of the mean of the preactivations,
\be
\widehat{\nabla G}\equiv \le(\frac{1}{\hid}\sum_{i=1}^{\hid}z_i\ri)^2\, .
\ee
With these definitions, we can concisely rewrite the baby object~\eqref{eq:kernelLN-simple} as
\be
\kernelLN=\frac{G+\widehat{\Delta G}-\widehat{\nabla G} }{\le[\sqrt{G+\epsilon+\widehat{\Delta G}-\widehat{\nabla G}}\ri]^2}\, .
\ee
We'll now show that
\be
\E{\widehat{\Delta G}^p\widehat{\nabla G}^q}=O\le(\frac{1}{n}\ri)\, 
\ee
for any integers $p,q\geq0$ with $p+q>0$ so that, at leading order, we can neglect these finite-$\hid$ corrections.\footnote{The result we'll derive was used in Ref.~\cite{doshi2021critical} on intuitive ground; what follows is a slightly more careful analysis.}

Let's slowly roll out the expression in eight steps (a wordy description follows these equation):
\begin{align}
&\E{\widehat{\Delta G}^p\widehat{\nabla G}^q}\, \label{eq:eightfold}\\
=&\mathbb{E}\le\{\le[\sum_{r=0}^{p}{p \choose r}\frac{\le(-G\ri)^{p-r}}{\hid^{r}}\sum_{i_1,\ldots,i_r=1}^{\hid}z_{i_1}^2\cdots z_{i_r}^2\ri]\le[\frac{1}{\hid^{2q}}\sum_{j_1,\ldots,j_{2q}=1}^{\hid}z_{j_1}\cdots z_{j_{2q}}\ri]\ri\}\, \nonumber\\
=&\sum_{r=0}^{p}{p \choose r}\le(-G\ri)^{p-r}\frac{1}{\hid^{r+2q}}\sum_{i_1,\ldots,i_r=1}^{\hid}\sum_{j_1,\ldots,j_{2q}=1}^{\hid}\E{z_{i_1}^2\cdots z_{i_r}^2z_{j_1}\cdots z_{j_{2q}}}\, \nonumber\\
=&\sum_{r=0}^{p}{p \choose r}\le(-G\ri)^{p-r}\frac{\hid(\hid-1)\cdots (\hid-r-2q+1)}{\hid^{r+2q}}\E{z_1^2\cdots z_r^2 z_{r+1}\cdots z_{r+2q}}+O\le(\frac{1}{\hid}\ri)\, \nonumber\\
=&\sum_{r=0}^{p}{p \choose r}\le(-G\ri)^{p-r}\E{z_1^2}\cdots \E{z_r^2}\E{z_{r+1}} \cdots \E{z_{r+2q}}+O\le(\frac{1}{n}\ri)\, \nonumber\\
=&\sum_{r=0}^{p}{p \choose r}\le(-G\ri)^{p-r}G^r 0^q+O\le(\frac{1}{n}\ri)\, \nonumber\\
=&G^p\sum_{r=0}^{p}{p \choose r}\le(-1\ri)^{p-r} 0^q+O\le(\frac{1}{n}\ri)\, \nonumber\\
=&G^{p}(1-1)^p0^q+O\le(\frac{1}{n}\ri)\, \nonumber\\
=&O\le(\frac{1}{n}\ri)\, .\nonumber
\end{align}
In the first step, we explicitly wrote out the expressions for $\widehat{\Delta G}$ -- using the binomial formula -- and $\widehat{\nabla G}$; in the second step, we pulled out the non-random coefficients and the summation marks outside the expectation; in the third step, we separated the sum into the terms with the embedding indices $i_1,\ldots,i_r,j_1,\ldots,j_{2q}$ all distinct, which give rise to the putatively leading-order contribution written there, 
% -- where we also used the permutation symmetry among neurons --
and the rest of the terms that have at least one coincident embedding index, which combinatorially give rise to at most $O\le(1/\hid\ri)$ contributions; in the fourth step, we picked up the leading contribution from the first term and dumped the rest into $O\le(1/\hid\ri)$;\footnote{To be more verbose, this fourth step involved two mini-steps: one is a simple algebra, $\hid(\hid-1)\cdots (\hid-r-2q+1)/\hid^{r+2q}=1+O\le(1/\hid\ri)$; the other is to neglect correlations among distinct neurons as subleading, that is, $\E{z_1^2\cdots z_r^2 z_{r+1}\cdots z_{r+2q}}=\E{z_1^2}\cdots \E{z_r^2}\E{z_{r+1}} \cdots \E{z_{r+2q}}+O\le(1/\hid\ri)$. See Ref.~\cite{PDLT} for more on the latter.} in the fifth step, we simply evaluated the expectations; in the sixth step, we pulled the common factor of $G^{p}$ out of the sum; in the seventh step, we used the binomial formula in reverse; and in the eighth step, we used the fact that $0^p 0^q=0$ for any $p,q\geq0$ with $p+q>0$.

With all the dust settled, judiciously putting back sample--token--block indices, we get
\be\label{eq:LN-kernel}
\kernelLN^{(\blocki)}_{(\sa_1;t_1)(\sa_2;t_2)}=\E{\frac{1}{\hid}\sum_{i=1}^{\hid}s^{(\blocki)}_{\sa_1;t_1;i}s^{(\blocki)}_{\sa_2;t_2;i}}=\frac{G^{(\blocki)}_{(\sa_1;t_1)(\sa_2;t_2)}}{\sqrt{G^{(\blocki)}_{(\sa_1;t_1)(\sa_1;t_1)}+\epsilon}\sqrt{G^{(\blocki)}_{(\sa_2;t_2)(\sa_2;t_2)}+\epsilon}}+O\le(\frac{1}{n}\ri)\, .
\ee
As a quick sanity check, when $(\sa_1;t_1)=(\sa_2;t_2)$ and $\epsilon=0$, we have  $\kernelLN^{(\blocki)}_{(\sa;t)(\sa;t)}=1$, as should be the case because $\frac{1}{\hid}\sum_{i=1}^{\hid}s_{\sa;t;i}s_{\sa;t;i}=1$ by construction.
Most importantly, even if we got an exponentially large kernel $G^{(\blocki)}\sim e^{\text{large}}$, they cancel between the numerator  and denominator to yield order-one numbers as $F^{(\blocki)}\sim \frac{e^{\text{large}}}{\sqrt{e^{\text{large}}+\epsilon}\sqrt{e^{\text{large}}+\epsilon}}=O(1)$. This way, normalization layers prevent exponentially exploding signal problems from ever happening, at least from block to block.
%It would also prevent a putative vanishing problem --  which is actually not a problem because skip connections add --  though it would give up if the kernel ever got less than epsilon.

\subsubsection{Multi-Head Self-Attention Block}\label{subsubsec:initialization-MHSA}
As explained in \S\ref{subsubsec:forward-MHSA}, the residual path of the MHSA block outputs
\be\label{eq:matrix-product-2-review}
r_{\sa;t;i}=\sum_{h=1}^{H}\sum_{t'=1}^{T}\sum_{c=1}^{C}\sum_{j=1}^{\hid} \ws^{h}_{\sa;tt'}U_{ic}^{h}V_{cj}^{h}s_{\sa;t';j}\, ,
\ee
where  the self-attention matrix $\ws^{h}_{\sa;tt'}=\ws^{h}_{\sa;tt'}\le[\wst\le(Q,K;s\ri)\ri]$ is an order-one function of the query--key dot product matrix $\wst_{\sa;\tilde{t}\tilde{t}'}^{h}$~\eqref{eq:matrix-product-1} given by
\be\label{eq:matrix-product-1-review}
\wst_{\sa;\tilde{t}\tilde{t}'}^{h}\equiv\frac{1}{\sqrt{C}}\sum_{c=1}^{C}q_{\sa;\tilde{t};c}^{h}k_{\sa;\tilde{t}';c}^{h}=\frac{1}{\sqrt{C}}\sum_{c=1}^{C}\sum_{i_1,i_2=1}^{\hid}Q_{ci_1}^{h}K_{ci_2}^{h}s_{\sa;\tilde{t};i_1}s_{\sa;\tilde{t}';i_2}\, .
\ee
We initialize the associated weights $Q_{ci}^{h}$, $K_{ci}^{h}$, $V_{ci}^{h}$, and $U_{ic}^{h}$ by drawing them from mean-zero distributions with the covariances
\begin{align}
\mathbb{E}\le[Q_{c_1i_1}^{h_1}Q_{c_2i_2}^{h_2}\ri]=&\le(\frac{C_Q}{\hid}\ri)\delta_{c_1c_2}\delta_{i_1i_2}\delta^{h_1h_2}\, ,\\
\mathbb{E}\le[K_{c_1i_1}^{h_1}K_{c_2i_2}^{h_2}\ri]=&\le(\frac{C_K}{\hid}\ri)\delta_{c_1c_2}\delta_{i_1i_2}\delta^{h_1h_2}\, ,\\
\mathbb{E}\le[V_{c_1i_1}^{h_1}V_{c_2i_2}^{h_2}\ri]=&\le(\frac{C_V}{\hid}\ri)\delta_{c_1c_2}\delta_{i_1i_2}\delta^{h_1h_2}\, ,\\
\mathbb{E}\le[U_{i_1c_1}^{h_1}U_{i_2c_2}^{h_2}\ri]=&\le(\frac{C_U}{\hid}\ri)\delta_{i_1i_2}\delta_{c_1c_2}\delta^{h_1h_2}\, ,
\end{align}
with order-one initialization hyperparameters $C_Q$, $C_K$, $C_V$, and $C_U$. (The associated bias parameters -- when they exist -- are typically initialized to zero.)
%In the backward path, after all the nightmare with the notations are settled, the gist is that we should upscale its learning rate by a factor of $\hid$ with respect to those for weights, though we would then probably have to further tune relative learning-rate factors for bias parameters and element-wise affine parameters.
To see why these are the right scalings, let's calculate the statistics of the query--key dot product~\eqref{eq:matrix-product-1-review} and residual-path output~\eqref{eq:matrix-product-2-review}.

First, the mean of the query--key dot product vanishes at initialization as
\begin{align}\label{eq:first-dot}
\E{\wst_{\sa;tt'}^{h}}=&\E{\frac{1}{\sqrt{C}}\sum_{c=1}^{C}\sum_{i,j=1}^{\hid}Q_{ci}^{h}K_{cj}^{h}s_{\sa;t;i}s_{\sa;t';j}}\, \\
=&\frac{1}{\sqrt{C}}\sum_{c=1}^{C}\sum_{i,j=1}^{\hid}\E{Q_{ci}^{h}}\E{K_{cj}^{h}}\E{s_{\sa;t;i}s_{\sa;t';j}}=0\, ,\nonumber
\end{align}
since these weights are independently drawn from mean-zero distributions. As for the covariance,
\begin{align}\label{eq:second-dot}
\E{\wst_{\sa_1;t_1t'_1}^{h_1}\wst_{\sa_2;t_2t'_2}^{h_2}}=&\frac{1}{C}\sum_{c_1,c_2=1}^{C}\sum_{i_1,i_2,j_1,j_2=1}^{\hid}\E{Q_{c_1i_1}^{h_1}K_{c_1j_1}^{h_1}s_{\sa_1;t_1;i_1}s_{\sa_1;t'_1;j_1}Q_{c_2i_2}^{h_2}K_{c_2j_2}^{h_2}s_{\sa_2;t_2;i_2}s_{\sa_2;t'_2;j_2}}\, \nonumber \\
=&\frac{1}{C}\sum_{c_1,c_2=1}^{C}\sum_{i_1,i_2,j_1,j_2=1}^{\hid}\frac{C_Q}{\hid}\frac{C_K}{\hid}\delta_{c_1c_2}\delta_{i_1i_2}\delta_{j_1j_2}\delta^{h_1h_2}\E{s_{\sa_1;t_1;i_1}s_{\sa_1;t'_1;j_1}s_{\sa_2;t_2;i_2}s_{\sa_2;t'_2;j_2}}\, \nonumber\\
=&\delta^{h_1h_2}C_Q C_K\E{\le(\frac{1}{\hid}\sum_{i=1}^{\hid}s_{\sa_1;t_1;i}s_{\sa_2;t_2;i}\ri)\le(\frac{1}{\hid}\sum_{j=1}^{\hid}s_{\sa_1;t'_1;j}s_{\sa_2;t'_2;j}\ri)}\, \nonumber\\
=&\delta^{h_1h_2}C_Q C_K \kernelLN_{(\sa_1;t_1)(\sa_2;t_2)}\kernelLN_{(\sa_1;t'_1)(\sa_2;t'_2)}+O\le(\frac{1}{n}\ri)\, ,
\end{align}
where in the last step we used our eightfold result~\eqref{eq:eightfold} to truncate away the $1/\hid$ corrections. Together, we expect $\wst_{tt'}^{h}$ to have zero mean and order-one covariance. Here, the rescaling of the query--key dot product with $1/\sqrt{C}=\sqrt{H/\hid}$  was critical as otherwise the covariance would have scaled nontrivially with the width $\hid$ and/or the number of heads $H$.\footnote{The intuition behind this lengthy math -- that the sum over $C$ mean-zero order-one random numbers scales like $\sqrt{C}$ --  was explained in the footnote 4 of the original Transformer paper~\cite{vaswani2017attention}.\label{foot:tensor-boo-boo} In contrast, the Tensor Program V~\cite{yang2022tensor} puts forth a different scaling strategy -- $1/C$ instead of $1/\sqrt{C}$ -- based on heuristics that query and key vectors might get correlated at some point in training.} Consequently, we expect the self-attention matrix to be of order one.\footnote{See Appendix~\ref{app:self-attention} for more fun with the statistics of the self-attention.}

Second, turning our attention to the residual-path output, it has vanishing mean at initialization,
\be
\E{r_{\sa;t;i}}=\sum_{h=1}^{H}\sum_{t'=1}^{T}\sum_{c=1}^{C}\sum_{j=1}^{\hid} \E{U_{ic}^{h}}\E{V_{cj}^{h}}\E{\ws^{h}_{\sa;tt'}s_{\sa;j;t'}}=0\, ,
\ee
and order-one covariance,
\begin{align}
&\E{r_{\sa_1;t_1;i_1}r_{\sa_2;t_2;i_2}}\, \\
=&\sum_{h_1,h_2=1}^{H}\sum_{t'_1,t'_2=1}^{T}\sum_{c_1,c_2=1}^{C}\sum_{j_1,j_2=1}^{\hid} \E{\ws^{h_1}_{\sa_1;t_1t'_1}U_{i_1c_1}^{h_1}V_{c_1j_1}^{h_1}s_{\sa_1;t'_1;j_1}\ws^{h_2}_{\sa_2;t_2t'_2}U_{i_2c_2}^{h_2}V_{c_2j_2}^{h_2}s_{\sa_2;t'_2;j_2}}\, \nonumber\\
=&\sum_{h_1,h_2=1}^{H}\sum_{t'_1,t'_2=1}^{T}\sum_{c_1,c_2=1}^{C}\sum_{j_1,j_2=1}^{\hid}\frac{C_U}{\hid}\frac{C_V}{\hid}\delta_{c_1c_2}\delta_{i_1i_2}\delta_{j_1j_2}\delta^{h_1h_2} \E{\ws^{h_1}_{\sa_1;t_1t'_1}s_{\sa_1;t'_1;j_1}\ws^{h_2}_{\sa_2;t_2t'_2}s_{\sa_2;t'_2;j_2}}\, \nonumber\\
=&\delta_{i_1i_2}C_U C_V\frac{C}{\hid^2}\sum_{h=1}^{H}\sum_{t'_1,t'_2=1}^{T}\sum_{j=1}^{\hid}\E{\ws^{h}_{\sa_1;t_1t'_1}\ws^{h}_{\sa_2;t_2t'_2}s_{\sa_1;t'_1;j}s_{\sa_2;t'_2;j}}\, \nonumber\\
=&\delta_{i_1i_2}C_U C_V\sum_{t'_1,t'_2=1}^{T}\E{\le(\frac{1}{H}\sum_{h=1}^{H}\ws^{h}_{\sa_1;t_1t'_1}\ws^{h}_{\sa_2;t_2t'_2}\ri)\le(\frac{1}{\hid}\sum_{i=1}^{\hid}s_{\sa_1;t'_1;i}s_{\sa_2;t'_2;i}\ri)}\, .\nonumber
\end{align}
Here, the term in each pair of the parentheses is expected to be of order one.\footnote{To be more specific, we can use the results of Appendix~\ref{app:self-attention}  to factor the expectation as\label{foot:MHSA-stats-101}
\begin{align}
&\E{\le(\frac{1}{H}\sum_{h=1}^{H}\ws^{h}_{\sa_1;t_1t'_1}\ws^{h}_{\sa_2;t_2t'_2}\ri)\le(\frac{1}{\hid}\sum_{i=1}^{\hid}s_{\sa_1;t'_1;i}s_{\sa_2;t'_2;i}\ri)}\, \label{eq:MHSA-stats-101} \\
=&\frac{1}{H}\sum_{h=1}^{H}\le(\E{\ws^{h}_{\sa_1;t_1t'_1}\ws^{h}_{\sa_2;t_2t'_2}}\ri)\kernelLN_{(\sa_1;t'_1)(\sa_2;t'_2)}+O\le(\frac{1}{C}\ri)\, ,\nonumber
\end{align}
and further express the remaining expectation as a $\le(\vert\mathcal{D}\vert T^2\ri)$-dimensional Gaussian integral.}

Overall, adding the skip path back in and inking the block indices, we have the following kernel transformation through the MHSA block:
\begin{align}
G^{(\blocki+1)}_{(\sa_1;t_1)(\sa_2;t_2)}=&G^{(\blocki)}_{(\sa_1;t_1)(\sa_2;t_2)}\, \\
&+C_U C_V\sum_{t'_1,t'_2=1}^{T}\E{\le(\frac{1}{H}\sum_{h=1}^{H}\ws^{h}_{\sa_1;t_1t'_1}\ws^{h}_{\sa_2;t_2t'_2}\ri)\le(\frac{1}{\hid}\sum_{i=1}^{\hid}s^{(\blocki)}_{\sa_1;t'_1;i}s^{(\blocki)}_{\sa_1;t'_2;i}\ri)}\, .\nonumber
\end{align}
In particular, we expect the second additive piece to be order-one  functional of the order-one layer-normalized kernel $F^{(\blocki)}$~\eqref{eq:LN-kernel} and thus expect the overall kernel $G^{(\blocki)}$ to grow linearly.

\subsubsection{Multilayer Perceptron Block}\label{subsubsec:initialization-MLP}
As explained in \S\ref{subsubsec:forward-MLP}, the residual path of the MLP block outputs
\be
r_{\sa;t;i}=\sum_{j=1}^{M\hid}X_{ij}\sigma\le(w_{\sa;t;j}\ri)=\sum_{j=1}^{M\hid}X_{ij}\sigma\le(\sum_{k=1}^{\hid}W_{jk}s_{\sa;t;k}\ri)\, .
\ee
We initialize the associated weights $W_{ij}$ and $X_{ij}$ by drawing them from mean-zero distributions with covariances
\begin{align}
\E{W_{i_1j_1}W_{i_2j_2}}=&\le(\frac{C_W}{\hid}\ri)\delta_{i_1i_2}\delta_{j_1j_2}\, ,\label{eq:init-W}\\
\E{X_{i_1j_1}X_{i_2j_2}}=&\le(\frac{C_X}{M\hid }\ri)\delta_{i_1i_2}\delta_{j_1j_2}\, .\label{eq:init-X}
\end{align}
Let's see why these are the right scalings.

First, in the middle layer, we get vanishing mean and order-one covariance as
\begin{align}
\E{w_{\sa;t;i}}=&\sum_{j=1}^{\hid}\E{W_{ij}}\E{s_{\sa;t;j}}=0\, ,\\
\E{w_{\sa_1;t_1;i_1}w_{\sa_2;t_2;i_2}}=&\sum_{j_1,j_2=1}^{\hid}\E{W_{i_1j_1}W_{i_2j_2}s_{\sa_1;t_1;j_1}s_{\sa_2;t_2;j_2}}\, \nonumber\\
=&C_{W}\delta_{i_1i_2}\E{\frac{1}{\hid}\sum_{j=1}^{\hid}s_{\sa_1;t_1;j}s_{\sa_2;t_2;j}}\, \\
=&\delta_{i_1i_2}C_{W}\kernelLN_{(\sa_1;t_1)(\sa_2;t_2)}\, .\nonumber
\end{align}
Indeed, at leading order in the $1/n$ expansion, we can further show that $w_{\sa;t;i}$ is governed by mean-zero Gaussian distributions with covariance $\delta_{i_1i_2}C_W F_{(\sa_1;t_1)(\sa_2;t_2)}$~\cite{PDLT}.\footnote{What the MLP block does is extensively covered in Ref.~\cite{PDLT} -- both at infinite width and at finite width -- for those interested in more details than necessary here. That said, to give a flavor of (non-)Gaussianity, let's examine the fourth moment of the middle-layer preactivations,
\begin{align}
&\E{w_{\sa_1;t_1;i_1}w_{\sa_2;t_2;i_2}w_{\sa_3;t_3;i_3}w_{\sa_4;t_4;i_4}}\, \\
=&\sum_{j_1,j_2,j_3,j_4=1}^{\hid}\E{W_{i_1j_1}W_{i_2j_2}W_{i_3j_3}W_{i_4j_4}s_{\sa_1;t_1;j_1}s_{\sa_2;t_2;j_2}s_{\sa_3;t_3;j_3}s_{\sa_4;t_4;j_4}}\, \nonumber\\
=&C_W^2\Bigg\{\delta_{i_1i_2}\delta_{i_3i_4}\E{\le(\frac{1}{\hid}\sum_{j=1}^{\hid}s_{\sa_1;t_1;j}s_{\sa_2;t_2;j}\ri)\le(\frac{1}{\hid}\sum_{k=1}^{\hid}s_{\sa_3;t_3;k}s_{\sa_4;t_4;k}\ri)}\, \nonumber\\
&\ \ \ \ +\delta_{i_1i_3}\delta_{i_2i_4}\E{\le(\frac{1}{\hid}\sum_{j=1}^{\hid}s_{\sa_1;t_1;j}s_{\sa_3;t_3;j}\ri)\le(\frac{1}{\hid}\sum_{k=1}^{\hid}s_{\sa_2;t_2;k}s_{\sa_4;t_4;k}\ri)}\, \nonumber\\
&\ \ \ \ +\delta_{i_1i_4}\delta_{i_2i_3}\E{\le(\frac{1}{\hid}\sum_{j=1}^{\hid}s_{\sa_1;t_1;j}s_{\sa_4;t_4;j}\ri)\le(\frac{1}{\hid}\sum_{k=1}^{\hid}s_{\sa_2;t_2;k}s_{\sa_3;t_3;k}\ri)}\Bigg\}+O\le(\frac{1}{\hid}\ri)\, \nonumber\\
=&\E{w_{\sa_1;t_1;i_1}w_{\sa_2;t_2;i_2}}\E{w_{\sa_3;t_3;i_3}w_{\sa_4;t_4;i_4}}+\E{w_{\sa_1;t_1;i_1}w_{\sa_3;t_3;i_3}}\E{w_{\sa_2;t_2;i_2}w_{\sa_4;t_4;i_4}}\, \nonumber\\
&+\E{w_{\sa_1;t_1;i_1}w_{\sa_4;t_4;i_4}}\E{w_{\sa_2;t_2;i_2}w_{\sa_3;t_3;i_3}}+O\le(\frac{1}{\hid}\ri)\, .\nonumber
\end{align}
In the second equality, we used the i.i.d.-ness of the initialization distribution; specifically, the first term comes from when $(i_1,j_1)=(i_2,j_2)$ and $(i_3,j_3)=(i_4,j_4)$ and similarly for the other two terms. [For the normal distribution, this step is exact and is a special case of Wick's theorem; for non-normal distributions, there is a missing non-Gaussian term when $i_1=i_2=i_3=i_4$ and  $j_1=j_2=j_3=j_4$, which leads to the contribution proportional to $(1/\hid)\delta_{i_1i_2i_3i_4}=(1/\hid)\delta_{i_1i_2}\delta_{i_1i_3}\delta_{i_1i_4}$.]
In the last equality, we used our eightfold result~\eqref{eq:eightfold} to truncate away the $1/n$ correction.
Overall, this result on the fourth moment can be concisely summarized as
\be
\E{w_{\sa_1;t_1;i_1}w_{\sa_2;t_2;i_2}w_{\sa_3;t_3;i_3}w_{\sa_4;t_4;i_4}}\big\vert_{\text{connected}}=O\le(\frac{1}{\hid}\ri)\, .
\ee}
In particular, to evaluate various expectation values, we can use the factorization formula
\begin{align}\label{eq:GP}
\E{f_1(w_{\sa_1;t_1;i_1})\cdots f_p(w_{\sa_p;t_p;i_p})}=\le\langle f_1\le(\widetilde{w}_{\sa_1;t_1}\ri)\ri\rangle_{C_{W}\kernelLN}\cdots \le\langle f_p\le(\widetilde{w}_{\sa_p;t_p}\ri)\ri\rangle_{C_{W}\kernelLN}+O\le(\frac{1}{\hid}\ri)\, ,
\end{align}
for distinct embedding indices $i_1,\ldots,i_p$ and any functions $f_1,\ldots,f_p$. Here, $\langle \cdot\rangle_{K}$ with the kernel $K_{(\sa_1;t_1)(\sa_2;t_2)}$ in general denotes a $(\vert \mathcal{D}\vert T)$-dimensional Gaussian integral 
\begin{align}
\langle f\le(\widetilde{w}\ri)\rangle_{K}\equiv&\frac{1}{\sqrt{\det{\le(2\pi K\ri)}}}\int \le[\prod_{\sa\in\mathcal{D}}\prod_{t=1}^{T}d\widetilde{w}_{\sa;t}\ri] f\le(\widetilde{w}\ri)\, \\
&\ \ \ \ \ \ \ \ \ \ \ \ \ \ \ \ \ \ \ \ \ \ \times\exp\le[-\frac{1}{2}\sum_{\sa_1,\sa_2\in\mathcal{D}}\sum_{t_1,t_2=1}^{T}\widetilde{w}_{\sa_1;t_1} \le(K^{-1}\ri)^{(\sa_1;t_1)(\sa_2;t_2)} \widetilde{w}_{\sa_2;t_2}\ri]\, .\nonumber
\end{align}

Second, moving onto the residual-path output, we again have vanishing mean and order-one covariance as
\begin{align}
\E{r_{\sa;t;i}}=&\sum_{j=1}^{M\hid}\E{X_{ij}}\E{\sigma\le(w_{\sa;t;j}\ri)}=0\, ,\\
\E{r_{\sa_1;t_1;i_1}r_{\sa_2;t_2;i_2}}=&C_{X}\E{\frac{1}{M\hid}\sum_{j=1}^{M\hid} \sigma\le(w_{\sa_1;t_1;j}\ri)\sigma\le(w_{\sa_2;t_2;j}\ri)}\, \\
=&C_X\le\langle  \sigma\le(\widetilde{w}_{\sa_1;t_1}\ri) \sigma\le(\widetilde{w}_{\sa_2;t_2}\ri)\ri\rangle_{C_{W}\kernelLN}+O\le(\frac{1}{n}\ri)\, ,\nonumber
\end{align}
where in the last step we used the factorization formula~\eqref{eq:GP} just mentioned.

Combining two operations, adding the skip path back in, and inking the block indices, we have the kernel transformation through the MLP block
\be
G^{(\blocki+1)}_{(\sa_1;t_1)(\sa_2;t_2)}=G^{(\blocki)}_{(\sa_1;t_1)(\sa_2;t_2)}+C_X\le\langle  \sigma\le(\widetilde{w}_{\sa_1;t_1}\ri) \sigma\le(\widetilde{w}_{\sa_2;t_2}\ri)\ri\rangle_{C_{W}\kernelLN^{(\blocki)}}+O\le(\frac{1}{n}\ri)\, ,
\ee
where the layer-normalized kernel $F^{(\blocki)}$~\eqref{eq:LN-kernel} is given in terms of the previous-block kernel $G^{(\blocki)}$. In particular, as in the case of the MHSA block, we expect the second additive piece to be of order one due to the layer normalization and thus expect the kernel to linearly grow.\footnote{In contrast, if there were no normalization layer, then we would have instead gotten  -- say, e.g., for the \texttt{ReLU} activation --  something like $C_X\le\langle  \sigma\le(\widetilde{w}\ri) \sigma\le(\widetilde{w}\ri)\ri\rangle_{C_{W}G^{(\blocki)}}\sim \sharp G^{(\blocki)}$, which would have resulted in $G^{(\blocki+1)}\sim (1+\sharp)G^{(\blocki)}$ and hence an exponential explosion $\sim (1+\sharp)^{\blocki}$.}

\subsubsection{Head Block}\label{subsubsec:initialization-head}
\begin{center}\textit{Vision: linear classification layer}\end{center}

In the head block of Vision Transformers, which outputs the last-block preactivations $z^{(L)}_{\sa;t;i}=\headb_{i}+\sum_{j=1}^{\hid}\head_{ij} s^{(L-1)}_{\sa;t;j}$, we typically zero-initialize the head biases as $\headb_{i}=0$ while we draw the head weights $\head_{ij}$ from a mean-zero distribution with the covariance
\be
\E{\head_{i_1j_1}\head_{i_2j_2}}=\le(\frac{C_{\text{head}}}{\hid}\ri)\delta_{i_1i_2}\delta_{j_1j_2}\, .
\ee
As usual,  the mean of the output vanishes, while their covariance is given by
\be
\E{z^{(L)}_{\sa_1;t_1;i_1}z^{(L)}_{\sa_2;t_2;i_2}}= \delta_{i_1i_2} C_{\text{head}} \E{\frac{1}{\hid}\sum_{i=1}^{\hid}s^{(L-1)}_{\sa_1;t_1;i}s^{(L-1)}_{\sa_2;t_2;i}}=\delta_{i_1i_2} C_{\text{head}} F^{(L-1)}_{(\sa_1;t_1)(\sa_2;t_2)}\, .
\ee
That is, the output kernel
\be
G^{(L)}_{(\sa_1;t_1)(\sa_2;t_2)}=C_{\text{head}} F^{(L-1)}_{(\sa_1;t_1)(\sa_2;t_2)}\, ,
\ee
is manifestly of order one.\footnote{If we mean pool in the token direction as $\widetilde{z}^{(L)}_{\sa;i}=\frac{1}{T}\sum_{t=1}^{T}z^{(L)}_{\sa;t;i}$~\eqref{eq:mean-pool}, then the actual output kernel is given by
\be
\widetilde{G}^{(L)}_{\sa_1\sa_2}=\frac{1}{T^2} \sum_{t_1,t_2=1}^{T}G^{(L)}_{(\sa_1;t_1)(\sa_2;t_2)}\, .
\ee
If we attach a class token at $t=0$ and take $\widetilde{z}^{(L)}_{\sa;i}=z^{(L)}_{\sa;0;i}$ as the output, then the output kernel is given by $\widetilde{G}^{(L)}_{\sa_1\sa_2}=G^{(L)}_{(\sa_1;0)(\sa_2;0)}$.}
Note that even the linear growth of the kernel $G^{(\blocki)}$ with the depth $\blocki$ cancels out in this last step because of the very last normalization layer.\\

\begin{center}\textit{Language: word embedding, transposed (and rescaled)}\end{center}

In the head block of Language Transformers, which outputs the last-block preactivations $z^{(L)}_{\sa;t;i}=\rescale \sum_{j=1}^{\hid} \WE_{ji}s^{(L-1)}_{\sa;t;j}$, we've already specified the covariance of the word-embedding parameters to be of order one as $\E{\WE_{i_1j_1}\WE_{i_2j_2}}=\le(C_{\text{WE}}\ri) \delta_{i_1i_2}\delta_{j_1j_2}$, so let's focus on the rescaling factor $\rescale$. Intuitively, the need for this rescaling factor $\rescale$ should be clear: since we are summing $n$ mean-zero random order-one numbers, it calls for the rescaling by
\be
\rescale=\sqrt{\frac{1}{n}}\, ,\label{eq:rescaling}
\ee
so as to maintain the outputs to be of order one.\footnote{That said, this rescaling factor was absent in the original paper~\cite{press2016using}. This output rescaling factor of $1/\sqrt{\hid}$ has been implemented in, for example, Ref.~\cite{chowdhery2022palm} (though we could not track the original reference) and the maximal-update version -- $\rescale=1/\hid$ -- was first implemented in Ref.~\cite{yang2022tensor}.}
Non-intuitively, we have
\begin{align}
\E{z^{(L)}_{\sa_1;t_1;i_1}z^{(L)}_{\sa_2;t_2;i_2}}=&\rescale^2\delta_{i_1i_2} C_{\text{WE}} \sum_{j=1}^{\hid} \E{s^{(L-1)}_{\sa_1;t_1;j}s^{(L-1)}_{\sa_2;t_2;j}}\, \label{eq:naive-head}\\
=&\delta_{i_1i_2} C_{\text{WE}}  \E{\frac{1}{\hid}\sum_{j=1}^{\hid}s^{(L-1)}_{\sa_1;t_1;j}s^{(L-1)}_{\sa_2;t_2;j}}=\delta_{i_1i_2} C_{\text{WE}} F^{(L-1)}_{(\sa_1;t_1)(\sa_2;t_2)}\, ,\nonumber
\end{align}
where in the second equality we used the rescaling prescription~\eqref{eq:rescaling}.
We thus again have the order-one output kernel
\be
G^{(L)}_{(\sa_1;t_1)(\sa_2;t_2)}=C_{\text{WE}} F^{(L-1)}_{(\sa_1;t_1)(\sa_2;t_2)}\, .
\ee

Actually this entire discussion neglected the subtle interlayer correlation created by weight tying between the stem and head blocks, which is discussed in the footnote here.\footnote{
The discussion of general case below in the next paragraph will be condensed and not particularly illuminating, so we first recommend working out a simple toy model consisting of three linear layers with a skip connection, $z^{(3)}_{\sa;t;i}=\rescale\sum_{j,k}^{\hid}\sum_{m=1}^{\nvocab}\WE_{ji}\le(W^{(2)}_{jk}+\delta_{jk}\ri)\WE_{km}x_{\sa;t;m}$ where normally-distributed word-embedding weights $\WE_{ij}$ have the covariance $C_{\text{WE}}$ and middle-block weights $W^{(2)}_{ij}$ have the covariance $C_{W^{(2)}}/\hid$.\label{foot:long} Then we get
\begin{align}
&\E{z^{(3)}_{\sa_1;t_1;i_1}z^{(3)}_{\sa_2;t_2;i_2}}\, \\
=&\delta_{i_1i_2}C_{\text{WE}}^2 \le(C_{W^{(2)}}+1\ri)G^{(0)}_{(\sa_1;t_1)(\sa_2;t_2)}+ C_{\text{WE}}^2\le(\hid+\frac{C_{W^{(2)}}}{\hid}\ri)x_{\sa_1;t_1;i_1}x_{\sa_2;t_2;i_2}+\frac{1}{\hid}\le(C_{W^{(2)}}+1\ri)\le(x_{\sa_1;t_1;i_2}x_{\sa_2;t_2;i_1}\ri)\, .\nonumber
\end{align}
Here the first term is the one discussed in the main text and the rest is largely ignorable, except the contribution $\hid C_{\text{WE}}^2 x_{\sa_1;t_1;i_1}x_{\sa_2;t_2;i_2}$ which comes from the skip path $\rescale\sum_{j=1}^{\hid}\sum_{m=1}^{\nvocab}\WE_{ji}\WE_{jm}x_{\sa;t;m}$. At first sight, this contact term seems the most dominant contribution, but note that the main contribution with $\delta_{i_1i_2}$ has $\nvocab$ nonzero components of order one, while this contact term has only one nonzero component of order $\hid$ so, as long as $\hid\lesssim\nvocab$, it is not obviously problematic. Nonetheless, if we'd like to get rid of this contact term, then the easiest solution would be to just drop the skip path in the second block.\\
To deal with more general cases, we can first study correlators $\E{\WE_{j_1i_1}\WE_{j_2i_2} z^{(1)}_{\sa_1;t_1;k_1} \cdots z^{(1)}_{\sa_p;t_p;k_p}}$ and show that, for any function $\mathcal{F}$ that depends on $\WE_{km}$ only through the first-block preactivations $z^{(1)}_{\sa;t;k}$,
\begin{align}
\E{\WE_{j_1i_1}\WE_{j_2i_2}\mathcal{F}}=&\frac{1}{\hid}C_{\text{WE}}\delta_{i_1i_2} \delta_{j_1j_2}\E{\mathcal{F}}+\frac{1}{\hid}C_{\text{WE}}^2\sum_{\widetilde{\sa}_1,\widetilde{\sa}_2\in\mathcal{D}}\sum_{\tilde{t}_1,\tilde{t}_2=1}^{T}x_{\widetilde{\sa}_1;\tilde{t}_1;i_1}x_{\widetilde{\sa}_2;\tilde{t}_2;i_2}\E{\frac{\partial^2\mathcal{F}}{\partial z^{(1)}_{\widetilde{\sa}_1;\tilde{t}_1;j_1}\partial z^{(1)}_{\widetilde{\sa}_2;\tilde{t}_2;j_2}}}\, .\nonumber
\end{align}
So, putting in $\mathcal{F}=s^{(L-1)}_{\sa_1;t_1;j_1}s^{(L-1)}_{\sa_2;t_2;j_2}$, the first term gives the naive contribution~\eqref{eq:naive-head} discussed in the main text while the second term gives rise to four more contributions
\begin{align}
&\frac{1}{\hid}C_{\text{WE}}^2 \sum_{\tilde{t}_1,\tilde{t}_2=1}^{T}x_{\sa_1;\tilde{t}_1;i_1}x_{\sa_2;\tilde{t}_2;i_2}\E{\le(\sum_{j_1=1}^{\hid}\frac{\partial s^{(L-1)}_{\sa_1;t_1;j_1}}{\partial z^{(1)}_{\sa_1;\tilde{t}_1;j_1}}\ri)\le(\sum_{j_2=1}^{\hid}\frac{\partial s^{(L-1)}_{\sa_2;t_2;j_2}}{\partial z^{(1)}_{\sa_2;\tilde{t}_2;j_2}}\ri)}\, \\
+&\frac{1}{\hid}C_{\text{WE}}^2 \sum_{j_1,j_2=1}^{\hid}\sum_{\tilde{t}_1,\tilde{t}_2=1}^{T}x_{\sa_1;\tilde{t}_1;i_2}x_{\sa_2;\tilde{t}_2;i_1}\E{\frac{\partial s^{(L-1)}_{\sa_1;t_1;j_2}}{\partial z^{(1)}_{\sa_1;\tilde{t}_1;j_1}}\frac{\partial s^{(L-1)}_{\sa_2;t_2;j_2}}{\partial z^{(1)}_{\sa_2;\tilde{t}_2;j_1}}}\, \nonumber\\
+&\frac{1}{\hid}C_{\text{WE}}^2 \sum_{j_1,j_2=1}^{\hid}\sum_{\tilde{t}_1,\tilde{t}_2=1}^{T}x_{\sa_1;\tilde{t}_1;i_1}x_{\sa_1;\tilde{t}_2;i_2}\E{\frac{\partial^2 s^{(L-1)}_{\sa_1;t_1;j_1}}{\partial z^{(1)}_{\sa_1;\tilde{t}_1;j_1}\partial z^{(1)}_{\sa_1;\tilde{t}_2;j_2}}s^{(L-1)}_{\sa_2;t_2;j_2}}\, \nonumber\\
+&\frac{1}{\hid}C_{\text{WE}}^2 \sum_{j_1,j_2=1}^{\hid}\sum_{\tilde{t}_1,\tilde{t}_2=1}^{T}x_{\sa_2;\tilde{t}_1;i_1}x_{\sa_2;\tilde{t}_2;i_2}\E{s^{(L-1)}_{\sa_1;t_1;j_1}\frac{\partial^2 s^{(L-1)}_{\sa_2;t_2;j_2}}{\partial z^{(1)}_{\sa_2;\tilde{t}_1;j_1}\partial z^{(1)}_{\sa_2;\tilde{t}_2;j_2}}}\, .\nonumber
\end{align}
Out of these four terms, the first one contains the contact term.  Specifically, to see that, we use the chain rule $\partial s^{(L-1)}_{\sa;t;j}/\partial z^{(1)}_{\sa;\tilde{t};j}=\sum_{k=1}^{\hid}\le(\partial s^{(L-1)}_{\sa;t;j}/\partial z^{(L-1)}_{\sa;t;k}\ri)\le(\partial z^{(L-1)}_{\sa;t;k}/\partial z^{(1)}_{\sa;\tilde{t};j}\ri)$, use the future mnemonic~\eqref{eq:LN-mnemonic} to pick up the leading term $\partial s^{(L-1)}_{\sa;t;j}/\partial z^{(L-1)}_{\sa;t;k}\approx \delta_{jk}/\sqrt{G^{(L-1)}_{(\sa;t)(\sa;t)}+\epsilon}$, and then finally note that the dominant contribution to $\partial z^{(L-1)}_{\sa;t;j}/\partial z^{(1)}_{\sa;\tilde{t};j}$ comes from the skip path (essentially because the ``$j$'' in the $j$-th component of the residual-path output $\Resopp^{(\blocki)}_{t;j}$ has no meaningful association with the ``$j$'' in the $j$-th component of the first-block preactivation $z^{(1)}_{\sa;t;j}$ and hence its contribution is diluted and is down by powers of $1/\hid$).}
\newpage
\
\newpage
\subsection{A Crash Course on Neural Tangent Kernels}\label{subsec:NTK}
In this short crash course, we'll give a condensed introduction to the concept of neural tangent kernels~\cite{jacot2018neural}, both for the vanilla stochastic gradient descent (SGD) optimizer (\S\ref{subsubsec:SGD}) and for the AdamW optimizer (\S\ref{subsubsec:AdamW}); for a vaporized introduction, see, e.g., Ref.~\cite{PDLT}. This introduction paves the way for our effective-theory analysis of the backward path in \S\ref{subsec:backward}, where we'll figure out how to scale a relative learning-rate factor for each group of model parameters in Transformers.

\subsubsection{Vanilla SGD}\label{subsubsec:SGD}
The SGD update equation is given by
\be\label{eq:SGD-simple}
\theta_{\mu}(t)=\theta_{\mu}(t-1)-\eta_t \frac{\partial \loss_{\A_{t}}}{\partial \theta_{\mu}}\bigg\vert_{\theta=\theta(t-1)}\, ,
\ee
where the model-parameter index $\mu$ runs over all the $P$ model parameters $\theta_{\mu}$ in the architecture, $\eta_t$ is a learning rate at iteration $t$, $\loss_{\A_{t}}$ denotes a loss function evaluated on a minibatch $\A_t$ at iteration $t$, and $\theta_{\mu}(0)$ are drawn from the initialization distribution that was extensively discussed in \S\ref{subsec:initialization}.\footnote{We apologize to the letter $t$ for our using it to represent both the iteration index $t$ and the token index $t$.} In this standard form, we assign the single learning rate $\eta_t$ for all the model parameters, but in theory we'll soon find that the learning rate for each group $G$ of model parameters must be scaled differently as we embiggen Transformers. [To see how model parameters in a Transformer can get partitioned into groups, look back at equations~\eqref{eq:init-preview-stem}--\eqref{eq:init-preview-head} in \S\ref{subsec:initialization} or look ahead to equations~\eqref{eq:SGDlambdaG-preview-stem}--\eqref{eq:AdamWlambdaG-preview-head} in \S\ref{subsec:backward}.] To that end, we generalize the SGD update equation~\eqref{eq:SGD-simple} to
\be\label{eq:SGD}
\theta_{\mu}(t)=\theta_{\mu}(t-1)-\eta_t\LambdaGmu \frac{\partial \loss_{\A_{t}}}{\partial \theta_{\mu}}\bigg\vert_{\theta=\theta(t-1)}\, ,
\ee
where $G(\mu)$ denotes a group $G$ to which the $\mu$-th model parameter belongs, and relative learning-rate factors $\LambdaG$'s can be used to balance the degrees of updates among various groups in the architecture. In particular, the standard SGD naively sets $\LambdaG=1$ for all $G$'s, i.e., it uses a uniform learning rate for all the model parameters, while we'll figure out how to sophisticatedly scale them with widths for all the model-parameter groups in Transformers.

To do so, we look at how the network function $f_{\sa;i}\equiv z^{(L)}_{\sa;i}$ gets updated (where we temporarily suppress the token index $t$ to mitigate the notational conflict). That is, Taylor-expanding the change in the function to the first order in the model-parameter update~\eqref{eq:SGD}, we get
\begin{align}\label{eq:Taylor-SGD}
f_{\sa;i}(t)=&f_{\sa;i}(t-1)-\eta_t \sum_{\mu=1}^{P}\LambdaGmu\le(\frac{\partial \loss_{\A_t}}{\partial \theta_{\mu}} \frac{\partial f_{\sa;i}}{\partial \theta_{\mu}}\ri)\bigg\vert_{\theta=\theta(t-1)}+\ldots\, \\
=&f_{\sa;i}(t-1)-\eta_t \sum_{\mu=1}^{P}\LambdaGmu\le[\sum_{\widetilde{\sa}\in\A_t}\sum_{j=1}^{n_{\text{out}}} \frac{\partial \loss_{\A_t}}{\partial f_{\widetilde{\sa};j}}\le(\frac{\partial f_{\widetilde{\sa};j}}{\partial \theta_{\mu}} \frac{\partial f_{\sa;i}}{\partial \theta_{\mu}}\ri)\ri]\Bigg\vert_{\theta=\theta(t-1)}+\ldots\, \nonumber\\
=&f_{\sa;i}(t-1)-\eta_t\sum_{\widetilde{\sa}\in\A_t}\sum_{j=1}^{n_{\text{out}}} \le(\frac{\partial \loss_{\A_t}}{\partial f_{\widetilde{\sa};j}} \NTKM_{(\widetilde{\sa};j)(\sa;i)}\ri)\bigg\vert_{\theta=\theta(t-1)}+\ldots\, .\nonumber
\end{align}
Here, in the last line, we've defined the neural tangent kernel as
\be\label{eq:NTK-SGD}
\NTKM_{(\sa_1;i_1)(\sa_2;i_2)}\equiv \sum_{\mu=1}^{P}\LambdaGmu\frac{\partial f_{\sa_1;i_1}}{\partial \theta_{\mu}} \frac{\partial f_{\sa_2;i_2}}{\partial \theta_{\mu}}\, .
\ee
In particular, as the derivative of the loss with respect to the network outputs, $\partial \loss_{\A_t}/\partial f_{\widetilde{\sa};j}$, is expected to be of order one for generic losses, our objective in \S\ref{subsec:backward} will be to scale relative learning-rate factors $\LambdaG$'s such that the neural tangent kernel $\NTKM_{(\sa_1;i_1)(\sa_2;i_2)}$ -- and hence the change in the network function -- stays of order one as Transformers get wider.

In passing, we mention that, in principle, the higher-order terms ``$\ldots$'' in the Taylor expansion~\eqref{eq:Taylor-SGD} can matter, and the neural tangent kernel itself can dynamically change too. However, these effects are all  $1/\hid$-suppressed and hence can be neglected at the (leading) order we are working in in this note.\footnote{For a more thorough treatment of these $1/n$ effects, see Ref.~\cite{PDLT} (see also Refs.~\cite{dyer2019asymptotics,hanin2019finite} for some original work). More generally, the leading perturbative corrections scale with width as $1/\hid^{1-s}$~\cite{yaida2022meta}, which reduces to $1/\hid$ for the neural-tangent scaling strategy at $s=0$ while being never perturbative for the maximal-update scaling strategy at $s=1$.}

\subsubsection{AdamW}\label{subsubsec:AdamW}
The update equations for AdamW~\cite{kingma2014adam,loshchilov2017decoupled}  -- here again generalized with the relative learning-rate factors $\LambdaGA$'s -- can be written as
\begin{align}
g^{\mu}(t)\equiv&\frac{\partial \loss_{\A_{t}}}{\partial \theta_{\mu}}\bigg\vert_{\theta=\theta(t-1)}\, ,\label{eq:AdamW1}\\
v^{\mu}(t)=&\beta_1 v^{\mu}(t-1)+(1-\beta_1)g^{\mu}(t)\, ,\label{eq:AdamW2}\\
u^{\mu\mu}(t)=&\beta_2 u^{\mu\mu}(t-1)+(1-\beta_2)\le[g^{\mu}(t)\ri]^2\, ,\label{eq:AdamW3}\\
\theta_{\mu}(t)=&(1-\eta_t\cdot \wdt)\theta_{\mu}(t-1)-\LambdaGAmu\frac{\eta_t}{\sqrt{\frac{u^{\mu\mu}(t)}{(1-\beta_2^{t})}}+\epsilon} \frac{v^{\mu}(t)}{(1-\beta_1^{t})}\, ,\label{eq:AdamW4}
\end{align}
where $v^{\mu}(0)=u^{\mu\mu}(0)=0$ while $\theta_{\mu}(0)$ are again drawn from the initialization distribution. Here, the optimizer hyperparameters $\beta_1$ and $\beta_2$ respectively set the decay rates for the running averages of the first and second moments of per-parameter gradients -- roughly taking averages over the past $1/(1-\beta_{\sharp})$ iterations -- while the hyperparameters $\eta_t$ and $\wdt$ are the global learning rate and weight decay, respectively. In particular, in words, the last update equation~\eqref{eq:AdamW4} can be viewed as adapting the \textit{per-parameter} learning rate as $\eta \LambdaGAmu/\sqrt{u^{\mu\mu}}$, i.e., according to the inverse square root of the \textit{per-parameter} running average of the squared gradients.\footnote{Let us explain other factors in the last AdamW update equation~\eqref{eq:AdamW4} in more detail: the dividing factors of $(1-\beta_1^{t})$ and $(1-\beta_2^{t})$ are the bias-correction terms for the moment estimators $v^{\mu}$ and $u^{\mu\mu}$, respectively~\cite{kingma2014adam}, and the regularization factor of $\epsilon$ in the denominator sets the cutoff scale for gradients, that is, for the model parameter component $\mu$ with $\vert g^{\mu}\vert\gg\epsilon$ the update is basically $\eta\cdot\text{sign}(g^{\mu})$ whereas for $\vert g^{\mu}\vert\ll\epsilon$ the magnitude of the update starts to diminish in proportion to the magnitude of the gradient, albeit with an inflated learning rate $\eta/\epsilon$.\\
While we are on the topic of the AdamW optimizer hyperparameters, intuitively, with the neural-tangent scalings we employ in this note, gradient information doesn't change at leading order, so we can view gradients as fixed at initialization and forget about the effects of $\beta_1$ and $\beta_2$ as far as the leading-order theoretical analysis goes. As for the weight decay, properly carrying through the exercise on the regularization proposed in the footnote of \S10 in Ref.~\cite{PDLT}, we find that the product of the global  learning rate and weight decay, $\eta_t\cdot\wdt$, stays of order one: roughly, in their notation, the learning-rate tensor has lower indices as $\lambda_{\mu\nu}$ while the weight-decay tensor comes in as $\sum_{\mu,\nu=1}^{P}a^{\mu\nu}\theta_{\mu}\theta_{\nu}$ with upper indices, and $\lambda_{\mu\mu}$ scales like $1/\fanin$ while $a^{\mu\mu}$ scales like $\fanin$, so their product $\lambda_{\mu\mu}a^{\mu\mu}$ is of order one.}

Let us gedanken-Taylor-expand the change in the function to the first order in the model-parameter update~\eqref{eq:AdamW4}, as we did before for SGD. We then note that the essential difference between SGD and AdamW is the factor of $\sqrt{u^{\mu\mu}}\sim\vert g^{\mu}\vert$ in the denominator of the parameter update, that is, we get the (AdamW-modified) neural tangent kernel of the form
\be\label{eq:NTK-AdamW}
\widetilde{\NTKM}_{(\sa_1;i_1)(\sa_2;i_2)}\equiv \sum_{\mu=1}^{P}\frac{\LambdaGAmu}{\vert g^{\mu}\vert}\frac{\partial f_{\sa_1;i_1}}{\partial \theta_{\mu}} \frac{\partial f_{\sa_2;i_2}}{\partial \theta_{\mu}}\, .
\ee
We'll accordingly need to scale $\LambdaGA$'s to keep this modified neural tangent kernel of order one so that, again, the change in the network function stays of order one.

\sbreak
In our analysis in \S\ref{subsec:backward}, to be economical, we'll take the following strategy to deal with $\LambdaG$ for SGD and $\LambdaGA$ for AdamW together: for each group $G$ of model parameters, we'll \textit{(i)} first figure out the appropriate scaling of the relative learning-rate factor $\LambdaG$ for SGD so that the neural tangent kernel~\eqref{eq:NTK-SGD} stays of order one, \textit{(ii)} then estimate the width scaling of the expected per-parameter gradient magnitude $\vert g^{\mu}\vert$ for model parameters in the group $G$, and \textit{(iii)} accordingly adjust the relative learning-rate factor $\LambdaGAmu\sim\LambdaGmu \vert g^{\mu}\vert$ for AdamW so that the modified neural tangent kernel~\eqref{eq:NTK-AdamW} stays of order one.

\newpage
\subsection{Statistics of Neural Tangent Kernels}\label{subsec:backward}
In the previous few pages, we've identified the neural tangent kernel,
\be\label{eq:outputNTK-def}
\NTK_{(\sa_1;t_1;i_1)(\sa_2;t_2;i_2)}\equiv \sum_{\mu=1}^{P}\LambdaGmu\frac{\partial z^{(L)}_{\sa_1;t_1;i_1}}{\partial \theta_{\mu}} \frac{\partial z^{(L)}_{\sa_2;t_2;i_2}}{\partial \theta_{\mu}}\, ,
\ee
as the central object of interest in identifying the proper scalings of  per-group learning rates $\LambdaG$'s. That is, the neural tangent kernel dictates the update in the network function, and we would like to keep it of order one so that it doesn't explode or vanish in scaling up the models. (Here, we've decorated the neural tangent kernel with a \textit{hat}~\cite{PDLT} to connote that we'll be studying the statistics -- or, in this note, just the mean -- of this object \textit{at initialization} and also restored the token indices $t_1, t_2$ as there won't be any conflict with the iteration index $t$ at this point.\footnote{As mentioned in \S\ref{subsec:NTK}, the neural tangent kernel is fixed at initialization at leading order~\cite{jacot2018neural,PDLT}.})

To carry out our analysis, it is useful to introduce the $\blocki$-th-block neural tangent kernel
\be\label{eq:blockNTK-def}
\NTK^{(\blocki)}_{(\sa_1;t_1;i_1)(\sa_2;t_2;i_2)}\equiv\sum_{\mu=1}^{P}\LambdaGmu\frac{\partial z^{(\blocki)}_{\sa_1;t_1;i_1}}{\partial \theta_{\mu}}\frac{\partial z^{(\blocki)}_{\sa_2;t_2;i_2}}{\partial \theta_{\mu}}\, ,
\ee
which equals the actual neural tangent kernel~\eqref{eq:outputNTK-def} when $\blocki=L$. In particular, these objects enable us to carry out a \textit{forward} recursive analysis that very much mirrors our analysis of the statistics of preactivations in \S\ref{subsec:initialization}. Specifically, by recalling the main forward equation~\eqref{eq:resrec} and using the chain rule to roll out derivatives, we have
\begin{align}
&\NTK^{(\blocki+1)}_{(\sa_1;t_1;i_1)(\sa_2;t_2;i_2)}\, \label{eq:NTK-forward}\\
=&\sum_{\mu\in(\blocki+1)\text{-th block}}\LambdaGmu\frac{\partial \mathcal{R}^{(\blocki+1)}_{\sa_1;t_1;i_1}}{\partial \theta_{\mu}^{(\blocki+1)}}\frac{\partial \mathcal{R}^{(\blocki+1)}_{\sa_2;t_2;i_2}}{\partial \theta_{\mu}^{(\blocki+1)}}\, \nonumber\\
&+\sum_{t'_1,t'_2=1}^{T}\sum_{j_1,j_2=1}^{\hid}\sum_{k_1,k_2=1}^{\hid}\le(\frac{\partial \mathcal{R}^{(\blocki+1)}_{\sa_1;t_1;i_1}}{\partial s^{(\blocki)}_{\sa_1;t'_1;j_1}}\frac{\partial \mathcal{R}^{(\blocki+1)}_{\sa_2;t_2;i_2}}{\partial s^{(\blocki)}_{\sa_2;t'_2;j_2}}\ri)\le(\frac{\partial s^{(\blocki)}_{\sa_1;t'_1;j_1}}{\partial z^{(\blocki)}_{\sa_1;t'_1;k_1}}\frac{\partial s^{(\blocki)}_{\sa_2;t'_2;j_2}}{\partial z^{(\blocki)}_{\sa_2;t'_2;k_2}}\ri)\NTK^{(\blocki)}_{(\sa_1;t'_1;k_1)(\sa_2;t'_2;k_2)}\, \nonumber\\
&+\NTK^{(\blocki)}_{(\sa_1;t_1;i_1)(\sa_2;t_2;i_2)}\, \nonumber\\
&+\sum_{t'_1=1}^{T}\sum_{j_1=1}^{\hid}\sum_{k_1=1}^{\hid}\le(\frac{\partial \mathcal{R}^{(\blocki+1)}_{\sa_1;t_1;i_1}}{\partial s^{(\blocki)}_{\sa_1;t'_1;j_1}}\ri)\le(\frac{\partial s^{(\blocki)}_{\sa_1;t'_1;j_1}}{\partial z^{(\blocki)}_{\sa_1;t'_1;k_1}}\ri)\NTK^{(\blocki)}_{(\sa_1;t'_1;k_1)(\sa_2;t_2;i_2)}\, \nonumber\\
&+\sum_{t'_2=1}^{T}\sum_{j_2=1}^{\hid}\sum_{k_2=1}^{\hid}\le(\frac{\partial \mathcal{R}^{(\blocki+1)}_{\sa_2;t_2;i_2}}{\partial s^{(\blocki)}_{\sa_2;t'_2;j_2}}\ri)\le(\frac{\partial s^{(\blocki)}_{\sa_2;t'_2;j_2}}{\partial z^{(\blocki)}_{\sa_2;t'_2;k_2}}\ri)\NTK^{(\blocki)}_{(\sa_1;t_1;i_1)(\sa_2;t'_2;k_2)}\, .\nonumber
\end{align}
In words, the first term is the additive contribution from the $(\blocki+1)$-th-block model parameters, the second term is the cumulative contribution from the residual path, the third term is the xeroxed term from the skip path, and the last two terms are the cross terms between the skip and residual paths. (For $\blocki+1=L$, the last three terms are absent as there is no skip path.)
In more words, within the cumulative term, the multiplicative factor in the first set of the parentheses describes how the gradient changes as signals pass through the $(\blocki+1)$-th block while the multiplicative factor in the second set of the parentheses describes how the normalization layer affects the gradient.\footnote{These multiplicative factors are identified as  the partial Jacobians in Ref.~\cite{doshi2021critical} -- here squared and decomposed.}

Going forward -- block by block -- our strategy will be to \textit{(i)} choose the relative learning-rate factor for SGD, $\LambdaG$, so that the first additive term is of order one, \textit{(ii)} estimate the width scaling of the expected per-parameter gradient magnitude $\vert g^{\mu}\vert$ for model parameters in the group $G$, \textit{(iii)} adjust the relative learning-rate factor for AdamW as $\LambdaGAmu\sim\LambdaGmu \vert g^{\mu}\vert$, and further \textit{(iv)} recursively show that our choices of the initialization hyperparameters~\eqref{eq:init-preview-stem}--\eqref{eq:init-preview-head}  lead to order-one multiplicative factors in the second cumulative term while cross terms vanish in expectation.

For SGD, such neural-tangent analysis of gradients will lead to  -- much like our analysis of preactivations led to the scalings~\eqref{eq:init-preview-stem}--\eqref{eq:init-preview-head} of the initialization hyperparameters -- the following set of the scalings for the relative learning-rate factors:
\begin{align}
\LambdaSuper_{\patch}=&\le(\frac{1}{\npatch}\ri)\lambdaSub_{\patch}\, \ \ \Big\vert\Big\vert\ \ \LambdaSuper_{\WE}=\le(1\ri)\cdot\lambdaSub_{\text{WE}}\, , \label{eq:SGDlambdaG-preview-stem}\\
\LambdaSuper_{\PE}=&\le(1\ri)\cdot\lambdaSub_{\text{PE}}\, ,\label{eq:SGDlambdaG-preview-PE}\\
\LambdaSuper_{Q}=&\le(\frac{1}{\hid}\ri)\lambdaSub_{Q}\, ,\label{eq:SGDlambdaG-preview-Q}\\
\LambdaSuper_{K}=&\le(\frac{1}{\hid}\ri)\lambdaSub_{K}\, ,\label{eq:SGDlambdaG-preview-K}\\
\LambdaSuper_{V}=&\le(\frac{1}{\hid}\ri)\lambdaSub_{V}\, ,\label{eq:SGDlambdaG-preview-V}\\
\LambdaSuper_{U}=&\le(\frac{1}{\hid}\ri)\lambdaSub_{U}\, ,\label{eq:SGDlambdaG-preview-U}\\
\LambdaSuper_{W}=&\le(\frac{1}{\hid}\ri)\lambdaSub_{W}\, ,\label{eq:SGDlambdaG-preview-W}\\
\LambdaSuper_{X}=&\le(\frac{1}{M\hid}\ri)\lambdaSub_{X}\, ,\label{eq:SGDlambdaG-preview-X}\\
\LambdaSuper_{\head}=&\le(\frac{1}{\hid}\ri)\lambdaSub_{\head}\, ,\ \  \LambdaSuper_{\headb}=\le(1\ri)\cdot\lambdaSub_{\headb}\, ,\label{eq:SGDlambdaG-preview-head}
\end{align}
where -- like the order-one initialization hyperparameters $C_{G}$'s -- $\lambdaSub_{G}$'s are order-one training hyperparameters which we could in principle tune but won't.\footnote{Had we employed the $\lambda_{\mu\nu}$ notation of Refs.~\cite{PDLT,yaida2022meta} instead of the $\LambdaG$ notation used herein, we would have denoted, e.g., the first equation~\eqref{eq:SGDlambdaG-preview-stem} as $\lambda_{\patch_{i_1j_1}\patch_{i_2j_2}}=\LambdaSuper_{\patch}\delta_{i_1i_2}\delta_{j_1j_2}=(\lambdaSub_{\patch}/\npatch)\delta_{i_1i_2}\delta_{j_1j_2}$. Slightly more confusing in translation is that order-one training hyperparameters such as $\lambda_{W}^{(\blocki)}$ in Ref.~\cite{PDLT} would be denoted as $\lambdaSub_{W^{(\blocki)}}$ herein.}
For AdamW,  accounting for the factor of $\sim\vert g^{\mu}\vert$ modifies the above relative learning-rate factors to
\begin{align}
\LambdaSuperA_{\patch}=&\le(\frac{1}{\npatch\sqrt{\hid}}\ri)\lambdaSubA_{\patch}\, \ \ \Big\vert\Big\vert\ \ \LambdaSuperA_{\WE}=\le(\frac{1}{\sqrt{\hid}}\ri)\lambdaSubA_{\WE}\, , \label{eq:AdamWlambdaG-preview-stem}\\
\LambdaSuperA_{\PE}=&\le(\frac{1}{\sqrt{\hid}}\ri)\lambdaSubA_{\PE}\, ,\label{eq:AdamWlambdaG-preview-PE}\\
\LambdaSuperA_{Q}=&\le(\frac{1}{\hid\sqrt{\hid}}\ri)\lambdaSubA_{Q}\, ,\label{eq:AdamWlambdaG-preview-Q}\\
\LambdaSuperA_{K}=&\le(\frac{1}{\hid\sqrt{\hid}}\ri)\lambdaSubA_{K}\, ,\label{eq:AdamWlambdaG-preview-K}\\
\LambdaSuperA_{V}=&\le(\frac{1}{\hid\sqrt{\hid}}\ri)\lambdaSubA_{V}\, ,\label{eq:AdamWlambdaG-preview-V}\\
\LambdaSuperA_{U}=&\le(\frac{1}{\hid\sqrt{\hid}}\ri)\lambdaSubA_{U}\, ,\label{eq:AdamWlambdaG-preview-U}\\
\LambdaSuperA_{W}=&\le(\frac{1}{\hid\sqrt{M\hid}}\ri)\lambdaSubA_{W}\, ,\label{eq:AdamWlambdaG-preview-W}\\
\LambdaSuperA_{X}=&\le(\frac{1}{M\hid\sqrt{\hid}}\ri)\lambdaSubA_{X}\, ,\label{eq:AdamWlambdaG-preview-X}\\
\LambdaSuperA_{\head}=&\le(\frac{1}{\hid\sqrt{n_{\text{out}}}}\ri)\lambdaSubA_{\head}\, ,\ \ \LambdaSuperA_{\headb}=\le(\frac{1}{\sqrt{n_{\text{out}}}}\ri)\lambdaSubA_{\headb}\, ,\label{eq:AdamWlambdaG-preview-head}
\end{align}
so as to ensure that the network update stays of order one for AdamW.\footnote{Let's harken back to footnote~\ref{foot:Meta1} and discuss generic meta-principled scaling strategies~\cite{yaida2022meta}.\label{foot:Meta2} For learning rates, there is gauge redundancy~\cite{yang2021tensor,yaida2022meta}, but one way to implement it is, for SGD, to multiply the right-hand sides of~\eqref{eq:SGDlambdaG-preview-stem}--\eqref{eq:SGDlambdaG-preview-X} by $n^{s}$ \textit{except} the head one~\eqref{eq:SGDlambdaG-preview-head} and, for AdamW, to multiply the right-hand sides of\eqref{eq:AdamWlambdaG-preview-stem}--\eqref{eq:AdamWlambdaG-preview-X} by $n^{\frac{s}{2}}$ \textit{except} the head one~\eqref{eq:AdamWlambdaG-preview-head}. Here the factor of $n^{\frac{s}{2}}$ difference between SGD and AdamW arises from non-head gradient magnitude $\vert g^{\mu}\vert$'s picking up a factor of $\sqrt{1/n^{s}}$ from the head initialization scaling~\eqref{eq:init-preview-Meta-head}.}
With these choices, for the rest of this section, we'll recursively show that the means of the neural tangent kernels at initialization take the form
\be
\E{\NTK^{(\blocki)}_{(\sa_1;t_1;i_1)(\sa_2;t_2;i_2)}}=\delta_{i_1i_2} \Theta^{(\blocki)}_{(\sa_1;t_1)(\sa_2;t_2)}\, ,\label{eq:backward-goal}
\ee
with order-one kernels $\Theta^{(\blocki)}_{(\sa_1;t_1)(\sa_2;t_2)}$.

With those goals in mind, let's proceed forward with our effective-theory analysis of gradients, block by block.

\subsubsection{Stem Block}\label{subsubsec:backward-stem}
To stem the recursive analysis, we analyze the first-block neural tangent kernel,
\be
\NTK^{(1)}_{(\sa_1;t_1;i_1)(\sa_2;t_2;i_2)}=\sum_{\mu\in \text{1st block}}\LambdaGmu\frac{\partial z^{(1)}_{\sa_1;t_1;i_1}}{\partial \theta_{\mu}^{(1)}}\frac{\partial z^{(1)}_{\sa_2;t_2;i_2}}{\partial \theta_{\mu}^{(1)}}\, ,\label{eq:NTK-stem}
\ee
first for Vision Transformers and then for Language Transformers.\\

\begin{center}\textit{Vision: patchify embedding and positional embedding}\end{center}

Taking derivatives of the stem-block preactivations
\be
z^{(1)}_{\sa;t;i}=\PE_{t;i}+\sum_{j=1}^{\npatch}\patch_{ij}x_{\sa;t;j}
\ee
with respect to the stem-block model parameters, we have
\begin{align}
\frac{\partial z^{(1)}_{\sa;t;i}}{\partial \patch_{jk}}=&\delta_{ij}x_{\sa;t;k}\, ,\\
\frac{\partial z^{(1)}_{\sa;t;i}}{\partial \PE_{t';j}}=&\delta_{tt'}\delta_{ij}\, .
\end{align}

The squared norms of these gradients are then given by
\begin{align}
\sum_{i=1}^{\hid}\sum_{j=1}^{\npatch}\frac{\partial z^{(1)}_{\sa_1;t_1;i_1}}{\partial \patch_{ij}}\frac{\partial z^{(1)}_{\sa_2;t_2;i_2}}{\partial \patch_{ij}}=&\delta_{i_1i_2}\npatch\le(\frac{1}{\npatch}\sum_{j=1}^{\npatch}x_{\sa_1;t_1;j}x_{\sa_2;t_2;j}\ri)\, \label{eq:NTK-vision-stem1}\\
=&\delta_{i_1i_2}\le[\le(\npatch\ri)\cdot G^{(0)}_{(\sa_1;t_1)(\sa_2;t_2)}\ri]\, ,\nonumber\\
\sum_{t=1}^{T}\sum_{i=1}^{\hid}\frac{\partial z^{(1)}_{\sa_1;t_1;i_1}}{\partial \PE_{t;i}}\frac{\partial z^{(1)}_{\sa_2;t_2;i_2}}{\partial \PE_{t;i}}=&\delta_{i_1i_2}\le[\le(1\ri)\cdot\delta_{t_1t_2}\ri]\, ,\label{eq:NTK-vision-stem2}
\end{align}
where in the middle we recalled the definition of the order-one input kernel~\eqref{eq:input-kernel-vision}.\footnote{In passing, we note that these relations hold even without taking expectation values, i.e., the first-block neural tangent kernel is deterministic.}
Note that, as promised~\eqref{eq:backward-goal}, these additive contributions are diagonal in the embedding direction.

For SGD, the additive contributions \eqref{eq:NTK-vision-stem1} and \eqref{eq:NTK-vision-stem2} from these  gradients to the stem-block neural tangent kernel~\eqref{eq:NTK-stem} are multiplied by $\LambdaSuper_{\patch}$ and $\LambdaSuper_{\PE}$, respectively. Thus, in order to make these contributions stay of order one, we need to scale relative learning-rate factors as 
\begin{align}
\LambdaSuper_{\patch}=\le(\frac{1}{\npatch}\ri)\lambdaSub_{\patch}\, ,\ \ \ \LambdaSuper_{\PE}=\le(1\ri)\cdot\lambdaSub_{\PE}\, \ \ \ \text{for}\ \ \ \text{SGD}\, ,
\end{align}
with order-one hyperparameters $\lambdaSub_{\patch}$ and $\lambdaSub_{\PE}$.
For AdamW, we need to account for the magnitude of the per-parameter gradient $\vert g^{\mu}\vert$. To estimate it for the patchify weights, we note that there are $\hid\cdot\npatch$ components adding up to the squared norm of the order $\npatch$, so each component of the gradient on average scales as $\vert g^{\mu}\vert\sim\sqrt{\npatch/(\hid\cdot\npatch)}=1/\sqrt{\hid}$. Similarly for the positional-embedding parameters, we estimate $\vert g^{\mu}\vert\sim\sqrt{1/\hid}=1/\sqrt{\hid}$.\footnote{In this note, we implicitly focus on the scalings of hyperparameters with widths and do \textit{not} investigate the scalings with the batch size $\vert\mathcal{A}_t\vert$, sequence length $T$, etc.} Thus, to offset these factors in the AdamW optimizer, we set
\begin{align}
\LambdaSuperA_{\patch}=\le(\frac{1}{\npatch\sqrt{\hid}}\ri)\lambdaSubA_{\patch}\, ,\ \ \ \LambdaSuperA_{\PE}=\le(\frac{1}{\sqrt{\hid}}\ri)\lambdaSubA_{\PE}\, \ \ \ \text{for}\ \ \ \text{AdamW}\, ,
\end{align}
with order-one hyperparameters $\lambdaSubA_{\patch}$ and $\lambdaSubA_{\PE}$.\\
\begin{center}\textit{Language: word embedding and positional embedding}\end{center}

Very similarly to the vision case, taking derivatives of the stem-block preactivations
\be
z^{(1)}_{\sa;t;i}=\PE_{t;i}+\sum_{j=1}^{\nvocab}\WE_{ij}x_{\sa;t;j}\, ,
\ee
with respect to the stem-block model parameters, we have
\begin{align}
\frac{\partial z^{(1)}_{\sa;t;i}}{\partial \WE_{jk}}\Bigg\vert_{\text{stem}}=&\delta_{ij}x_{\sa;t;k}\, ,\label{eq:gradient-WE-stem}\\
\frac{\partial z^{(1)}_{\sa;t;i}}{\partial \PE_{t';j}}=&\delta_{tt'}\delta_{ij}\, .
\end{align}
Here, for the derivative with respect to the word-embedding parameters, we emphasized that these gradients come from the stem block because there will be another contribution from the head block, to be discussed in \S\ref{subsubsec:backward-head}.

Anyhow, the squared norms of these gradients are then given by
\begin{align}
\sum_{i=1}^{\hid}\sum_{j=1}^{\nvocab}\frac{\partial z^{(1)}_{\sa_1;t_1;i_1}}{\partial \WE_{ij}}\Bigg\vert_{\text{stem}}\frac{\partial z^{(1)}_{\sa_2;t_2;i_2}}{\partial \WE_{ij}}\Bigg\vert_{\text{stem}}=&\delta_{i_1i_2}\le(\sum_{j=1}^{\nvocab}x_{\sa_1;t_1;j}x_{\sa_2;t_2;j}\ri)\, \label{eq:LaT-NTK-stem-WE}\\
=&\delta_{i_1i_2}\le[\le(1\ri)\cdot G^{(0)}_{(\sa_1;t_1)(\sa_2;t_2)}\ri]\, ,\nonumber\\
\sum_{t=1}^{T}\sum_{i=1}^{\hid}\frac{\partial z^{(1)}_{\sa_1;t_1;i_1}}{\partial \PE_{t;i}}\frac{\partial z^{(1)}_{\sa_2;t_2;i_2}}{\partial \PE_{t;i}}=&\delta_{i_1i_2}\le[\le(1\ri)\cdot\delta_{t_1t_2}\ri]\, ,
\end{align}
where in the middle we recalled the definition of the order-one input kernel~\eqref{eq:input-kernel-language}. Note that, as promised~\eqref{eq:backward-goal}, these additive contributions are diagonal in the embedding direction.

For SGD, to ensure order-one additive contributions to the stem-block neural tangent kernel~\eqref{eq:NTK-stem}, we scale the relative learning-rate factors as
\begin{align}
\LambdaSuper_{\WE}=\le(1\ri)\cdot\lambdaSub_{\WE}\, ,\ \ \ \LambdaSuper_{\PE}=\le(1\ri)\cdot\lambdaSub_{\PE}\, \ \ \ \text{for}\ \ \ \text{SGD}\, ,\label{eq:WE-scale-stem-SGD}
\end{align}
i.e., we keep both of them of order one.
For AdamW, we adjust for the magnitude of the per-parameter gradient $\vert g^{\mu}\vert\sim\sqrt{1/\hid}=1/\sqrt{\hid}$ both for word-embedding and positional-embedding parameters, yielding
\begin{align}
\LambdaSuperA_{\WE}=\le(\frac{1}{\sqrt{\hid}}\ri)\lambdaSubA_{\WE}\, ,\ \ \ \LambdaSuperA_{\PE}=\le(\frac{1}{\sqrt{\hid}}\ri)\lambdaSubA_{\PE}\, \ \ \ \text{for}\ \ \ \text{AdamW}\, .\label{eq:WE-scale-stem-AdamW}
\end{align}
Here, in estimating the gradient magnitude for each component of the word-embedding parameters, we didn't divide by a factor of $\nvocab$, for a reason: unlike in the vision case where gradients are spread across all the $\npatch$ input components, here in the language case the gradients are mostly zero due to the one-hot structure of inputs, i.e., $x_{\sa;t;j}=\delta_{j j_{\star}(\sa;t)}$.\footnote{This reasoning, however, becomes more dubious as the batch size $\vert \mathcal{A}_t\vert$ grows larger: for sufficiently large batch sizes, all the gradients will in principle become nonzero, albeit inhomogeneously due to non-uniform token distributions in typical vocabularies~\cite{zipf2016human} (which incidentally forbids us from naively scaling $\LambdaSuperA_{\WE}$ with a simple power of $\nvocab$).\label{foot:nvocab-stem}
There is another subtlety that arises in the head block: see footnote~\ref{foot:nvocab-head}.\\
Overall, we leave the analysis of the interplay between the batch size $\vert\mathcal{A}_t\vert$, sequence length $T$, vocabulary size $\nvocab$, and token distribution to future work.}

\subsubsection{Layer Normalization}\label{subsubsec:backward-LN}
Mirroring \S\ref{subsubsec:initialization-LN}, we'll temporarily drop the sample--token--block indices until the dust settles.

To analyze how normalization layers affect the gradients, looking back at the forward equation for the neural tangent kernels~\eqref{eq:NTK-forward}, it would appear that we need to analyze the multiplicative factor
\be
\frac{\partial s_{j_1}}{\partial z_{k_1}}\frac{\partial s_{j_2}}{\partial z_{k_2}}\, ,
\ee
for general combinations of the embedding indices $(j_1,j_2,k_1,k_2)$. Fortunately, because of the recursively-expected  diagonal form~\eqref{eq:backward-goal} of the neural tangent kernel in the embedding direction, we can focus on the ones with $k_1=k_2$. Further we'll recursively show  in the following \S\S\ref{subsubsec:backward-MHSA}--\ref{subsubsec:backward-head} [in particular equations~\eqref{eq:promised-MHSA-1}, \eqref{eq:promised-MHSA-2}, \eqref{eq:promised-MLP}, and \eqref{eq:promised-head}] that the (putatively) leading-order contributions always come in the forms of the following two particular linear combinations:
\be
\frac{1}{\hid}\sum_{j=1}^{\hid}\sum_{k=1}^{\hid}\frac{\partial s_{j}}{\partial z_{k}}\frac{\partial s_{j}}{\partial z_{k}}\, ,\label{eq:LN-NTK-baby-1}
\ee
and
\be
\frac{1}{\hid^2}\sum_{j_1,j_2=1}^{\hid}\sum_{k=1}^{\hid}s_{j_1}s_{j_2}\frac{\partial s_{j_1}}{\partial z_{k}}\frac{\partial s_{j_2}}{\partial z_{k}}\, .\label{eq:LN-NTK-baby-2}
\ee

Let's start with the first one~\eqref{eq:LN-NTK-baby-1}.  Mindlessly rolling it out, we get
\begin{align}
&\frac{1}{\hid}\sum_{j=1}^{\hid}\sum_{k=1}^{\hid}\frac{\partial s_{j}}{\partial z_{k}}\frac{\partial s_{j}}{\partial z_{k}}\, \\
=&\frac{1}{\hid}\sum_{j=1}^{\hid}\sum_{k=1}^{\hid} \le\{\frac{\partial }{\partial z_k}\le[\frac{z_j-\frac{1}{\hid}\sum_{i=1}^{\hid}z_i}{\sqrt{\frac{1}{\hid}\sum_{i=1}^{\hid}z_i^2-\le(\frac{1}{\hid}\sum_{i=1}^{\hid}z_i\ri)^2+\epsilon}}\ri]\ri\}^2\, \nonumber\\
=&\frac{1}{\hid}\sum_{j=1}^{\hid}\sum_{k=1}^{\hid} \le[\frac{1}{\sqrt{\frac{1}{\hid}\sum_{i=1}^{\hid}z_i^2-\le(\frac{1}{\hid}\sum_{i=1}^{\hid}z_i\ri)^2+\epsilon}}  \le(\delta_{jk}-\frac{1}{\hid}-\frac{1}{\hid}s_js_k\ri)\ri]^2\, \nonumber\\
=&\frac{1}{\le[\sqrt{\frac{1}{\hid}\sum_{i=1}^{\hid}z_i^2-\le(\frac{1}{\hid}\sum_{i=1}^{\hid}z_i\ri)^2+\epsilon}\ri]^2}\!\!  \le[1-\frac{1}{\hid}-\frac{2}{\hid}\le(\frac{1}{\hid}\sum_{j=1}^{\hid}s_j^2\ri)\! +\! \frac{2}{\hid}\le(\frac{1}{\hid}\sum_{j=1}^{\hid}s_j\ri)^2\!\!\! +\! \frac{1}{\hid}\le(\frac{1}{\hid}\sum_{j=1}^{\hid}s_j^2\ri)^2\ri]\, .\nonumber
\end{align}
Recalling from \S\ref{subsubsec:forward-LN} that $\frac{1}{\hid}\sum_{i=1}^{\hid}z_i^2=G+\Delta G$ and $\le(\frac{1}{\hid}\sum_{i=1}^{\hid}z_i\ri)^2=\nabla G$ and that $\Delta G$ and $\nabla G$ are ignorable at leading order, we get 
\be
\frac{1}{\hid}\sum_{j=1}^{n}\sum_{k=1}^{n}\frac{\partial s_{j}}{\partial z_{k}}\frac{\partial s_{j}}{\partial z_{k}}=\frac{1}{\le(\sqrt{G+\epsilon}\ri)^2} +O\le(\frac{1}{\hid}\ri)\, .
\ee
Next, mindlessly rolling out the second one~\eqref{eq:LN-NTK-baby-2}, we get
\begin{align}
&\frac{1}{\hid^2}\sum_{j_1,j_2=1}^{n}\sum_{k=1}^{\hid}s_{j_1}s_{j_2}\frac{\partial s_{j_1}}{\partial z_{k}}\frac{\partial s_{j_2}}{\partial z_{k}}\, \\
=&\sum_{k=1}^{\hid} \le\{\frac{1}{\hid}\sum_{j=1}^{\hid}s_j\frac{\partial }{\partial z_k}\le[\frac{z_j-\frac{1}{\hid}\sum_{i=1}^{\hid}z_i}{\sqrt{\frac{1}{\hid}\sum_{i=1}^{\hid}z_i^2-\le(\frac{1}{\hid}\sum_{i=1}^{\hid}z_i\ri)^2+\epsilon}}\ri]\ri\}^2\, \nonumber\\
=&\sum_{k=1}^{\hid} \le\{\frac{1}{\sqrt{\frac{1}{\hid}\sum_{i=1}^{\hid}z_i^2-\le(\frac{1}{\hid}\sum_{i=1}^{\hid}z_i\ri)^2+\epsilon}}  \le[\frac{s_k}{\hid}-\frac{1}{\hid}\le(\frac{1}{\hid}\sum_{j=1}^{\hid}s_j\ri)-\frac{1}{\hid}\le(\frac{1}{\hid}\sum_{j=1}^{\hid}s_j^2\ri)s_k\ri]\ri\}^2\, \nonumber\\
=&O\le(\frac{1}{\hid}\ri)\, .\nonumber
\end{align}
That is, we can neglect this type of contributions at the leading order.

As the dust has settled,  carefully putting all the indices back in, we get
\be
\frac{1}{\hid}\sum_{j=1}^{n}\sum_{k=1}^{n}\frac{\partial s^{(\blocki)}_{\sa_1;t_1;j}}{\partial z^{(\blocki)}_{\sa_1;t_1;k}}\frac{\partial s^{(\blocki)}_{\sa_2;t_2;j}}{\partial z^{(\blocki)}_{\sa_2;t_2;k}}=\frac{1}{\sqrt{G^{(\blocki)}_{(\sa_1;t_1)(\sa_1;t_1)}+\epsilon}\sqrt{G^{(\blocki)}_{(\sa_2;t_2)(\sa_2;t_2)}+\epsilon}}+O\le(\frac{1}{\hid}\ri)\, .
\ee
Note that, since we expect the kernel $G^{(\blocki)}$ to grow linearly with the depth $\blocki$, we expect this multiplicative factor to be linearly suppressed~\cite{doshi2021critical}.

Overall, in retrospect and in prospect, we can simply use the quick intuitive mnemonics
\be
s_{\sa;t;j}\approx\frac{z_{\sa;t;j}}{\sqrt{G_{(\sa;t)(\sa;t)}+\epsilon}} \ \ \ \text{and}\ \ \ \frac{\partial s_{\sa;t;j}}{\partial z_{\sa;t;k}}\approx \frac{1}{\sqrt{G_{(\sa;t)(\sa;t)}+\epsilon}}\delta_{jk}\, ,\label{eq:LN-mnemonic}
\ee
from which all of our results here in~\ref{subsubsec:backward-LN} and there in~\ref{subsubsec:initialization-LN}  follow immediately.

\subsubsection{Multi-Head Self-Attention Block}\label{subsubsec:backward-MHSA}
Taking derivatives of the residual MHSA-block outputs
\be
r_{\sa;t;i}=\sum_{h=1}^{H}\sum_{t'=1}^{T}\sum_{c=1}^{C}\sum_{j=1}^{\hid} \ws^{h}_{\sa;tt'}\le[\wst\le(Q,K;s\ri)\ri]U_{ic}^{h}V_{cj}^{h}s_{\sa;t';j}\, 
\ee
with respect to the MHSA-block model parameters and the incoming signals, we have
\begin{align}
\frac{\partial r_{\sa;t;i}}{\partial U_{jc}^{h}}=&\delta_{ij}\sum_{t'=1}^{T}\sum_{k=1}^{\hid} V_{ck}^{h}\ws^{h}_{\sa;tt'}s_{\sa;t';k}\, ,\\
\frac{\partial r_{\sa;t;i}}{\partial V_{cj}^{h}}=&U_{ic}^{h}\sum_{t'=1}^{T} \ws^{h}_{\sa;tt'}s_{\sa;t';j}\, ,\\
\frac{\partial r_{\sa;t;i}}{\partial K_{cj}^{h}}=&\sum_{t'=1}^{T}\sum_{c'=1}^{C}\sum_{k=1}^{\hid}U_{ic'}^{h}V_{c'k}^{h} \le(\sum_{\tilde{t},\tilde{t}'=1}^T\frac{\partial \ws^{h}_{\sa;tt'}}{\partial \wst^{h}_{\sa;\tilde{t}\tilde{t}'}}\frac{\partial \wst^{h}_{\sa;\tilde{t}\tilde{t}'}}{\partial K_{cj}^{h}}\ri)s_{\sa;t';k}\, ,\\
\frac{\partial r_{\sa;t;i}}{\partial Q_{cj}^{h}}=&\sum_{t'=1}^{T}\sum_{c'=1}^{C}\sum_{k=1}^{\hid}U_{ic'}^{h}V_{c'k}^{h} \le(\sum_{\tilde{t},\tilde{t}'=1}^T\frac{\partial \ws^{h}_{\sa;tt'}}{\partial \wst^{h}_{\sa;\tilde{t}\tilde{t}'}}\frac{\partial \wst^{h}_{\sa;\tilde{t}\tilde{t}'}}{\partial Q_{cj}^{h}}\ri)s_{\sa;t';k}\, ,\\
\frac{\partial r_{\sa;t;i}}{\partial s_{\sa;t'';k}}=&\sum_{t'=1}^{T}\sum_{h=1}^{H}\sum_{c=1}^{C}U_{ic}^{h}\sum_{j=1}^{\hid}V_{cj}^{h} \le[\delta_{jk}\ws^{h}_{\sa;tt''}+\le(\sum_{\tilde{t},\tilde{t}'=1}^T\frac{\partial \ws^{h}_{\sa;tt'}}{\partial \wst^{h}_{\sa;\tilde{t}\tilde{t}'}}\frac{\partial \wst^{h}_{\sa;\tilde{t}\tilde{t}'}}{\partial s_{\sa;t'';k}}\ri)s_{\sa;t';j}\ri]\, , \label{eq:dangling-MHSA}
\end{align}
where for the last three we used the chain rule to roll out the derivatives of the self-attention matrix $\ws^{h}_{\sa;tt'}$. A little bit more explicitly, the derivatives of the query--key dot product~\eqref{eq:matrix-product-1},
\be
\wst_{\sa;\tilde{t}\tilde{t}'}^{h}=\frac{1}{\sqrt{C}}\sum_{c=1}^{C}q_{\sa;\tilde{t};c}^{h}k_{\sa;\tilde{t}';c}^{h}=\frac{1}{\sqrt{C}}\sum_{c=1}^{C}\sum_{i_1,i_2=1}^{\hid}Q_{ci_1}^{h}K_{ci_2}^{h}s_{\sa;\tilde{t};i_1}s_{\sa;\tilde{t}';i_2}\, ,
\ee
are given by
\begin{align}
\frac{\partial \wst^{h}_{\sa;\tilde{t}\tilde{t}'}}{\partial K_{cj}^{h}}=&\frac{1}{\sqrt{C}}\sum_{m=1}^{\hid}Q_{cm}^{h}s_{\sa;\tilde{t};m}s_{\sa;\tilde{t}';j}=\frac{1}{\sqrt{C}}q_{\sa;\tilde{t};c}^{h}s_{\sa;\tilde{t}';j}\, ,\\
\frac{\partial \wst^{h}_{\sa;\tilde{t}\tilde{t}'}}{\partial Q_{cj}^{h}}=&\frac{1}{\sqrt{C}}\sum_{m=1}^{\hid}K_{cm}^{h}s_{\sa;\tilde{t};j}s_{\sa;\tilde{t}';m}=\frac{1}{\sqrt{C}}s_{\sa;\tilde{t};j}k_{\sa;\tilde{t'};c}^{h}\, ,\\
\frac{\partial \wst^{h}_{\sa;\tilde{t}\tilde{t}'}}{\partial s_{\sa;t'';k}}=&\frac{1}{\sqrt{C}}\sum_{c=1}^{C}\sum_{m=1}^{\hid}\le(\delta_{t'' \tilde{t}}Q_{ck}^{h}K_{cm}^{h}s_{\sa;\tilde{t}';m}+\delta_{t'' \tilde{t}'}Q_{cm}^{h}K_{ck}^{h}s_{\sa;\tilde{t};m}\ri)\, \\
=&\frac{1}{\sqrt{C}}\sum_{c=1}^{C}\le(\delta_{t'' \tilde{t}}Q_{ck}^{h}k_{\sa;\tilde{t'};c}^{h}+\delta_{t'' \tilde{t}'}q_{\sa;\tilde{t};c}^{h}K_{ck}^{h}\ri)\, .\nonumber
\end{align}

Expectation values for the squared gradient norms are then given by -- recalling $n=HC$ --
\begin{align}
\label{eq:Uback}
&\!\!\!\!\!\!\!\!\!\!\E{\sum_{h=1}^H\sum_{c=1}^{C}\sum_{j=1}^{\hid}\frac{\partial r_{\sa_1;t_1;i_1}}{\partial U_{jc}^{h}}\frac{\partial r_{\sa_2;t_2;i_2}}{\partial U_{jc}^{h}}}\, \\
=&\delta_{i_1i_2}\le\{\le(\hid\ri)\cdot C_V\sum_{t'_1,t'_2=1}^{T}\E{\le(\frac{1}{H}\sum_{h=1}^{H}\ws^{h}_{\sa_1;t_1t'_1}\ws^{h}_{\sa_2;t_2t'_2}\ri)\le(\frac{1}{\hid}\sum_{j=1}^{\hid}s_{\sa_1;t'_1;j}s_{\sa_2;t'_2;j}\ri)}\ri\}\, ,\nonumber\\
\label{eq:Vback}
&\!\!\!\!\!\!\!\!\!\!\E{\sum_{h=1}^H\sum_{c=1}^{C}\sum_{j=1}^{\hid}\frac{\partial r_{\sa_1;t_1;i_1}}{\partial V_{cj}^{h}}\frac{\partial r_{\sa_2;t_2;i_2}}{\partial V_{cj}^{h}}}\, \\
=&\delta_{i_1i_2} \le\{\le(\hid\ri)\cdot  C_U\sum_{t'_1,t'_2=1}^{T} \E{\le(\frac{1}{H}\sum_{h=1}^{H}\ws^{h}_{\sa_1;t_1t'_1}\ws^{h}_{\sa_2;t_2t'_2}\ri)\le(\frac{1}{\hid}\sum_{j=1}^{\hid}s_{\sa_1;t'_1;j}s_{\sa_2;t'_2;j}\ri)}\ri\}\, ,\nonumber\\
\label{eq:Kback}
&\!\!\!\!\!\!\!\!\!\!\E{\sum_{h=1}^H\sum_{c=1}^{C}\sum_{j=1}^{\hid} \frac{\partial r_{\sa_1;t_1;i_1}}{\partial K_{cj}^{h}}\frac{\partial r_{\sa_2;t_2;i_2}}{\partial K_{cj}^{h}}}\, \\
=& \delta_{i_1i_2} \Bigg\{\le(\hid\ri)\cdot  C_U C_V\sum_{t'_1,t'_2,\tilde{t}_1,\tilde{t}_2,\tilde{t}'_1,\tilde{t}'_2=1}^{T}\mathbb{E}\Bigg[\Bigg(\frac{1}{H}\sum_{h=1}^{H}\frac{\partial \ws^{h}_{\sa_1;t_1t'_1}}{\partial \wst^{h}_{\sa_1;\tilde{t}_1\tilde{t}'_1}}\frac{\partial \ws^{h}_{\sa_2;t_2t'_2}}{\partial \wst^{h}_{\sa_2;\tilde{t}_2\tilde{t}'_2}}\Bigg)\le(\frac{1}{C}\sum_{c=1}^{C}q_{\sa_1;\tilde{t}_1;c}^{h}q_{\sa_2;\tilde{t}_2;c}^{h}\ri)\, \nonumber\\
&\ \ \ \ \ \ \ \ \ \ \ \ \ \ \ \ \ \ \ \ \ \ \ \ \ \ \ \ \ \ \ \ \ \ \ \ \ \ \ \ \ \ \ \ \ \ \times\le(\frac{1}{\hid}\sum_{j=1}^{\hid}s_{\sa_1;\tilde{t}'_1;j}s_{\sa_2;\tilde{t}'_2;j}\ri)\le(\frac{1}{\hid}\sum_{k=1}^{\hid}s_{\sa_1;t'_1;k}s_{\sa_2;t'_2;k}\ri)\Bigg]\Bigg\}\, ,\nonumber\\
\label{eq:Qback}
&\!\!\!\!\!\!\!\!\!\!\E{\sum_{h=1}^H\sum_{c=1}^{C}\sum_{j=1}^{\hid} \frac{\partial r_{\sa_1;t_1;i_1}}{\partial Q_{cj}^{h}}\frac{\partial r_{\sa_2;t_2;i_2}}{\partial Q_{cj}^{h}}}\, \\
=& \delta_{i_1i_2} \Bigg\{\le(\hid\ri)\cdot  C_U C_V\sum_{t'_1,t'_2,\tilde{t}_1,\tilde{t}_2,\tilde{t}'_1,\tilde{t}'_2=1}^{T}\mathbb{E}\Bigg[\Bigg(\frac{1}{H}\sum_{h=1}^{H}\frac{\partial \ws^{h}_{\sa_1;t_1t'_1}}{\partial \wst^{h}_{\sa_1;\tilde{t}_1\tilde{t}'_1}}\frac{\partial \ws^{h}_{\sa_2;t_2t'_2}}{\partial \wst^{h}_{\sa_2;\tilde{t}_2\tilde{t}'_2}}\Bigg)\le(\frac{1}{C}\sum_{c=1}^{C}k_{\sa_1;\tilde{t}'_1;c}^{h}k_{\sa_2;\tilde{t}'_2;c}^{h}\ri)\, \nonumber\\
&\ \ \ \ \ \ \ \ \ \ \ \ \ \ \ \ \ \ \ \ \ \ \ \ \ \ \ \ \ \ \ \ \ \ \ \ \ \ \ \ \ \ \ \ \ \ \times\le(\frac{1}{\hid}\sum_{j=1}^{\hid}s_{\sa_1;\tilde{t}_1;j}s_{\sa_2;\tilde{t}_2;j}\ri)\le(\frac{1}{\hid}\sum_{k=1}^{\hid}s_{\sa_1;t'_1;k}s_{\sa_2;t'_2;k}\ri)\Bigg]\Bigg\}\, .\nonumber
\end{align}
Here, the term in each pair of the parentheses in expectation is expected to be of order one and thus the squared gradient norms all scale as $\hid$.\footnote{In particular, for instance, we can use the results of Appendix~\ref{app:self-attention}  to factor the expectation as\label{foot:MHSA-stats-102}
\begin{align}
&\mathbb{E}\Bigg[\Bigg(\frac{1}{H}\sum_{h=1}^{H}\frac{\partial \ws^{h}_{\sa_1;t_1t'_1}}{\partial \wst^{h}_{\sa_1;\tilde{t}_1\tilde{t}'_1}}\frac{\partial \ws^{h}_{\sa_2;t_2t'_2}}{\partial \wst^{h}_{\sa_2;\tilde{t}_2\tilde{t}'_2}}\Bigg)\le(\frac{1}{C}\sum_{c=1}^{C}k_{\sa_1;\tilde{t}_1c}^{h}k_{\sa_2;\tilde{t}_2c}^{h}\ri)\le(\frac{1}{\hid}\sum_{j=1}^{\hid}s_{\sa_1;\tilde{t}'_1;j}s_{\sa_2;\tilde{t}'_2;j}\ri)\le(\frac{1}{\hid}\sum_{k=1}^{\hid}s_{\sa_1;t'_1;k}s_{\sa_2;t'_2;k}\ri)\Bigg]\, \nonumber\\
=&\mathbb{E}\Bigg[\Bigg(\frac{1}{H}\sum_{h=1}^{H}\frac{\partial \ws^{h}_{\sa_1;t_1t'_1}}{\partial \wst^{h}_{\sa_1;\tilde{t}_1\tilde{t}'_1}}\frac{\partial \ws^{h}_{\sa_2;t_2t'_2}}{\partial \wst^{h}_{\sa_2;\tilde{t}_2\tilde{t}'_2}}\Bigg)\Bigg]\E{\frac{1}{C}\sum_{c=1}^{C}k_{\sa_1;\tilde{t}_1c}^{h}k_{\sa_2;\tilde{t}_2c}^{h}}\kernelLN_{(\sa_1;\tilde{t}'_1)(\sa_2;\tilde{t}'_2)}\kernelLN_{(\sa_1;t'_1)(\sa_2;t'_2)}+O\le(\frac{1}{C}\ri)\, \label{eq:MHSA-stats-102}
\end{align}
where
\be
\E{\frac{1}{C}\sum_{c=1}^{C}k_{\sa_1;\tilde{t}_1c}^{h}k_{\sa_2;\tilde{t}_2c}^{h}}=\E{\frac{1}{C}\sum_{c=1}^{C}\sum_{i_1,i_2=1}^{\hid} K^{h}_{ci_1}K^{h}_{ci_2}s_{\sa_1;\tilde{t}_1;i_1}s_{\sa_2;\tilde{t}_2;i_2}}=C_{K} \kernelLN_{(\sa_1;\tilde{t}_1)(\sa_2;\tilde{t}_2)}\, ,
\ee
and further express the remaining expectation as a $\le(\vert\mathcal{D}\vert T^2\ri)$-dimensional Gaussian integral.}
Note that, as promised~\eqref{eq:backward-goal}, these additive contributions are diagonal in the embedding direction.

For SGD, to ensure order-one additive contributions to the neural tangent kernel, we thus should scale relative learning-rate factors $\LambdaG$ as 
\begin{align}
\LambdaSuper_{Q}=\le(\frac{1}{\hid}\ri)\lambdaSub_{Q}\, ,\ \ \ \LambdaSuper_{K}=\le(\frac{1}{\hid}\ri)\lambdaSub_{K}\, ,\ \ \ \LambdaSuper_{V}=\le(\frac{1}{\hid}\ri)\lambdaSub_{V}\, ,\ \ \ \LambdaSuper_{U}=\le(\frac{1}{\hid}\ri)\lambdaSub_{U}\, \ \ \ \text{for}\ \ \ \text{SGD}\, .
\end{align}
For AdamW, we need to adjust for the magnitude of the per-parameter gradient. Since all these weights have $\hid^2$ components adding up to the squared norm of order $\hid$, each component of the gradient scales as $\vert g^{\mu}\vert\sim\sqrt{n/\le(\hid^2\ri)}=1/\sqrt{\hid}$. Thus, to offset these factors in the AdamW optimizer, we set
\begin{align}
\LambdaSuperA_{Q}=\le(\frac{1}{\hid\sqrt{\hid}}\ri)\lambdaSubA_{Q}\, ,\ \ \LambdaSuperA_{K}=\le(\frac{1}{\hid\sqrt{\hid}}\ri)\lambdaSubA_{K}\, ,\ \ \LambdaSuperA_{V}=\le(\frac{1}{\hid\sqrt{\hid}}\ri)\lambdaSubA_{V}\, ,\ \ \LambdaSuperA_{U}=\le(\frac{1}{\hid\sqrt{\hid}}\ri)\lambdaSubA_{U}\, \ \ \text{for}\ \ \ \text{AdamW}\, .
\end{align}

Finally, moving onto the recursive contributions, the cross terms vanish as
\be
\E{\frac{\partial r_{\sa;t;i}}{\partial s_{\sa;t'';k}}\mathcal{F}^{(\text{previous})}}=0\, ,
\ee
due to the dangling mean-zero weights $U$ and $V$ in the signal derivative~\eqref{eq:dangling-MHSA}, where here $\mathcal{F}^{(\text{previous})}$ is a function of observables from the preceding blocks.
For the cumulative contribution, we have
 \begin{align}
&\E{\sum_{k_1,k_2=1}^{\hid}\frac{\partial r_{\sa_1;t_1;i_1}}{\partial s_{\sa_1;t''_1;k_1}}\frac{\partial r_{\sa_2;t_2;i_2}}{\partial s_{\sa_2;t''_2;k_2}}\mathcal{F}_{k_1k_2}^{(\text{previous})}}\, \\
=&\delta_{i_1i_2}C_U C_V\frac{1}{H}\sum_{h=1}^{H}\sum_{t'_1,t'_2=1}^{T}\frac{1}{\hid}\sum_{k_1,k_2=1}^{\hid}\, \nonumber\\
&\times\mathbb{E}\Bigg\{\Bigg[\delta_{k_1k_2}\ws^{h}_{\sa_1;t_1t''_1}\ws^{h}_{\sa_2;t_2t''_2}\, \nonumber\\
&\ \ \ \ \ \ \ \ \ \ +s_{\sa_2;t'_2;k_1}\ws^{h}_{\sa_1;t_1t''_1}\sum_{\tilde{t}_2,\tilde{t}'_2=1}^T\frac{\partial \ws^{h}_{\sa_2;t_2t'_2}}{\partial \wst^{h}_{\sa_2;\tilde{t}_2\tilde{t}'_2}}\frac{1}{\sqrt{C}}\sum_{c_2=1}^{C}\le(\delta_{t''_2 \tilde{t}_2}Q_{c_2 k_2}^{h}k^{h}_{\sa_2;\tilde{t}'_2;c_2}+\delta_{t''_2 \tilde{t}'_2}K_{c_2k_2}^{h}q^{h}_{\sa_2;\tilde{t}_2;c_2}\ri)\, \nonumber\\
&\ \ \ \ \ \ \ \ \ \ +s_{\sa_1;t'_1;k_2}\ws^{h}_{\sa_2;t_2t''_2}\sum_{\tilde{t}_1,\tilde{t}'_1=1}^T\frac{\partial \ws^{h}_{\sa_1;t_1t'_1}}{\partial \wst^{h}_{\sa_1;\tilde{t}_1\tilde{t}'_1}}\frac{1}{\sqrt{C}}\sum_{c_1=1}^{C}\le(\delta_{t''_1 \tilde{t}_1}Q_{c_1 k_1}^{h}k^{h}_{\sa_1;\tilde{t}'_1;c_1}+\delta_{t''_1 \tilde{t}'_1}K_{c_1k_1}^{h}q^{h}_{\sa_1;\tilde{t}_1;c_1}\ri)\, \nonumber\\
&\ \ \ \ \ \ \ \ \ \ +\hid\le(\frac{1}{\hid}\sum_{j=1}^{\hid}s_{\sa_1;t'_1;j}s_{\sa_2;t'_2;j}\ri)\sum_{\tilde{t}_1,\tilde{t}'_1,\tilde{t}_2,\tilde{t}'_2=1}^T\frac{\partial \ws^{h}_{\sa_1;t_1t'_1}}{\partial \wst^{h}_{\sa_1;\tilde{t}_1\tilde{t}'_1}}\frac{\partial \ws^{h}_{\sa_2;t_2t'_2}}{\partial \wst^{h}_{\sa_2;\tilde{t}_2\tilde{t}'_2}}\frac{1}{C}\sum_{c_1,c_2=1}^{C}\sum_{m_1,m_2=1}^{\hid}\, \nonumber\\
&\ \ \ \ \ \ \ \ \ \ \ \ \ \ \ \ \ \ \ \times\le(\delta_{t''_1 \tilde{t}_1}Q_{c_1 k_1}^{h}K_{c_1m_1}^{h}s_{\sa_1;\tilde{t}'_1;m_1}+\delta_{t''_1 \tilde{t}'_1}Q_{c_1m_1}^{h}K_{c_1k_1}^{h}s_{\sa_1;\tilde{t}_1;m_1}\ri)\, \nonumber\\
&\ \ \ \ \ \ \ \ \ \ \ \ \ \ \ \ \ \ \ \times\le(\delta_{t''_2 \tilde{t}_2}Q_{c_2 k_2}^{h}K_{c_2m_2}^{h}s_{\sa_2;\tilde{t}'_2;m_2}+\delta_{t''_2 \tilde{t}'_2}Q_{c_2m_2}^{h}K_{c_2k_2}^{h}s_{\sa_2;\tilde{t}_2;m_2}\ri)\Bigg]\mathcal{F}_{k_1k_2}^{(\text{previous})}\Bigg\}\, .\nonumber
\end{align}
Now there are four terms in expectation. The first term gives rise to the contribution
 \begin{align}
\delta_{i_1i_2}C_U C_V\frac{1}{H}\sum_{h=1}^{H}\sum_{t'_1,t'_2=1}^{T}\E{\ws^{h}_{\sa_1;t_1t''_1}\ws^{h}_{\sa_2;t_2t''_2}\frac{1}{\hid}\le(\sum_{k=1}^{\hid}\mathcal{F}_{kk}^{(\text{previous})}\ri)}\, ,\label{eq:promised-MHSA-1}
\end{align}
which has the first promised form~\eqref{eq:LN-NTK-baby-1} of the layer-normalization multiplicative factor and also is diagonal in the embedding direction~\eqref{eq:backward-goal}.
The next two cross terms actually vanish at this order.\footnote{The argument essentially boils down to doing algebra of the form
\begin{align}
&\frac{1}{\hid}\sum_{k_1,k_2=1}^{\hid}\E{\le(\wst^{h}_{\sa;tt'}\ri)s_{\alpha_2;t'_2;k_1}\frac{1}{\sqrt{C}}Q^{h}_{c_2k_2}k^{h}_{\sa_2;\tilde{t}'_2;c_2}\mathcal{F}_{k_1k_2}^{(\text{previous})}}\, \\
=&\frac{1}{\hid}\sum_{k_1,k_2=1}^{\hid}\frac{1}{C}\sum_{c_2=1}^{C} \E{\le(\sum_{c=1}^{C}\sum_{i_1,i_2=1}^{\hid}Q_{ci_1}^{h}K_{ci_2}^{h}s_{\sa;t;i_1}s_{\sa;t';i_2}\ri)s_{\alpha_2;t'_2;k_1}Q^{h}_{c_2k_2}\sum_{m_2=1}^{\hid}K^{h}_{c_2 m_2}s_{\sa_2;\tilde{t}'_2;m_2}\mathcal{F}_{k_1k_2}^{(\text{previous})}}\, \nonumber\\
=&C_Q C_K\frac{1}{\hid^3}\sum_{i_1,i_2,k_1,k_2,m_2=1}^{\hid} \E{\delta_{i_1k_2}\delta_{i_2 m_2}s_{\sa;t;i_1}s_{\sa;t';i_2}s_{\alpha_2;t'_2;k_1}s_{\sa_2;\tilde{t}'_2;m_2}\mathcal{F}_{k_1k_2}^{(\text{previous})}}\, \nonumber\\
=&C_Q C_K\E{\le(\frac{1}{\hid}\sum_{m_2=1}^{\hid}s_{\sa;t';m_2}s_{\sa_2;\tilde{t}'_2;m_2}\ri)\le(\frac{1}{\hid^2}\sum_{k_1,k_2=1}^{\hid} s_{\sa;t;k_2}s_{\alpha_2;t'_2;k_1}\mathcal{F}_{k_1k_2}^{(\text{previous})}\ri)}=O\le(\frac{1}{\hid}\ri)\, ,\nonumber
\end{align}
where in the very last equality we noticed that it results in the second promised form~\eqref{eq:LN-NTK-baby-2} of the layer-normalization multiplicative factor, which can be dropped as $O(1/\hid)$.
The rest of the argument is then similar to the one in Appendix~\ref{app:self-attention}, doing the same algebra with -- instead of just one query--key dot product $\wst^{h}_{\sa;tt'}$ -- any odd number of the query--key dot products.}
For the last term, integrating over $Q$ and $K$ weights, we get
 \begin{align}
&\delta_{i_1i_2}C_U C_V C_Q C_K\frac{1}{H}\sum_{h=1}^{H}\sum_{t'_1,t'_2,\tilde{t}_1,\tilde{t}'_1,\tilde{t}_2,\tilde{t}'_2=1}^{T}\, \label{eq:promised-MHSA-2}\\
&\times\mathbb{E}\Bigg[\frac{\partial \ws^{h}_{\sa_1;t_1t'_1}}{\partial \wst^{h}_{\sa_1;\tilde{t}_1\tilde{t}'_1}}\frac{\partial \ws^{h}_{\sa_2;t_2t'_2}}{\partial \wst^{h}_{\sa_2;\tilde{t}_2\tilde{t}'_2}}\le(\frac{1}{\hid}\sum_{j=1}^{\hid}s_{\sa_1;t'_1;j}s_{\sa_2;t'_2;j}\ri)\frac{1}{\hid^2}\sum_{k_1,k_2,m_1,m_2=1}^{\hid}\, \nonumber\\
&\ \ \ \ \ \ \times\Bigg(\delta_{t''_1 \tilde{t}_1}\delta_{t''_2 \tilde{t}_2}\delta_{k_1k_2}\delta_{m_1m_2}s_{\sa_1;\tilde{t}'_1;m_1}s_{\sa_2;\tilde{t}'_2;m_2}+\delta_{t''_1 \tilde{t}_1}\delta_{t''_2 \tilde{t}'_2}\delta_{k_1m_2}\delta_{k_2m_1}s_{\sa_1;\tilde{t}'_1;m_1}s_{\sa_2;\tilde{t}_2;m_2}\, \nonumber\\
&\ \ \ \ \ \ \ \ \ +\delta_{t''_1 \tilde{t}'_1}\delta_{t''_2 \tilde{t}_2}\delta_{k_1m_2}\delta_{k_2m_1}s_{\sa_1;\tilde{t}_1;m_1}s_{\sa_2;\tilde{t}'_2;m_2}+\delta_{t''_1 \tilde{t}'_1}\delta_{t''_2 \tilde{t}'_2}\delta_{k_1k_2}\delta_{m_1m_2}s_{\sa_1;\tilde{t}_1;m_1}s_{\sa_2;\tilde{t}_2;m_2}\Bigg)\mathcal{F}_{k_1k_2}^{(\text{previous})}\Bigg]\, \nonumber\\
=&\delta_{i_1i_2}C_U C_V C_Q C_K\frac{1}{H}\sum_{h=1}^{H}\sum_{t'_1,t'_2,\tilde{t}_1,\tilde{t}'_1,\tilde{t}_2,\tilde{t}'_2=1}^{T}\, \nonumber\\
&\times\mathbb{E}\Bigg\{\frac{\partial \ws^{h}_{\sa_1;t_1t'_1}}{\partial \wst^{h}_{\sa_1;\tilde{t}_1\tilde{t}'_1}}\frac{\partial \ws^{h}_{\sa_2;t_2t'_2}}{\partial \wst^{h}_{\sa_2;\tilde{t}_2\tilde{t}'_2}}\le(\frac{1}{\hid}\sum_{j=1}^{\hid}s_{\sa_1;t'_1;j}s_{\sa_2;t'_2;j}\ri)\, \nonumber\\
&\ \ \ \ \ \ \times\le[\delta_{t''_1 \tilde{t}_1}\delta_{t''_2 \tilde{t}_2}\le(\frac{1}{\hid}\sum_{m=1}^{\hid}s_{\sa_1;\tilde{t}'_1;m}s_{\sa_2;\tilde{t}'_2;m}\ri)+\delta_{t''_1 \tilde{t}'_1}\delta_{t''_2 \tilde{t}'_2}\le(\frac{1}{\hid}\sum_{m=1}^{\hid}s_{\sa_1;\tilde{t}_1;m}s_{\sa_2;\tilde{t}_2;m}\ri)\ri]\le(\frac{1}{\hid}\sum_{k=1}^{\hid}\mathcal{F}_{kk}^{(\text{previous})}\ri)\Bigg\}\, \nonumber\\
+&\delta_{i_1i_2}C_U C_V C_Q C_K\frac{1}{H}\sum_{h=1}^{H}\sum_{t'_1,t'_2,\tilde{t}_1,\tilde{t}'_1,\tilde{t}_2,\tilde{t}'_2=1}^{T}\, \nonumber\\
&\times\mathbb{E}\Bigg\{\frac{\partial \ws^{h}_{\sa_1;t_1t'_1}}{\partial \wst^{h}_{\sa_1;\tilde{t}_1\tilde{t}'_1}}\frac{\partial \ws^{h}_{\sa_2;t_2t'_2}}{\partial \wst^{h}_{\sa_2;\tilde{t}_2\tilde{t}'_2}}\le(\frac{1}{\hid}\sum_{j=1}^{\hid}s_{\sa_1;t'_1;j}s_{\sa_2;t'_2;j}\ri)\, \nonumber\\
&\ \ \ \ \ \ \times\Bigg[\delta_{t''_1 \tilde{t}_1}\delta_{t''_2 \tilde{t}'_2}\le(\frac{1}{\hid^2}\sum_{m_1,m_2=1}^{\hid}s_{\sa_1;\tilde{t}'_1;m_1}s_{\sa_2;\tilde{t}_2;m_2}\mathcal{F}_{m_2m_1}^{(\text{previous})}\ri)\, \nonumber\\
&\ \ \ \ \ \ \ \ \ \ +\delta_{t''_1 \tilde{t}'_1}\delta_{t''_2 \tilde{t}_2}\le(\frac{1}{\hid^2}\sum_{m_1,m_2=1}^{\hid}s_{\sa_1;\tilde{t}_1;m_1}s_{\sa_2;\tilde{t}'_2;m_2}\mathcal{F}_{m_2m_1}^{(\text{previous})}\ri)\Bigg]\Bigg\}\, .\nonumber
\end{align}
The first contribution has the first promised form~\eqref{eq:LN-NTK-baby-1} of the layer-normalization multiplicative factor and also is diagonal in the embedding direction~\eqref{eq:backward-goal}. Meanwhile, the second contribution has the second promised form~\eqref{eq:LN-NTK-baby-2}, which we can drop as $O(1/\hid)$.

\subsubsection{Multilayer Perceptron Block}\label{subsubsec:backward-MLP}
Taking derivatives of the residual MLP-block outputs
\be
r_{\sa;t;i}=\sum_{j=1}^{M\hid}X_{ij}\sigma\le(w_{\sa;t;j}\ri)=\sum_{j=1}^{M\hid}X_{ij}\sigma\le(\sum_{k=1}^{\hid}W_{jk}s_{\sa;t;k}\ri)\, ,
\ee
with respect to the MLP-block model parameters and the incoming signals, we have
\begin{align}
\frac{\partial r_{\sa;t;i}}{\partial X_{jk}}=&\delta_{ij}\sigma\le(w_{\sa;t;k}\ri)\, ,\\
\frac{\partial r_{\sa;t;i}}{\partial W_{jk}}=&X_{ij}\sigma'\le(w_{\sa;t;j}\ri)s_{\sa;t;k}\, ,\\
\frac{\partial r_{\sa;t;i}}{\partial s_{\sa;t;k}}=&\sum_{j=1}^{M\hid}X_{ij} \sigma'\le(w_{\sa;t;j}\ri) W_{jk}\, .\label{eq:dangling-MLP}
\end{align}

Expectation values for the squared gradient norms are then given by
\begin{align}
\E{\sum_{j=1}^{\hid}\sum_{k=1}^{M\hid}\frac{\partial r_{\sa_1;t_1;i_1}}{\partial X_{jk}}\frac{\partial r_{\sa_2;t_2;i_2}}{\partial X_{jk}}}=&M\hid \delta_{i_1i_2} \E{\frac{1}{M\hid}\sum_{k=1}^{M\hid}\sigma\le(w_{\sa_1;t_1;k}\ri)\sigma\le(w_{\sa_2;t_2;k}\ri)}\, \\
=&\delta_{i_1i_2} \le\{\le(M\hid\ri)\cdot\le[\le\langle \sigma\le(\widetilde{w}_{\sa_1;t_1}\ri)\sigma\le(\widetilde{w}_{\sa_2;t_2}\ri)\ri\rangle_{C_W F}+O\le(\frac{1}{\hid}\ri)\ri]\ri\}\, ,\nonumber\\
\E{\sum_{j=1}^{M\hid}\sum_{k=1}^{\hid}\frac{\partial r_{\sa_1;t_1;i_1}}{\partial W_{jk}}\frac{\partial r_{\sa_2;t_2;i_2}}{\partial W_{jk}}}=&\hid C_X \delta_{i_1i_2} \mathbb{E}\Bigg\{\Bigg[\frac{1}{M\hid}\sum_{j=1}^{M\hid}\sigma'\le(w_{\sa_1;t_1;j}\ri)\sigma'\le(w_{\sa_2;t_2;j}\ri)\Bigg]\, \\
&\ \ \ \ \ \ \ \ \ \ \ \ \ \ \ \ \ \ \ \ \ \ \ \ \ \ \ \ \ \ \times\le(\frac{1}{\hid}\sum_{k=1}^{\hid}s_{\sa_1;t_1;k}s_{\sa_2;t_2;k}\ri)\Bigg\}\, \nonumber\\
=&\delta_{i_1i_2}\le\{\le(\hid\ri)\cdot \le[ C_X\le\langle \sigma'\le(\widetilde{w}_{\sa_1;t_1}\ri)\sigma'\le(\widetilde{w}_{\sa_2;t_2}\ri)\ri\rangle_{C_W F}F_{(\sa_1;t_1)(\sa_2;t_2)}+O\le(\frac{1}{\hid}\ri)\ri]\ri\}\, ,\nonumber
\end{align}
where we in particular used the factorization formula~\eqref{eq:GP} to simplify the expressions. Note that, as promised~\eqref{eq:backward-goal}, these additive contributions are diagonal in the embedding direction.

For SGD, to ensure order-one additive contributions to the neural tangent kernel, we thus should scale relative learning-rate factors $\LambdaG$ as 
\begin{align}
\LambdaSuper_{W}=\le(\frac{1}{\hid}\ri)\lambdaSub_{W}\, ,\ \ \ \LambdaSuper_{X}=\le(\frac{1}{M\hid}\ri)\lambdaSub_{X}\, \ \ \ \text{for}\ \ \ \text{SGD}\, .
\end{align}
For AdamW, we need to adjust for the magnitude of the per-parameter gradient. For the $W$ weights, each component of the gradient scales as $\vert g^{\mu}\vert\sim\sqrt{\hid/\le(M\hid\cdot \hid\ri)}=1/\sqrt{M\hid}$ and, for the $X$ weights, $\vert g^{\mu}\vert\sim\sqrt{Mn/\le(\hid\cdot M\hid\ri)}=1/\sqrt{\hid}$. Thus, to offset these factors in the AdamW optimizer, we set
\begin{align}
\LambdaSuperA_{W}=\le(\frac{1}{\hid\sqrt{M\hid}}\ri)\lambdaSubA_{W}\, ,\ \ \ \LambdaSuperA_{X}=\le(\frac{1}{M\hid\sqrt{\hid}}\ri)\lambdaSubA_{X}\, \ \ \ \text{for}\ \ \ \text{AdamW}\, .
\end{align}

Finally, moving onto the recursive contributions, the cross terms vanish as
\begin{align}
 \E{\frac{\partial r_{\sa;t;i}}{\partial s_{\sa;t;j}}\mathcal{F}^{(\text{previous})}}=&0\, ,
 \end{align}
 due to the dangling mean-zero weight $X$ in the signal derivative~\eqref{eq:dangling-MLP}, where here $\mathcal{F}^{(\text{previous})}$ is a function of observables from the preceding blocks.
 For the cumulative contribution, we have
 \begin{align}
&\E{\sum_{k_1,k_2=1}^{\hid}\frac{\partial r_{\sa_1;t_1;i_1}}{\partial s_{\sa_1;t_1;k_1}}\frac{\partial r_{\sa_2;t_2;i_2}}{\partial s_{\sa_2;t_2;k_2}}\mathcal{F}_{k_1k_2}^{(\text{previous})}}\label{eq:promised-MLP}\\
=&C_X \delta_{i_1i_2}\mathbb{E}\le\{\le[\frac{1}{M\hid}\sum_{j=1}^{M\hid}\sigma'\le(w_{\sa_1;t_1;j}\ri)\sigma'\le(w_{\sa_2;t_2;j}\ri)\ri]\le[\sum_{k_1,k_2=1}^{\hid}W_{jk_1}W_{j k_2}\mathcal{F}_{k_1 k_2}^{(\text{previous})}\ri]\ri\}\, \nonumber\\
=&C_X C_W\delta_{i_1i_2}\mathbb{E}\le\{\le[\frac{1}{M\hid}\sum_{j=1}^{M\hid}\sigma'\le(w_{\sa_1;t_1;j}\ri)\sigma'\le(w_{\sa_2;t_2;j}\ri)\ri]\le[\frac{1}{\hid}\sum_{k=1}^{\hid}\mathcal{F}_{kk}^{(\text{previous})}\ri]\ri\}+O\le(\frac{1}{\hid}\ri)\, \nonumber\\
=&\delta_{i_1i_2}\le[C_X C_W\le\langle \sigma'\le(\widetilde{w}_{\sa_1;t_1}\ri)\sigma'\le(\widetilde{w}_{\sa_2;t_2}\ri)\ri\rangle_{C_W F}\ri]\E{\frac{1}{\hid}\sum_{k=1}^{\hid}\mathcal{F}^{(\text{previous})}_{kk}}+O\le(\frac{1}{\hid}\ri)\, .\nonumber
\end{align}
Here, in the first equality, we integrated out the $X$ weights; in the second inequality, we integrated out the $W$ weights (where we ignored the interlayer correlations -- see Ref.~\cite{PDLT} -- as subleading $1/\hid$ corrections); in the last equality we used the factorization formula~\eqref{eq:GP} again to simplify the expression. This results in the first promised form~\eqref{eq:LN-NTK-baby-1} of the layer-normalization multiplicative factor and also is diagonal in the embedding direction~\eqref{eq:backward-goal}.

\subsubsection{Head Block}\label{subsubsec:backward-head}
\begin{center}\textit{Vision: linear classification layer}\end{center}

Taking derivatives of the network outputs
\be
z^{(L)}_{\sa;t;i}=\headb_{i}+\sum_{j=1}^{\hid}\head_{ij}s^{(L-1)}_{\sa;t;j}\, 
\ee
with respect to the head-block model parameters and the incoming signals, we have
\begin{align}
\frac{\partial z^{(L)}_{\sa;t;i}}{\partial\headb_j}=&\delta_{ij}\, ,\\
\frac{\partial z^{(L)}_{\sa;t;i}}{\partial\head_{jk}}=&\delta_{ij}s^{(L-1)}_{\sa;t;k}\, ,\\
\frac{\partial z^{(L)}_{\sa;t;i}}{\partial s^{(L-1)}_{\sa;t;j}}=& \head_{ij}\, .\label{eq:dangling-head-vision}
\end{align}

Expectation values for the squared gradient norms are then given by
\begin{align}
\E{\sum_{j=1}^{n_{\text{out}}}\frac{\partial z^{(L)}_{\sa_1;t_1;i_1}}{\partial \headb_{j}}\frac{\partial z^{(L)}_{\sa_2;t_2;i_2}}{\partial \headb_{j}}}=&\delta_{i_1i_2}\cdot\le(1\ri)\, ,\label{eq:additiveNTK-headb}\\
\E{\sum_{j=1}^{n_{\text{out}}}\sum_{k=1}^{\hid}\frac{\partial z^{(L)}_{\sa_1;t_1;i_1}}{\partial \head_{jk}}\frac{\partial z^{(L)}_{\sa_2;t_2;i_2}}{\partial \head_{jk}}}=&\delta_{i_1i_2}\E{\sum_{k=1}^{\hid} s^{(L-1)}_{\sa_1;t_1;k}s^{(L-1)}_{\sa_2;t_2;k}}=\delta_{i_1i_2}\hid\ \E{\frac{1}{\hid}\sum_{k=1}^{\hid} s^{(L-1)}_{\sa_1;t_1;k}s^{(L-1)}_{\sa_2;t_2;k}}\, \nonumber\\
=&\delta_{i_1i_2}\le[\le(\hid\ri)\cdot\kernelLN^{(L-1)}_{(\sa_1;t_1)(\sa_2;t_2)}\ri]\, .\label{eq:additiveNTK-head}
\end{align}
Note that, as promised~\eqref{eq:backward-goal}, these additive contributions are diagonal in the embedding direction.

For SGD, to ensure order-one additive contributions to the neural tangent kernel, we thus should scale relative learning-rate factors as 
\begin{align}
\LambdaSuper_{\headb}=\le(1\ri)\cdot\lambdaSub_{\headb}\, ,\ \ \ \LambdaSuper_{\head}=\le(\frac{1}{\hid}\ri)\lambdaSub_{\head}\, \ \ \ \text{for}\ \ \ \text{SGD}\, .
\end{align}
For AdamW, we need to adjust for the magnitude of the per-parameter gradient. For the biases, each component of the gradient scales as $\vert g^{\mu}\vert\sim1/\sqrt{n_{\text{out}}}$ and, similarly for the weights, $\vert g^{\mu}\vert\sim\sqrt{\hid/(\hid\cdot n_{\text{out}})}=1/\sqrt{n_{\text{out}}}$. Thus, to offset these factors in the AdamW optimizer, we set
\begin{align}
\LambdaSuperA_{\headb}=\le(\frac{1}{\sqrt{n_{\text{out}}}}\ri)\cdot\lambdaSubA_{\headb}\, ,\ \ \ \LambdaSuperA_{\head}=\le(\frac{1}{\hid\sqrt{n_{\text{out}}}}\ri)\lambdaSubA_{\head}\, \ \ \ \text{for}\ \ \ \text{AdamW}\, .
\end{align}

Finally, moving onto the recursive contributions, as parenthetically mentioned right after our  block-to-block recursive formula for neural tangent kernels~\eqref{eq:NTK-forward}, we don't have the skip--residual cross terms in the last block. For the cumulative contribution, we have
\begin{align}
\E{\sum_{j_1,j_2=1}^{\hid}\frac{\partial z^{(L)}_{\sa_1;t_1;i_1}}{\partial s^{(L-1)}_{\sa_1;t_1;j_1}}\frac{\partial z^{(L)}_{\sa_2;t_2;i_2}}{\partial s^{(L-1)}_{\sa_2;t_2;j_2}}\mathcal{F}_{j_1j_2}^{(L-1)}}=&\sum_{j_1,j_2=1}^{\hid}\E{\head_{i_1j_1}\head_{i_2j_2}\mathcal{F}_{j_1j_2}^{(L-1)}}\, \label{eq:promised-head}\\
=&\delta_{i_1i_2}C_{\text{head}}\E{\frac{1}{\hid}\sum_{j=1}^{\hid}\mathcal{F}_{jj}^{(L-1)}}\, .\nonumber
\end{align}
This results in the first promised form~\eqref{eq:LN-NTK-baby-1} of the layer-normalization multiplicative factor and also is diagonal in the embedding direction~\eqref{eq:backward-goal}.\\

\begin{center}\textit{Language: word embedding, transposed (and rescaled)}\end{center}

Taking derivatives of the network outputs
\be
z^{(L)}_{\sa;t;i}=\rescale \sum_{j=1}^{\hid} \WE_{ji}s^{(L-1)}_{\sa;t;j}\, ,
\ee
with respect to the word-embedding parameters and the incoming signals, we have
\begin{align}
\frac{\partial z^{(L)}_{\sa;t;i}}{\partial\WE_{jk}}\Bigg\vert_{\text{head}}=&\rescale\delta_{ik} s^{(L-1)}_{\sa;t; j}\, ,\label{eq:gradient-WE-head}\\
\frac{\partial z^{(L)}_{\sa;t;i}}{\partial s^{(L-1)}_{\sa;t;j}}=&\rescale\WE_{ji}\, .\label{eq:dangling-head-language}
\end{align}
Note that there was another contribution from the stem block~\eqref{eq:gradient-WE-stem} and here we are disambiguating the one from the head.

Then -- recalling $\rescale=\sqrt{1/\hid}$~\eqref{eq:rescaling} -- we have the expectation value for the squared gradient norm given by
\begin{align}
\E{\sum_{i=1}^{\nvocab}\sum_{j=1}^{\hid}\frac{\partial z^{(L)}_{\sa_1;t_1;i_1}}{\partial \WE_{ji}}\Bigg\vert_{\text{head}}\frac{\partial z^{(L)}_{\sa_2;t_2;i_2}}{\partial \WE_{ji}}\Bigg\vert_{\text{head}}}=&\delta_{i_1i_2}\E{\le(\frac{1}{\hid}\sum_{j=1}^{\hid} s^{(L-1)}_{\sa_1;t_1;j}s^{(L-1)}_{\sa_2;t_2;j}\ri)}\, \label{eq:LaT-grad-additive}\\
 =&\delta_{i_1i_2}\le[\le(1\ri)\cdot\kernelLN^{(L-1)}_{(\sa_1;t_1)(\sa_2;t_2)}\ri]\, ,\nonumber
 \end{align}
Note that, as promised~\eqref{eq:backward-goal}, these additive contributions are diagonal in the embedding direction.\footnote{To be complete, in addition to this additive term~\eqref{eq:LaT-grad-additive} and the cumulative term, we need to consider stem--head cross gradient terms. As in long footnote~\ref{foot:long}, the most efficient way to get at the crux of the matter is to work out the toy model  $z^{(3)}_{\sa;t;i}=\rescale\sum_{j,k}^{\hid}\sum_{m=1}^{\nvocab}\WE_{ji}\le(W^{(2)}_{jk}+\delta_{jk}\ri)\WE_{km}x_{\sa;t;m}$, for which we find
\be
\E{\sum_{i=1}^{\nvocab}\sum_{j=1}^{\hid}\frac{\partial z^{(3)}_{\sa_1;t_1;i_1}}{\partial \WE_{ji}}\Bigg\vert_{\text{head}}\frac{\partial z^{(3)}_{\sa_2;t_2;i_2}}{\partial \WE_{ji}}\Bigg\vert_{\text{stem}}}=C_{\text{WE}}\le(1+\frac{C_{W^{(2)}}}{\hid}\ri)x_{\sa_1;t_1;i_2}x_{\sa_2;t_2;i_1}\, ,\label{eq:LaT-grad-cross} 
 \ee
 with the leading term coming from the skip path. Similarly to our observations in footnote~\ref{foot:long}, this contact contribution is zero except for one specific component and the same contribution arises in more general cases.} 

For SGD, to ensure the order-one additive contributions to the neural tangent kernel, we thus should scale the relative learning-rate factor as 
\begin{align}
\LambdaSuper_{\WE}=\le(1\ri)\cdot\lambdaSub_{\WE}\, \ \ \ \text{for}\ \ \ \text{SGD}\, .
\end{align}
For AdamW, we need to adjust for the magnitude of the per-parameter gradient $\vert g^{\mu}\vert\sim1/\sqrt{\hid}$. Thus, to offset these factors in the AdamW optimizer, we set
\begin{align}
\LambdaSuperA_{\WE}=\le(\frac{1}{\sqrt{\hid}}\ri)\lambdaSubA_{\WE}\, \ \ \ \text{for}\ \ \ \text{AdamW}\, .
\end{align}
Importantly, both of these scalings are consistent with the scalings~\eqref{eq:WE-scale-stem-SGD} and~\eqref{eq:WE-scale-stem-AdamW} we decided to set for them by analyzing the gradient contribution from  the stem block.\footnote{Further consistently, as discussed in footnote \ref{foot:nvocab-stem} for the stem block, the scaling of the relative AdamW learning-rate factor $\LambdaSuperA_{\WE}$ with the vocabulary size $\nvocab$ is subtle -- if not subtler -- for the head block.\label{foot:nvocab-head} Specifically, when using the cross-entropy loss, the gradient respect to the word-embedding parameter $\WE_{ij}$ gets a factor of $p_{\sa;t;j}-q_{\sa;t;j}$ where $p_{\sa;t;j}$ is the softmax (in the embedding direction $j$) distribution of the output $z^{(L)}_{\sa;t;j}$ and $q_{\sa;t;j}$ is the target (typically one-hot) distribution for the task at hand. There is then a subtlety even for a small batch size $\vert \mathcal{A}_t\vert$ as then most of these gradients are small but nonzero (roughly of order $\sim1/\nvocab$ at initialization) and their noisy signals get amplified in the AdamW optimizer -- unless one decides to explicitly focus on the top-few components in the $j$ direction or implicitly regularize with appropriately chosen $\epsilon$.}

Finally, moving onto the recursive contributions, as parenthetically mentioned right after our  block-to-block recursive formula for neural tangent kernels~\eqref{eq:NTK-forward}, we don't have the skip--residual cross terms in the last layer. For the cumulative contribution, we have
\begin{align}
\E{\sum_{j_1,j_2=1}^{\hid}\frac{\partial z^{(L)}_{\sa_1;t_1;i_1}}{\partial s^{(L-1)}_{\sa_1;t_1;j_1}}\frac{\partial z^{(L)}_{\sa_2;t_2;i_2}}{\partial s^{(L-1)}_{\sa_2;t_2;j_2}}\mathcal{F}_{j_1j_2}^{(L-1)}}=&\rescale^2\sum_{j_1,j_2=1}^{\hid}\E{\WE_{j_1i_1}\WE_{j_2i_2}\mathcal{F}_{j_1j_2}^{(L-1)}}\, \\
=&\delta_{i_1i_2}C_{\text{WE}}\E{\frac{1}{\hid}\sum_{j=1}^{\hid}\mathcal{F}_{jj}^{(L-1)}}\, .\nonumber
\end{align}
This results in the first promised form~\eqref{eq:LN-NTK-baby-1} of the layer-normalization multiplicative factor and also is diagonal in the embedding direction~\eqref{eq:backward-goal}.\footnote{Continuing -- and ending --  our usual operation of hiding the subtle interlayer correlation due to the stem--head weight tying, we here simply note that the first-block neural tangent kernel $\NTK^{(1)}_{(\sa_1;t_1;i_1)(\sa_2;t_2;i_2)}$ doesn't depend on the word-embedding parameters~\eqref{eq:LaT-NTK-stem-WE}, and using the leading-order layer-normalization mnemonics~\eqref{eq:LN-mnemonic}, there is nothing to worry about here.}

\newpage

%\section{Modern Applications}\label{sec:applications}
\section{Practical Applications}\label{sec:applications}

\setlength{\epigraphwidth}{0.21\textwidth}
\epigraph{\textit{Such heroic nonsense.}}{Megatron}

In Part~\ref{sec:foundations} of this note, we've developed an effective theory of Transformers at leading order. Such an analysis in particular suggested proper width scalings of initialization and training hyperparameters that collectively ensure benevolent limiting behaviors when widening Transformers. These suggestions, however, would be theoretical nonsense if they don't better the practice in one way or the other. In this Part~\ref{sec:applications}, we'll thus put our theoretical suggestions to practical tests, training both Vision and Language Transformers.

In \S\ref{subsec:ViT}, we'll train Vision Transformers for an image classification task. For them, initialization hyperparameters are often -- though not always -- scaled well with width, so we'll focus on comparing the standard uniform learning rate -- that is, $\LambdaGA=1$ in our language -- with our non-uniform neural-tangent scalings~\eqref{eq:AdamWlambdaG-preview-stem}--\eqref{eq:AdamWlambdaG-preview-head} of learning rates. In our training setup, we'll find that the neural-tangent scaling strategy can not only improve the model performance but also reduce the frequency of mid-training spikes when compared with the uniform learning rate.\footnote{We'll also test the maximal-update scaling strategy~\cite{yang2021tensor} and the hybrid of neural-tangent and maximal-update scaling strategies~\cite{yaida2022meta} for Vision Transformers. We'll find that they both reduce the frequency of mid-training spikes and -- among all the four scaling strategies tested herein -- the hybrid scaling strategy yields the best performance while the maximal-update scaling strategy yields the worst.}

In \S\ref{subsec:SLUDGE}, we'll pretrain Language Transformers for a span denoising task. For them, we'll make three changes to the standard uniform (non-)scaling of hyperparameters: \textit{(i)} the initialization hyperparameter for the word-embedding parameters will be cranked up from the standard $C_{\text{WE}}=(0.02)^{2}$ to $C_{\text{WE}}=1$; \textit{(ii)} concomitantly the network output is rescaled with $\rescale=1/\sqrt{\hid}$ instead of the standard non-rescaling $\rescale=1$; and \textit{(iii)} again the relative learning-rate factors $\LambdaGA$ are set according to the neural-tangent scaling strategy. Here, our results are slightly more mixed than those in \S\ref{subsec:ViT}, but we'll nonetheless include them as encouraging anecdata for larger scales.

\
\\
\textit{Programming note 1}: mirroring our theoretical treatment in Part~\ref{sec:foundations} of this note,
%and fearing further hyperparameter tunings
for all models, we'll drop bias parameters from MHSA and MLP blocks -- as practiced, e.g., by some large language models such as PaLM~\cite{chowdhery2022palm} -- and keep element-wise affine parameters in normalization layers fixed.\\
%In PyTorch, this means that we set $\texttt{bias}=\texttt{False}$ and $\texttt{elementwise\_affine}=\texttt{False}$.
\textit{Programming note 2}: for all experiments, we'll use PyTorch's automatic mixed precision~\cite{micikevicius2017mixed}, to better reflect the current practices and speed up training.

\newpage
\subsection{Image Classification with Encoder-Only Transformers}\label{subsec:ViT}
Here we'll train encoder-only Vision Transformers with an image classification objective.\footnote{Following Ref.~\cite{liu2022convnet}, all models were trained by using the codebase available at \url{https://github.com/facebookresearch/convnext}; to it, we added our own model code for Vision Transformers and also introduced the relative learning-rate factors~$\LambdaGA$'s to the AdamW optimizer code.}
We'll describe our dataset and task in \S\ref{subsubsec:data_ViT}, architecture hyperparameters in \S\ref{subsubsec:architecture_ViT}, initialization hyperparameters in \S\ref{subsubsec:initialization_ViT}, and optimizer and training hyperparameters in \S\ref{subsubsec:train_ViT}.
In \S\ref{subsubsec:versus_ViT}, we'll then compare the standard uniform scaling of learning rates with the neural-tangent scalings.

\subsubsection{Dataset and Task}\label{subsubsec:data_ViT}
As for the dataset, we use the ILSVRC challenge version~\cite{russakovsky2015imagenet} of the ImageNet-1k dataset~\cite{deng2009imagenet}, containing RGB images of 1000 different categories. The dataset is split into the training set containing 1,281,167 images and the validation set containing 50,000 images. Both for the training set and for the validation set,  as is customary, each image is preprocessed by subtracting the $\texttt{mean}=[0.485, 0.456, 0.406]$ and dividing by the standard deviation $\texttt{std}=[0.229,0.224, 0.225]$. 

As for the training objective, we augment images first with the by-now standard random resizing and cropping to $224$-by-$224$ and random horizontal flip,
then with RandAugment \cite{cubuk2020randaugment} ($n=2$, $m=9$), one of Mixup \cite{zhang2017mixup} ($\alpha=0.8$) or CutMix \cite{yun2019cutmix} ($\alpha=1.0$) chosen with equal probability $50\%$, and apply a single Random Erasing \cite{zhong2020random} rectangle with probability $25\%$ and with values set randomly per-pixel within the block.\footnote{As a full specification for RandAugment, we use \texttt{rand-m9-mstd0.5-inc1} in \texttt{timm}~\cite{rw2019timm}, which applies magnitude noise of the standard deviation $\texttt{std}=0.5$.}
We then optimize the cross-entropy loss between the model outputs and target labels, with label smoothing~\cite{szegedy2016rethinking} of 0.1.
%Eric Mintun taught S.Y.: timm silently turns off color jitter when randaug is on, so it is effectively not included.

At evaluation, images in the validation set are scaled to $256$-by-$256$ and center cropped to $224$-by-$224$, without any data argumentation. We then evaluate the top-one prediction accuracy.

\subsubsection{Architecture Hyperparameters}\label{subsubsec:architecture_ViT}
Overall, our architectural design follows the original Vision Transformers~\cite{dosovitskiy2020image} with one difference: as noted in the introduction of this Part~\ref{sec:applications}, we turn off all the bias parameters in MHSA and MLP blocks~\cite{chowdhery2022palm} and don't train element-wise affine parameters in normalization layers.

With that difference in mind, our Vision-Transformer architecture is exactly that described in the \textit{Vision} track of \S\ref{subsec:forward}, with the patch dimension $\npatch=16\cdot16\cdot3=768$, the number of tokenized patches $T=196$, the embedding dimension $n=768$, the normalization layer regularization $\epsilon=10^{-6}$,  the number of MHSA heads $H=12$, the MLP multiplier $M=4$, \texttt{GELU} activation functions in the MLP blocks, and output dimension $n_{\text{out}}=1000$; in the bulk, $12$ encoders -- each encoder consisting of one bidirectional MHSA block followed by one MLP block -- are stacked.

To make sure we are on the same footing, please check that those architectural choices result in $P=768\cdot768+196\cdot768+768^2\cdot(4+2\cdot 4)\cdot12+768\cdot1000+1000\approx 86\cdot 10^6$ model parameters.

\subsubsection{Initialization Hyperparameters}\label{subsubsec:initialization_ViT}
As for the initialization, the practices vary in the literature. In our experiments, for both standard and neural-tangent runs, we decided to more or less follow the PyTorch default at some points in time, that is, in the bulk, for the weights in the MHSA blocks we use mean-zero uniform distributions with $C_Q=C_K=C_V=2/(3+1)=1/2$ and $C_U=1/3$ and for the weights in the MLP blocks we use mean-zero uniform distributions with $C_W=2/(4+1)=2/5$, and $C_X=2\cdot4/(4+1)=8/5$.\footnote{The most of them derives directly from applying the standard Xavier initialization -- where we in particular note that $Q$--$K$--$V$ weights are often coded as one linear layer instead of three linear layers -- while for $U$ the factor of $1/3$ in the covariance derives from uniformly distributing weights in the interval $[-1/\sqrt{\hid},1/\sqrt{\hid}]$.} These are acceptable as order-one numbers. As for the stem and head blocks, we simply use a mean-zero normal distribution with $C_{\text{patch}}=C_{\text{head}}=1$ and $\headb_i=0$ while $C_{\text{PE}}=(0.02)^2$ for the positional-embedding parameters as practiced sometimes.\footnote{After our theoretical exposition, this last choice may look unnaturally small but\label{foot:002} -- since the positional-embedding parameters act like bias parameters (i.e., not multiplicative but additive) and hence we could in principle set this hyperparameter to any order-one number including $C_{\text{PE}}=0$ -- we adopt it as acceptable. In general, we tried to minimize the amount of changes with respect to the literature.}

\subsubsection{Optimizer and Training Hyperparameters}\label{subsubsec:train_ViT}
Overall, our training recipe almost exactly follows the one used in Ref.~\cite{liu2022convnet} -- except that we here ablate \textit{(i)} \texttt{DropPath}s~\cite{huang2016deep} and \textit{(ii)} exponential moving averaging~\cite{polyak1992acceleration}, in order to make the comparison slightly less confounded on their respective hyperparameters.

As for the optimizer,  we use AdamW~\cite{kingma2014adam, loshchilov2017decoupled}~\eqref{eq:AdamW1}--\eqref{eq:AdamW4} with $(\beta_1,\beta_2,\epsilon)= (0.9, 0.999,10^{-8})$;
also note that weight decay is \textit{not} applied to
the bias parameters in the head block~\cite{liu2022convnet}.

As for the learning schedule, we use a batch size of $4096$ -- which results in $312$ iterations per epoch -- with linear warmup~\cite{goyal2017accurate} for the first $20$ epochs, followed by a cosine learning-rate decay~\cite{loshchilov2016sgdr} over the next $280$ epochs, that is,
\begin{align}
\eta_t=\begin{cases}
\lrt\cdot\le(\frac{t}{312\cdot 20}\ri)\, \ \ \ \ \ \ \ \ \ \ \ \ \ \ \ \ \ \ \ \ \ \ \ \ \text{for}\ \ \ t\leq312\cdot 20\, , \\
\lrt_{\text{min}}+(\lrt-\lrt_{\text{min}})\cdot\le\{\frac{1}{2}+\frac{1}{2}\cos\le[\frac{\le(t-312\cdot 20\ri)\pi}{312\cdot 280}\ri]\ri\}\, \ \ \text{for}\ \ \ t>312\cdot 20\, ,
\end{cases}
\end{align}
with $\lrt_{\text{min}}=10^{-6}$.\footnote{In retrospect, the effect of $\lrt_{\text{min}}=10^{-6}$ kicks in roughly only in the last $\sim 5$ epochs for the standard scaling runs (at $\lrt^{\star}=0.001$) and in the last $\sim10$ iterations for the neural-tangent scaling runs (at $\lrt^{\star}=20$).}
We train the models for the full $300$ epochs; we'll discuss the settings of the overall learning rate $\lrt$ and the weight decay $\wdt$ shortly  in \S\ref{subsubsec:versus_ViT}.

As for the per-group learning-rate factors $\LambdaGA$'s, for the standard runs, we use the standard uniform scaling $\LambdaGA=1$, while for the neural-tangent runs, we follow our theoretical suggestions~\eqref{eq:AdamWlambdaG-preview-stem}--\eqref{eq:AdamWlambdaG-preview-head}, except that we ignore the factor of $M=4$: see Table~\ref{table:ViT-NTK}.\footnote{To closely follow the theoretical suggestions, we should've set $\LambdaSuperA_{W}=1/(\hid\sqrt{M\hid})$ and $\LambdaSuperA_{X}=1/(M\hid\sqrt{\hid})$ but we instead caved to $\LambdaSuperA_{W}=\LambdaSuperA_{X}=1/(\hid\sqrt{\hid})$, given that it is easier to implement for larger language models -- where sharding and flattening sometimes complicate the coding. However, if anyone scales $M$ large in the future, then $\LambdaSuperA_{W}$ and $\LambdaSuperA_{X}$ should be scaled properly with $M$.} Overall, with $\hid=768\sim 1000$, our neural-tangent scaling essentially boils down to cranking up the learning rate for positional-embedding parameters (and bias parameters in the head block) by a factor of $\hid\sim1000$ with respect to other model parameters.

\subsubsection{Comparison of Scaling Strategies}\label{subsubsec:versus_ViT}
We here compare the performances of the models trained with the standard uniform learning rate against the ones trained with the neural-tangent scalings of learning rates: see Table~\ref{table:ViT-NTK} below for a concise summary.
\begin{table}[h]\label{table:ViT-NTK} 
\scalebox{0.8}{
\begin{tabular}{|c|c|c|}
\hline
\multicolumn{3}{|c|}{AdamW--standard runs} \\\hline
    & initial \texttt{std} & relative \texttt{lr} factors \\\hline
patchify $\patch_{ij}$   &$768^{-\frac{1}{2}}$ &  1  \\\hline
positional embedding  $\PE_{t;i}$ &0.02 &  1 \\\hline
$Q$--$K$--$V$--$U$--$W$--$X$ weights   &$\sharp\hid^{-\frac{1}{2}}$ &  1 \\\hline
head weights $\head_{ij}$   &$\hid^{-\frac{1}{2}}$ &  1  \\\hline
head biases $\headb_{i}$   & 0 &  1  \\\hline
\end{tabular}
\begin{tabular}{|c|c|c|}
\hline
\multicolumn{3}{|c|}{AdamW--neural-tangent runs} \\\hline
    & initial \texttt{std} & relative \texttt{lr} factors \\\hline
patchify $\patch_{ij}$   &$768^{-\frac{1}{2}}$ & {\color{blue}$768^{-1}\cdot\hid^{-\frac{1}{2}}$}  \\\hline
positional embedding  $\PE_{t;i}$ &0.02 &  {\color{blue}$\hid^{-\frac{1}{2}}$}\\\hline
$Q$--$K$--$V$--$U$--$W$--$X$ weights   &$\sharp\hid^{-\frac{1}{2}}$ &  {\color{blue}$\hid^{-1}\cdot\hid^{-\frac{1}{2}}$} \\\hline
head weights $\head_{ij}$   &$\hid^{-\frac{1}{2}}$ &  {\color{blue}$n^{-1}\cdot1000^{-\frac{1}{2}}$}  \\\hline
head biases $\headb_{i}$   & 0 &  {\color{blue}$1000^{-\frac{1}{2}}$}  \\\hline
\end{tabular}
 }
\scalebox{0.8}{
\begin{tabular}{|c|c|c|}
\hline
\multicolumn{3}{|c|}{AdamW--$\sqrt{\text{neural-tangent}\cdot\text{maximal-update}}$ runs} \\\hline
    & initial \texttt{std} & relative \texttt{lr}  factors \\\hline
patchify $\patch_{ij}$   &$768^{-\frac{1}{2}}$ &  {\color{purple}$768^{-1}\cdot\hid^{-\frac{1}{4}}$}  \\\hline
positional embedding  $\PE_{t;i}$ &0.02 &  {\color{purple}$\hid^{-\frac{1}{4}}$}\\\hline
$Q$--$K$--$V$--$U$--$W$--$X$ weights   &$\sharp\hid^{-\frac{1}{2}}$ &  {\color{purple}$\hid^{-1}\cdot\hid^{-\frac{1}{4}}$} \\\hline
head weights $\head_{ij}$   &{\color{purple}$\hid^{-\frac{3}{4}}$} &  {\color{purple}$n^{-1}\cdot1000^{-\frac{1}{2}}$}  \\\hline
head biases $\headb_{i}$   & 0 &  {\color{purple}$1000^{-\frac{1}{2}}$}  \\\hline
\end{tabular}
\begin{tabular}{|c|c|c|}
\hline
\multicolumn{3}{|c|}{AdamW--maximal-update runs} \\\hline
    & initial \texttt{std} & relative \texttt{lr}  factors \\\hline
patchify $\patch_{ij}$   &$768^{-\frac{1}{2}}$ &  {\color{red}$768^{-1}$}  \\\hline
positional embedding  $\PE_{t;i}$ &0.02 &  {\color{red}$1.0$}\\\hline
$Q$--$K$--$V$--$U$--$W$--$X$ weights   &$\sharp\hid^{-\frac{1}{2}}$ &  {\color{red}$\hid^{-1}$} \\\hline
head weights $\head_{ij}$   &{\color{red}$\hid^{-1}$} &  {\color{red}$n^{-1}\cdot1000^{-\frac{1}{2}}$}  \\\hline
head biases $\headb_{i}$   & 0 &  {\color{red}$1000^{-\frac{1}{2}}$}  \\\hline
\end{tabular}
}
\end{table}

\begin{figure}[h]
   \centerline{
   \includegraphics[width=0.5\linewidth]{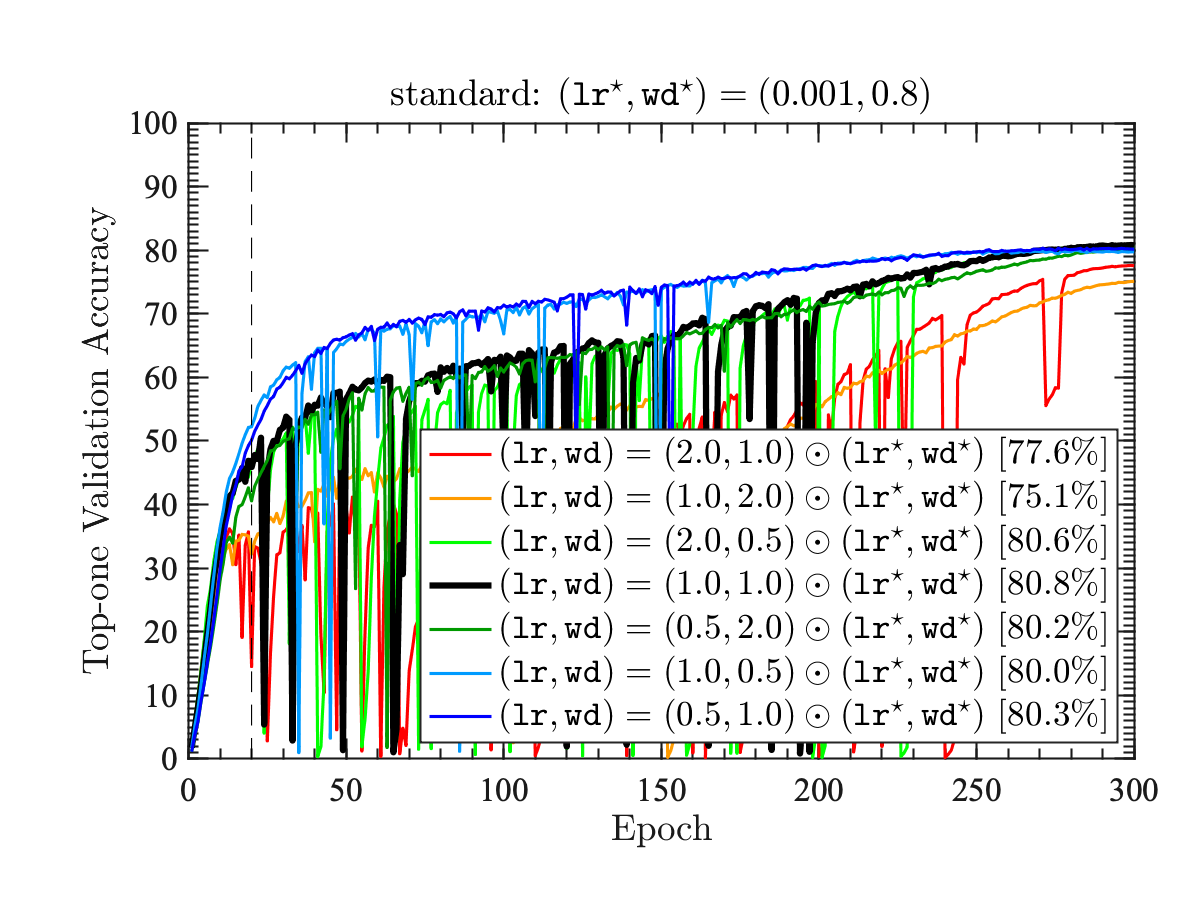}\!\!\!
   \includegraphics[width=0.5\linewidth]{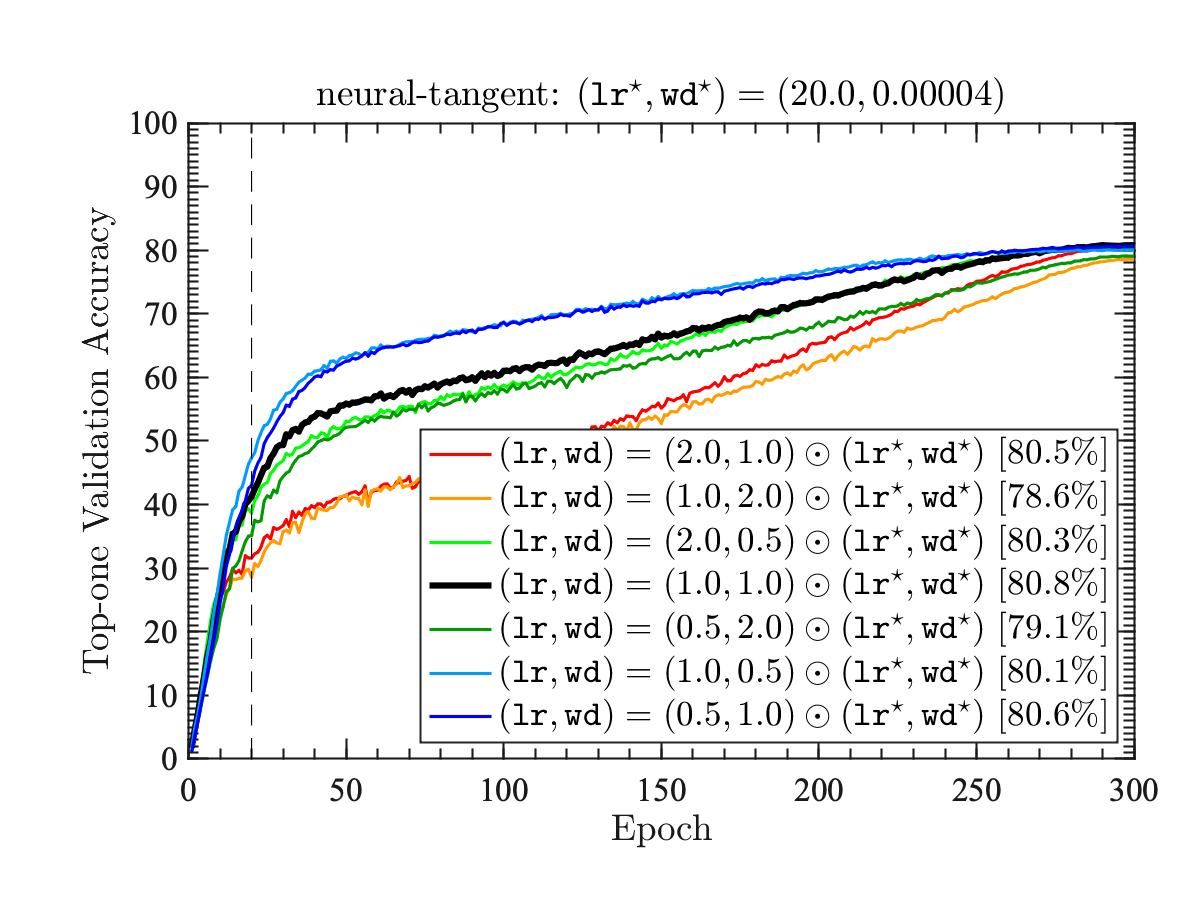}
   }
   \centerline{
   \includegraphics[width=0.5\linewidth]{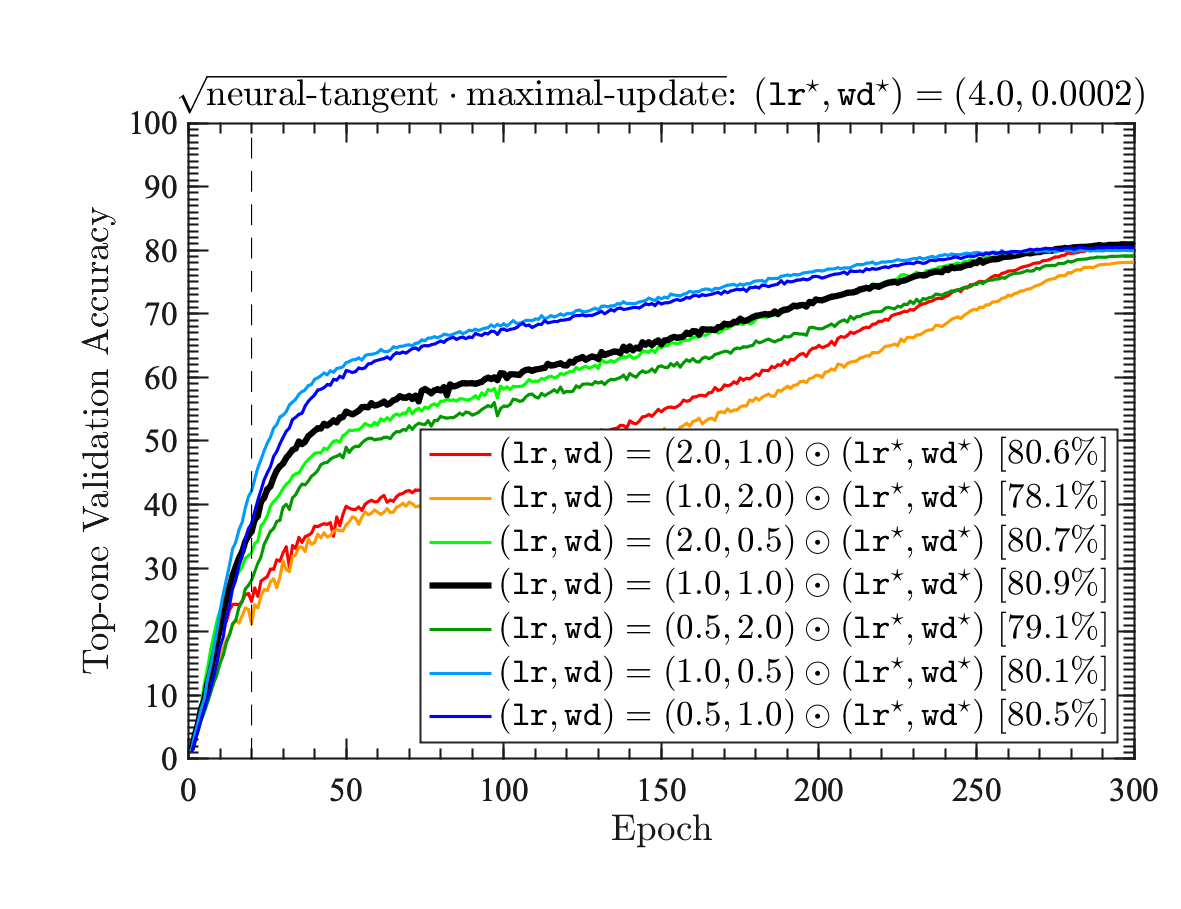}\!\!\!
   \includegraphics[width=0.5\linewidth]{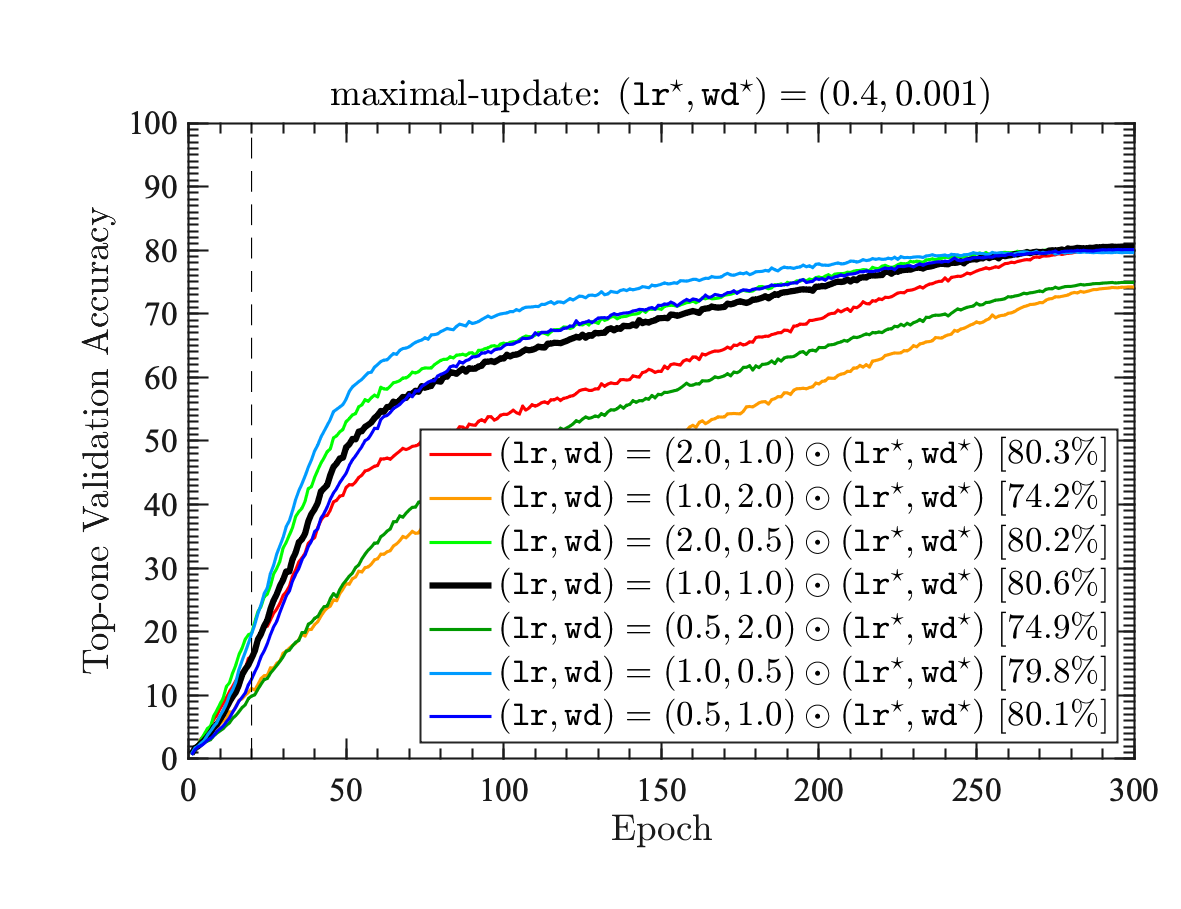}
   }
\caption{Hyperparameter searches for standard uniform (top left), neural-tangent (top right), hybrid neural-tangent--maximal-update (bottom left), and maximal-update (bottom right) scaling strategies for Vision Transformers trained by AdamW$_{(\beta_1,\beta_2,\epsilon)=(0.9,0.999,\text{1e-8})}$. For each, the top-one validation accuracy on the ImageNet-1k dataset is plotted as a function of training epochs. In the legend, we record the training hyperparameter pair $(\lrt,\wdt)$ [with the max top-one validation accuracy along each trajectory].}
\label{fig:hyper-ViT}
\end{figure}

\begin{figure}[h]
   \centerline{
   \includegraphics[width=0.75\linewidth]{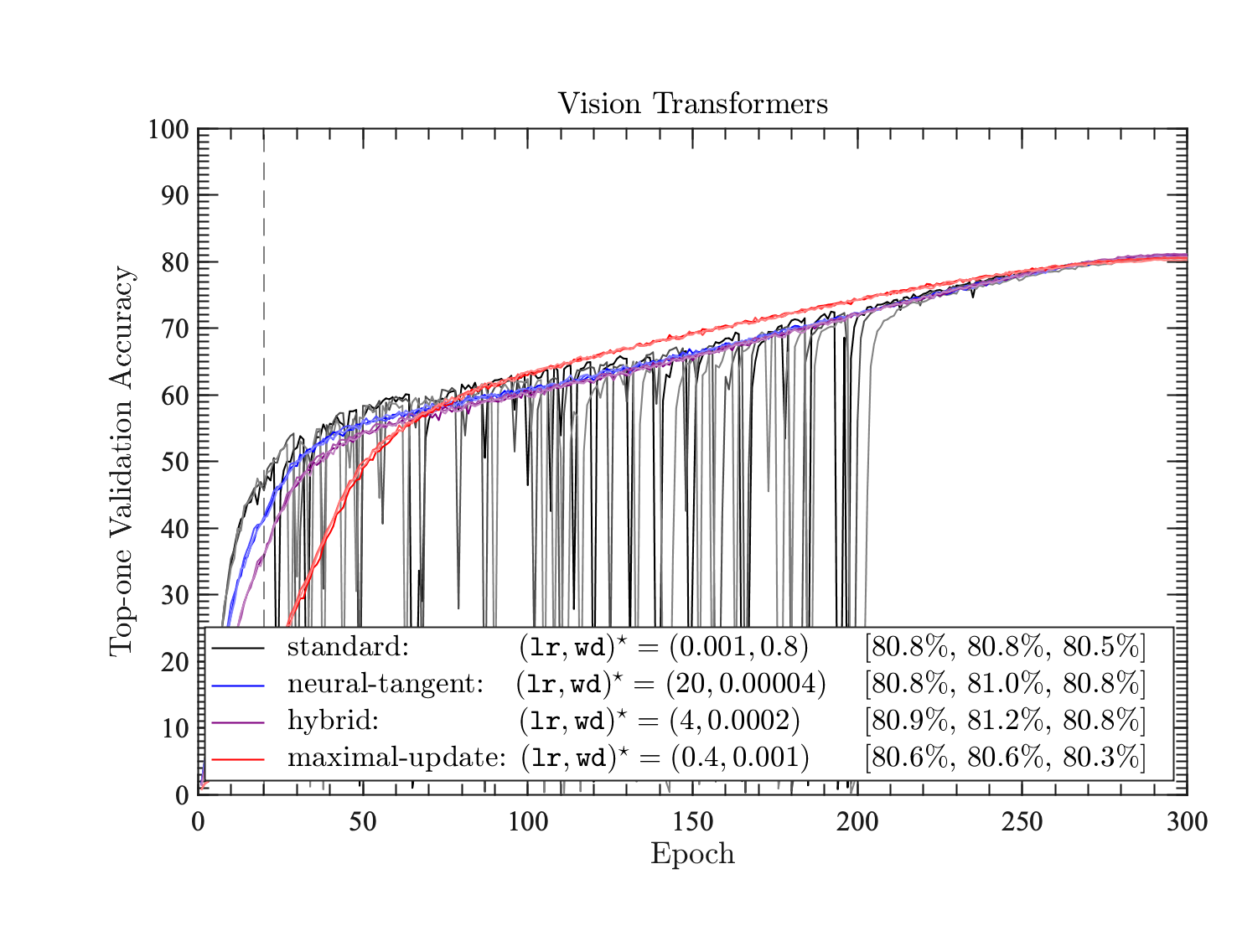}
   }
\caption{Comparison of the standard (black), neural-tangent (blue), hybrid neural-tangent--maximal-update (purple), and maximal-update (red) scaling strategies for Vision Transformers trained by AdamW$_{(\beta_1,\beta_2,\epsilon)=(0.9,0.999,\text{1e-8})}$. For each, the validation accuracy is plotted as a function of training epochs, for three different seeds (different whiteness). In the legend, we record the optimal training hyperparameter pair $(\lrt^{\star},\wdt^{\star})$ for each scaling strategy  [with the max top-one validation accuracy along each trajectory].}
\label{fig:seeds-ViT}
\end{figure}

In order to provide rigorous comparisons -- within reason -- we tune the overall learning rate $\lrt$  and the weight decay $\wdt$ for each scaling strategy: see Fig.~\ref{fig:hyper-ViT}. Specifically, we search for the optimal $(\lrt,\wdt)$ in $\log_{2}$ grid space; once we find the candidate optimum $(\lrt^{\star},\wdt^{\star})$, we make sure that the top-one validation accuracies are lower for $(\lrt,\wdt)=(2\cdot\lrt^{\star},\wdt^{\star}),\ (\lrt^{\star},2\cdot\wdt^{\star}),\ (2\cdot\lrt^{\star},0.5\cdot\wdt^{\star}),\ (0.5\cdot\lrt^{\star},2\cdot\wdt^{\star}),\ (\lrt^{\star},0.5\cdot\wdt^{\star}),\ (0.5\cdot\lrt^{\star},\wdt^{\star})$.\footnote{To save compute, we here omit  $(\lrt,\wdt)=(2\cdot\lrt^{\star},2\cdot\wdt^{\star}),\ (0.5\cdot\lrt^{\star},0.5\cdot\wdt^{\star})$ because, if doubling/halving $\lrt$ and $\wdt$ \textit{each individually} degrades the performance, then we expect that doubling/halving \textit{both together} degrades the performance as well.} 
Once the optimal pair $(\lrt^{\star},\wdt^{\star})$ is found for each scaling strategy, we ran the experiments for three different seeds: see Fig.~\ref{fig:seeds-ViT}.

Overall, in our experimental setup, we see that the neural-tangent scaling strategy not only statistically improves the performance but also reduces the frequency of mid-training spikes when compared to the standard scaling strategy.\footnote{We also observe the same reduction of spikes  for the neural-tangent--maximal-update hybrid and maximal-update scaling strategies: see Table~\ref{table:ViT-NTK} and Figs.~\ref{fig:hyper-ViT} and \ref{fig:seeds-ViT}.\label{foot:expensiveViT} Incidentally, in our experimental setup, among four scaling strategies, the hybrid scaling strategy yields the best performance  while the maximal-update scaling strategy yields the worst performance.}

We additionally note that, for the neural-tangent runs, the effective overall learning rate, $\lrt\cdot\LambdaGA$, for $Q$--$K$--$V$--$U$--$W$--$X$ weights at optimality is given by $20\cdot 768^{-3/2}\approx0.94\cdot10^{-4}\sim0.001$, which is essentially equal to the optimal learning rate for the standard runs; we further note that $\lrt^{\star}\cdot\wdt^{\star}=8\cdot10^{-4}$ for both scaling strategies. We thus conclude that the stability and performance improvement of the neural-tangent scaling strategy essentially arises from \textit{increasing} the learning rate for the positional-embedding parameters by a factor of $\hid\sim 1000$.\footnote{Technically speaking, we are also increasing the learning rate for the biases in the head block. However, we also performed experiments with the biases in the head block turned off and observed the qualitatively similar results, so we believe that the claimed benefits arise from our proper treatment of the positional-embedding parameters.}

\newpage
\subsection{Span Denoising with Encoder--Decoder Transformers}\label{subsec:SLUDGE}
Here we'll pretrain encoder--decoder Language Transformers with a span denoising objective.\footnote{Following Refs.~\cite{lewis2019bart,R2C2seeker2022}, all models were trained by using Fairseq~\cite{ott2019fairseq} available at \url{https://github.com/facebookresearch/fairseq}.}
We'll describe our dataset and task in \S\ref{subsubsec:data_SLUDGE}, architecture hyperparameters in \S\ref{subsubsec:architecture_SLUDGE}, initialization hyperparameters in \S\ref{subsubsec:initialization_SLUDGE}, and optimizer and training hyperparameters in \S\ref{subsubsec:train_SLUDGE}.
In \S\ref{subsubsec:versus_SLUDGE}, we'll then compare the standard uniform scalings of hyperparameters with the neural-tangent scalings.

\subsubsection{Dataset and Task}\label{subsubsec:data_SLUDGE}
As for the dataset, following Ref.~\cite{R2C2seeker2022}, we use the dataset that combines the corpora used by RoBERTa~\cite{roberta2019}  -- which by themselves are amalgamation of BookCorpus~\cite{zhu2015aligning} plus English Wikipedia~\cite{devlin2018bert}, CommonCrawl News~\cite{nagel2016cc}, OpenWebText~\cite{gokaslan2019openwebtext,radford2019language}, and Stories~\cite{trinh2018simple} -- with the English subset of the CC-100 corpus~\cite{conneau-etal-2020-unsupervised} and a dump of pushshift.io Reddit  compiled by a third party \cite{baumgartner2020pushshift,bb1roller2020}. We employ the same byte-pair encoding tokenizer as GPT-2~\cite{radford2019language}. Both for the training set and for the validation set, each input sample $\alpha$ is a set of sentences taken from these documents
% -- without crossing document boundaries --
with its total sequence length up to $512$, always starting and ending at the beginning of some sentence and at the end of some -- same or different -- sentence, where we also add a \texttt{[start]} token and an \texttt{[end]} token, respectively; we further pad \texttt{[pad]} tokens to make the full sequence length always be $514$.

As for the training objective, following Ref.~\cite{lewis2019bart}, we first corrupt documents by randomly permuting sentences and then repeately infilling a span of tokens with a \texttt{[MASK]} token -- where each span length is randomly sampled from a Poisson distribution with its mean and variance $3$ -- until 30\% of input sequence tokens are corrupted.
% each span begins at the beginning of a word
We then optimize the cross-entropy loss between the model output and the original uncorrupted sample -- discarding the \texttt{[pad]}s.

At evaluation, we corrupt documents and measure the cross-entropy in the same way, albeit now with the validation set.

\subsubsection{Architecture Hyperparameters}\label{subsubsec:architecture_SLUDGE}
Overall, our architectural design follows those of BART-large~\cite{lewis2019bart} and R2C2~\cite{R2C2seeker2022}, again with one difference:
as noted in the introduction of this Part~\ref{sec:applications}, we turn off all the bias parameters in MHSA and MLP blocks~\cite{chowdhery2022palm} and don't train element-wise affine parameters in normalization layers.\footnote{Like the original Language Transformer~\cite{vaswani2017attention}, the original BART-large~\cite{lewis2019bart} -- but not R2C2~\cite{R2C2seeker2022} -- placed normalization layers at the end of each block; as mentioned in \S\ref{subsec:forward}, as is standard nowadays, we here place normalization layers at the beginning of each residual path for both BART-large and R2C2.}

With that difference in mind, our BART-large (R2C2) architecture is \textit{almost} -- see the next paragraph -- that described in the \textit{Language} track of \S\ref{subsec:forward}, with the vocabulary size $\nvocab=50265$, the sequence length $T=514$, the width $n=1024$ ($2048$), the normalization layer regularization $\epsilon=10^{-5}$,  the number of attention heads $H=16$ ($32$), the MLP multiplier $M=4$, and \texttt{GELU} activation functions in the MLP blocks; in the bulk, $12$ ($22$) encoders -- each encoder consisting of one bidirectional MHSA block followed by one MLP block -- are stacked first and then $12$ ($22$) decoders -- each decoder consisting of one bidirectional MHSA block, followed by one soon-to-be-described encoder--decoder multi-headed \textit{mixed}-attention block, further followed by one MLP block  -- are stacked.

Compared to what's outlined in the \textit{Language} track of \S\ref{subsec:forward}, there are a few additional gadgets for the BART-large and R2C2 architecture:
\bi
\item Right after the stack of the encoder blocks and right before the stack of the decoder blocks, one additional set of positional-embedding parameters is placed, which is often considered as a stem block in the decoder stack.
\item In addition to the normalization layers applied at the beginning of each residual path,
%but \textit{not} applied to the xeroxed preactivations in the skip path: see Eq.~\eqref{eq:resrec}
there are three additional ones placed \textit{(1)} right after the first positional-embedding parameters in the stem block of the encoder stack, \textit{(2)} right after the stack of all the encoder blocks, and \textit{(3)} right after the second positional-embedding parameters in junction just mentioned above, i.e., in the stem block of the decoder stack.
\item In the encoder--decoder multi-headed mixed-attention blocks in the decoders, instead of the usual query--key--value vectors~\eqref{eq:q-vector}--\eqref{eq:v-vector} for bidirectional/masked MHSA blocks, we use 
\begin{align}
q_{\sa;t;c}^{h}\equiv&\sum_{i=1}^{\hid} Q_{ci}^{h}s_{\sa;t;i}\, ,\label{eq:q-vector-diff}\\
k_{\sa;t;c}^{h}\equiv&\sum_{i=1}^{\hid} K_{ci}^{h}\widetilde{s}_{\sa;t;i}\, ,\label{eq:k-vector-diff}\\
v_{\sa;t;c}^{h}\equiv&\sum_{i=1}^{\hid} V_{ci}^{h}\widetilde{s}_{\sa;t;i}\, ,\label{eq:v-vector-diff}
\end{align}
where -- as before -- $s_{\sa;t;i}$ is the layer-normalized signal from the preceding (masked MHSA) block while -- \textit{not} as before -- $\widetilde{s}_{\sa;t;i}$ is the layer-normalized signal from right before it enters the decoder stack, i.e., right before the second positional-embedding parameters are added.\footnote{In theory, the scaling analysis stays intact and the inner products $(1/\hid)\sum_{i=1}^{\hid} s_{\sa_1;t_1;i}s_{\sa_2;t_2;i}$ just get replaced by $(1/\hid)\sum_{i=1}^{\hid} s_{\sa_1;t_1;i}\widetilde{s}_{\sa_2;t_2;i}$ in some places and $(1/\hid)\sum_{i=1}^{\hid} \widetilde{s}_{\sa_1;t_1;i}\widetilde{s}_{\sa_2;t_2;i}$ in other places.}
Then, similarly to the bidirectional MHSA, we use the softmax~\eqref{eq:bidirectional-MHSA} over all the tokens except that we mask out the \texttt{[pad]}ed tokens from considerations.
\ei

To make sure we are on the same footing, please check that those architectural choices result in $P=50265\cdot1024+514\cdot1024+1024^2\cdot(4+2\cdot 4)\cdot12+514\cdot1024+1024^2\cdot(4+4+2\cdot 4)\cdot12\approx 405\cdot 10^6$ model parameters for BART-large and $P=50265\cdot2048+514\cdot2048+2048^2\cdot(4+2\cdot 4)\cdot22+514\cdot2048+2048^2\cdot(4+4+2\cdot 4)\cdot22\approx 2.7\cdot 10^9$ model parameters for R2C2.

\subsubsection{Initialization Hyperparameters}\label{subsubsec:initialization_SLUDGE}
As for the initialization, for standard runs, following Refs.~\cite{lewis2019bart,R2C2seeker2022}, we initialize all the model paprameters uniformly across all the groups by using the mean-zero normal distribution with the standard deviation $\texttt{std}=0.02$.\footnote{To be precise, we zero-initialize the word-embedding parameters $\WE_{ij}$ when $j$ corresponds to the \texttt{[pad]} token.} For BART-large with $\hid=1024$ and $M=4$, this means in our language that $C_Q=C_K=C_V=C_U=C_W=0.4096$, $C_X=1.6384$, and $C_{\text{PE}}=C_{\text{WE}}=0.02$; for R2C2 with $\hid=2048$ and $M=4$, this means that $C_Q=C_K=C_V=C_U=C_W=0.8192$, $C_X=3.2768$, and $C_{\text{PE}}=C_{\text{WE}}=(0.02)^2$. These are mostly acceptable as order-one numbers -- however aesthetically displeasing they are --  \textit{except} that the covariance $C_{\text{WE}}=(0.02)^2=0.0004$ for word-embedding parameters is far from being of order one and hence, for the neural-tangent runs, we adjust it to $C_{\text{WE}}=1$ and, concomitantly, change $\rescale=1$ for the standard runs to $\rescale=\sqrt{1/\hid}$ for the neural-tangent runs: attend to {\color{darkgreen}green} color in Table~\ref{table:SLUDGE}.\footnote{In contrast, we take $C_{\text{PE}}=(0.02)^2$ as acceptable for the same reason as described in footnote~\ref{foot:002}:  the positional-embedding parameters act like bias parameters (i.e., not multiplicative but additive) and hence we could in principle set this hyperparameter to any order-one number including $C_{\text{PE}}=0$.}
%In contrast, the word-embedding parameters act like weight parameters and hence setting them to zero would be problematic.

\subsubsection{Optimizer and Training Hyperparameters}\label{subsubsec:train_SLUDGE}
Overall, our training recipe mostly follows the ones used in Refs.~\cite{lewis2019bart,R2C2seeker2022}.\footnote{The most significant difference is that, from compute considerations for our study, our warmup and \textit{target} training length of 5,000 iterations and 200,000 iterations are respectively shorter than 10,000 (15,000) iterations and 500,000 (500,000) iterations used in Ref.~\cite{lewis2019bart} (Ref.~\cite{R2C2seeker2022}).}

As for the optimizer, we use AdamW~\cite{kingma2014adam, loshchilov2017decoupled}~\eqref{eq:AdamW1}--\eqref{eq:AdamW4} with $(\beta_1,\beta_2,\epsilon)= (0.9, 0.998,10^{-6})$;
we also set \texttt{--dropout} and \texttt{--attention-dropout} to $0.1$ and \texttt{--clip-norm} to $0.1$.

As for the learning schedule, we use an effective batch size of $512$ with linear warmup~\cite{goyal2017accurate} for the first 5,000 iterations, followed by a linear learning-rate decay over the next 195,000 iterations, that is,
\begin{align}
\eta_t=\begin{cases}
\lrt\cdot\le(\frac{t}{5000}\ri)\, \ \ \ \ \ \ \ \ \ \ \ \ \ \ \ \ \ \text{for}\ \ \ t\leq 5000\, , \\
\lrt\cdot\le\{\frac{200000-t}{195000}\ri\}\, \ \ \ \text{for}\ \ \ t> 5000 \, .
\end{cases}
\end{align}
We train the models for the full 200,000 iterations for BART-large models while we stop the runs at 15,000 iterations for R2C2 models due to the compute consideration; we'll discuss the settings of the overall learning rate $\lrt$ and the weight decay $\wdt$ shortly  in \S\ref{subsubsec:versus_SLUDGE}.

As for the per-group learning-rate factors $\LambdaGA$'s, for the standard runs, we use the standard uniform scaling $\LambdaGA=1$, while for the neural-tangent runs, we follow our theoretical suggestions~\eqref{eq:AdamWlambdaG-preview-stem}--\eqref{eq:AdamWlambdaG-preview-head}, except, again, that we ignore the factor of $M=4$: attend to {\color{blue}blue} color in Table~\ref{table:SLUDGE}.
Overall, our neural-tangent scalings of learning rates essentially boil down to cranking up the learning rates for both word-embedding and positional-embedding  parameters by a factor of the width -- $\hid=1024$ for BART-large and $\hid=2048$ for R2C2 -- with respect to other model parameters.

\subsubsection{Comparison of Scaling Strategies}\label{subsubsec:versus_SLUDGE}
We here compare the performances of the models trained with the standard uniform scalings of hyperparameters against the ones trained with the neural-tangent scalings of hyperparameters: see Table~\ref{table:SLUDGE} below for a concise summary.
\begin{table}[h]
\scalebox{0.8}{
\begin{tabular}{|c|c|c|c|}
\hline
\multicolumn{4}{|c|}{AdamW--standard runs} \\\hline
    & initial \texttt{std} & rescale & relative \texttt{lr} factors \\\hline
    word embedding $\WE_{ij}$   & 0.02 & & 1  \\\hline
positional embedding  $\PE_{t;i}$ & 0.02 & & 1 \\\hline
$Q$--$K$--$V$--$U$--$W$--$X$ weights   & 0.02 & &  1 \\\hline
output $\mathcal{N}$ & &1.0  & \\\hline
\end{tabular}
 \begin{tabular}{|c|c|c|c|}
\hline
\multicolumn{4}{|c|}{AdamW--neural-tangent runs} \\\hline
    & initial \texttt{std} & rescale & relative \texttt{lr} factors \\\hline
    word embedding $\WE_{ij}$   &{\color{darkgreen}1} & & ${\color{blue}\hid^{-\frac{1}{2}}}$ \\\hline
positional embedding  $\PE_{t;i}$ &0.02 & & ${\color{blue}\hid^{-\frac{1}{2}}}$\\\hline
$Q$--$K$--$V$--$U$--$W$--$X$ weights   &0.02{\tiny (yuck)} & &  ${\color{blue}\hid^{-\frac{3}{2}}}$ \\\hline
output $\mathcal{N}$ & & ${\color{darkgreen}\hid^{-\frac{1}{2}}}$ & \\\hline
\end{tabular}
 }
\label{table:SLUDGE} 
\end{table}

\begin{figure}[h]
   \centerline{
   \includegraphics[width=0.5\linewidth]{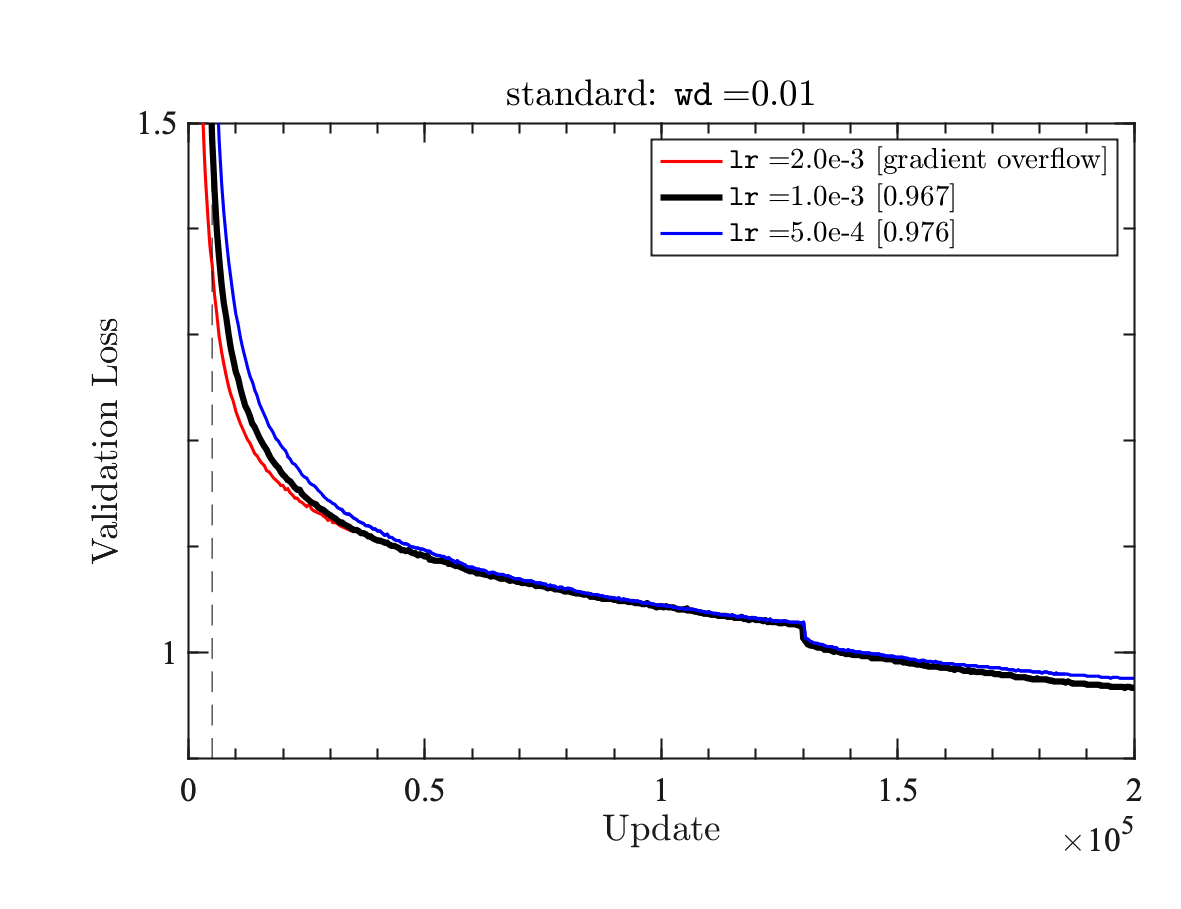}\!\!\!
   \includegraphics[width=0.5\linewidth]{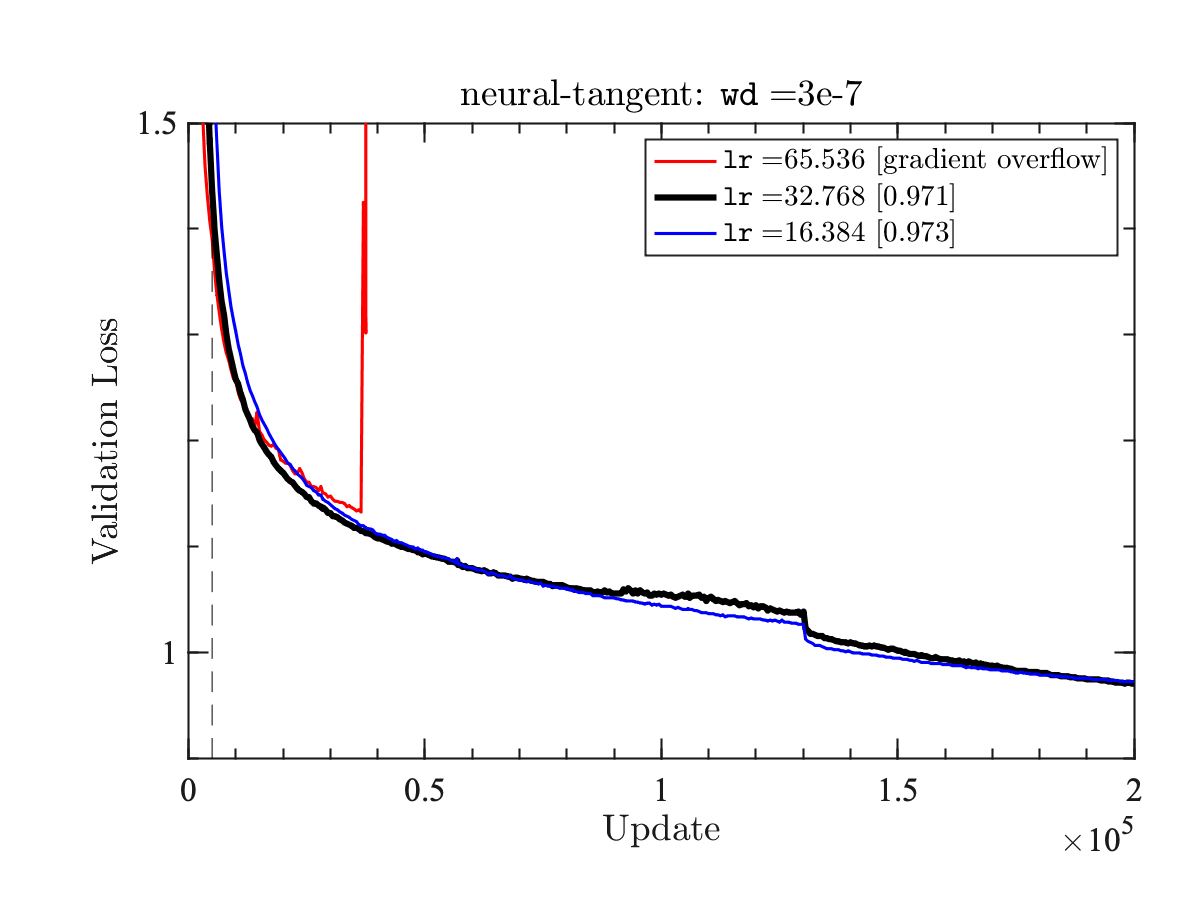}
   }
\caption{Hyperparameter searches for standard uniform (left) and neural-tangent (right) scaling strategies for BART-large trained by AdamW$_{(\beta_1,\beta_2,\epsilon)=(0.9,0.998,\text{1e-6})}$. For each, the validation loss is plotted as a function of training updates. In the legend, we record the global learning rate $\lrt$ [with the minimum validation loss along each trajectory]. Note that \textit{(i)} for both scaling strategies, the higher $\lrt$ runs (red) experienced gradient overflow,  \textit{(ii)} lowering $\lrt$ (going from black to blue) degrades the model performance more for the standard uniform scaling strategy than for the neural-tangent scaling strategy, and \textit{(iii)} for the neural-tangent scaling strategy, the $\lrt=32.768$ run -- despite its ``hiccup'' at around 90,000 iterations -- caught up with the $\lrt=16.384$ run in the end.}
\label{fig:hyper-BART}
\end{figure}

\begin{figure}[h]
   \centerline{
   \includegraphics[width=0.5\linewidth]{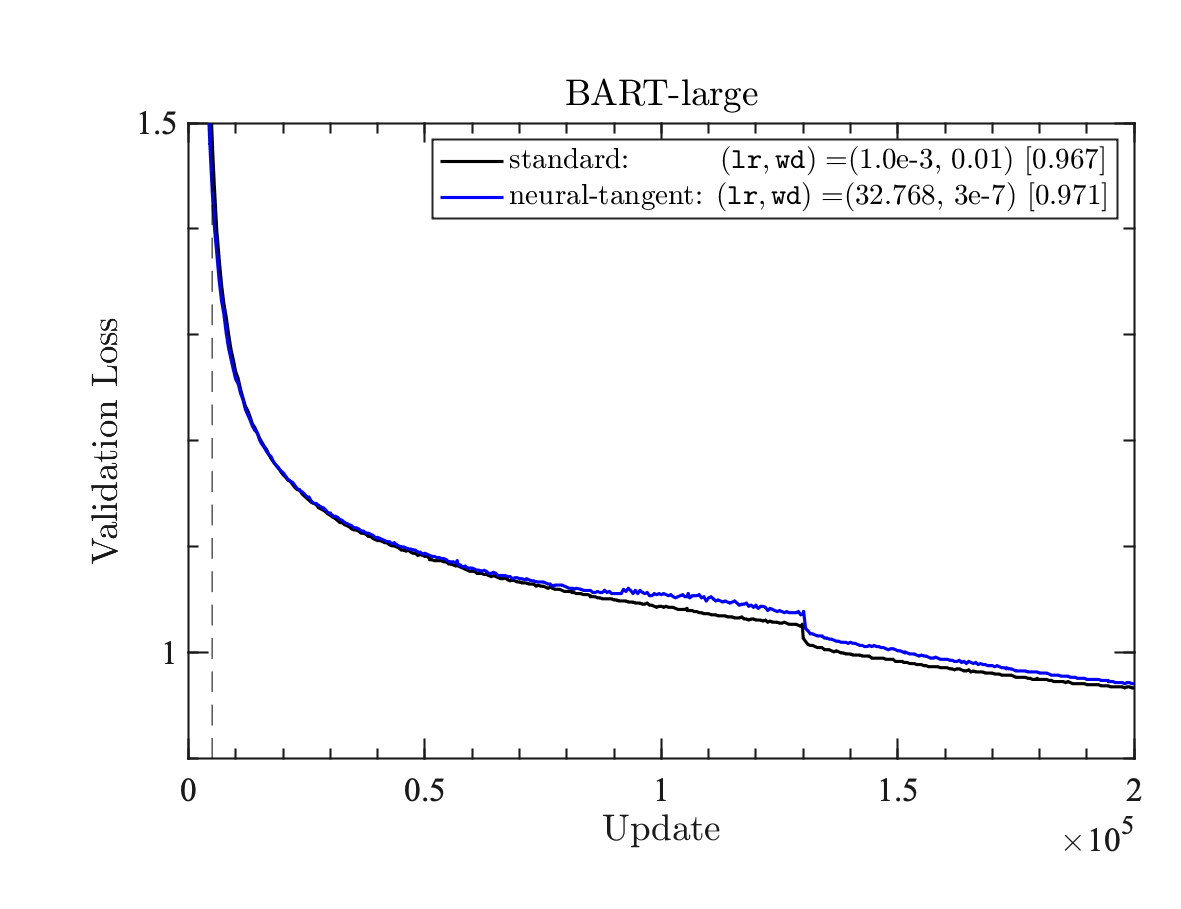}
   }
\caption{Comparison of the standard uniform (black) and neural-tangent (blue) scaling strategies for BART-large trained by AdamW$_{(\beta_1,\beta_2,\epsilon)=(0.9,0.998,\text{1e-6})}$. For each, the validation loss is plotted as a function of training updates. In the legend, we record the selected training hyperparameter pair $(\lrt,\wdt)$ for each scaling strategy [with the minimum validation loss along each trajectory].}
\label{fig:nonseeds-BART}
\end{figure}

In order to provide semi-rigorous comparisons -- within reason -- for BART-large, we tune the overall learning rate $\lrt$  but now that the models are five-fold larger than Vision Transformers treated in \S\ref{subsec:ViT}, we won't tune the weight decay $\wdt$. Specifically, for the standard runs, we fix the weight decay at $\wdt=0.01$ and search for the optimal overall learning rate $\lrt$ in $\log_{2}$ grid space; for the neural-tangent runs, given the remarks at the end of \S\ref{subsubsec:versus_ViT}, we shift the search space to $(\lrt,\wdt)\rightarrow (\hid^{3/2}\lrt,\hid^{-3/2}\wdt)$, now with fixed $\wdt=0.01\cdot 1024^{-3/2}\approx 3\text{e-7}$: see Fig.~\ref{fig:hyper-BART}. In fact, under this map, we see that the optimal $\lrt$ match between the standard uniform scaling strategy -- for which we find $\lrt^{\star}=1\text{e-3}$ -- and neural-tangent scaling strategy -- for which we find $\lrt^{\star}=1024^{3/2}\cdot1\text{e-3}=32.768$. We then compare the optimal runs: see Fig.~\ref{fig:nonseeds-BART}.

\begin{figure}[h]
   \centerline{
   \includegraphics[width=0.5\linewidth]{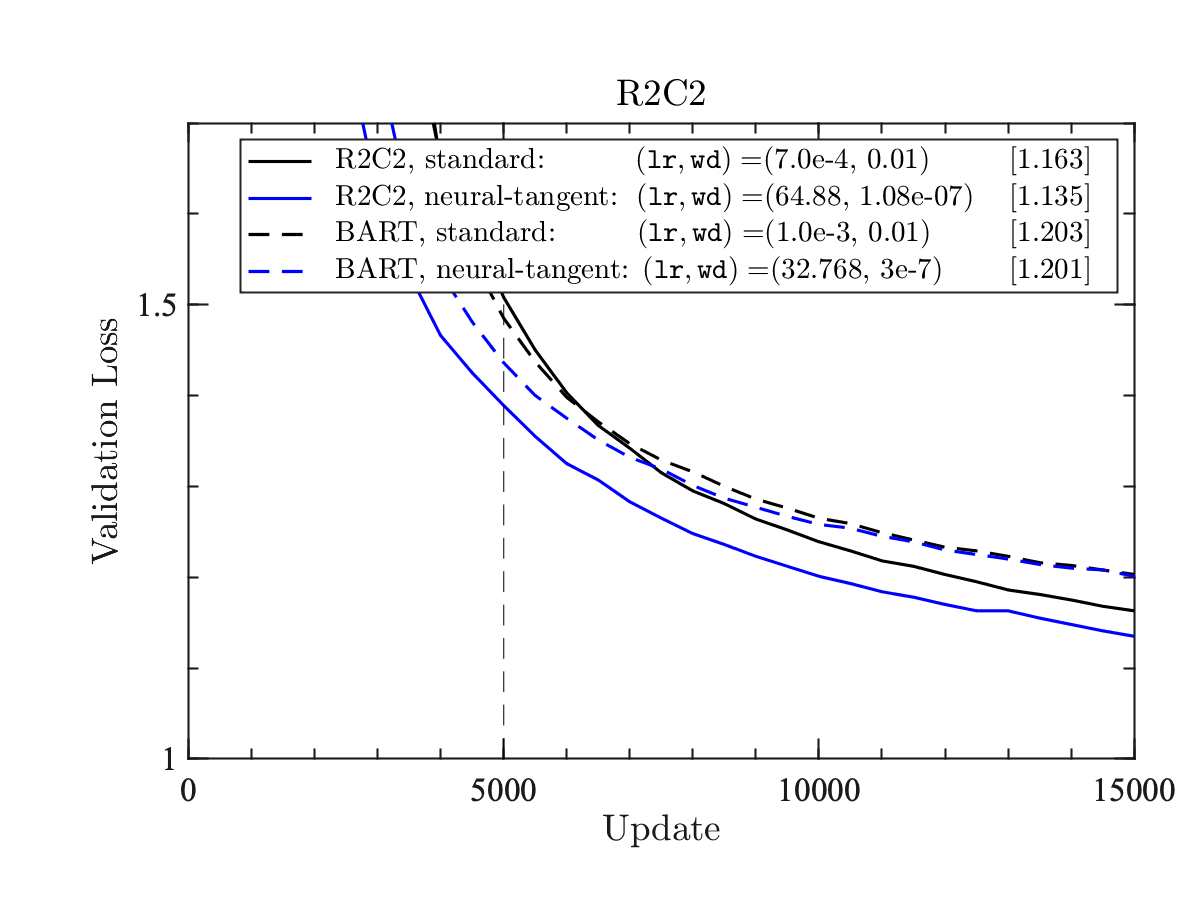}
   }
\caption{Comparison of the standard uniform (black) and neural-tangent (blue) scaling strategies for R2C2 trained by AdamW$_{(\beta_1,\beta_2,\epsilon)=(0.9,0.998,\text{1e-6})}$; for further comparison, we also include the similar runs for BART-large (dashed). For each, the validation loss is plotted as a function of training updates. In the legend, we record the zero-shotted training hyperparameter pair $(\lrt,\wdt)$ for each scaling strategy [with the minimum validation loss along each trajectory].}  
\label{fig:nonseeds-R2C2}
\end{figure}

For R2C2, since the models are seven-fold larger than BART-large -- and thirty-fold larger than Vision Transformers -- we won't even try to tune any of the hyperparameters. Instead, for the standard run, we follow the $(\lrt,\wdt)=(7\text{e-4},0.01)$ used in Ref.~\cite{R2C2seeker2022} -- albeit here with shorter warmup and target training lengths -- and, for the neural-tangent run, we shift it to $(\hid^{3/2}\lrt,\hid^{-3/2}\wdt)=(2048^{3/2}\cdot 7\text{e-4},2048^{-3/2}\cdot 0.01)\approx(64.88,1.08\text{e-7})$, motivated by our tuning results for both Vision Transformers and BART-large models: see Fig.~\ref{fig:nonseeds-R2C2}

For BART-large, the standard uniform scaling strategy seems to yield a marginally better model than the neural-tangent scaling strategy, though the latter strategy appears more robust against the change in the hyperparameter -- at least against lowering the global learning rate $\lrt$ -- and it would have been nice to run more experiments with distinct seeds.
On a more encouraging note, for R2C2, the neural-tangent scaling strategy seems to improve the convergence, at least at the early stage in training.
It would be interesting to further extend the experiments to the scale where Language Transformers start to suffer from mid-training spikes and see if the neural-tangent scaling strategy ameliorates them and if it results in further performance boost.

\newpage
\
\newpage
\newpage
\section*{Acknowledgements}
On personal notes, E.D.~is grateful to Stephen Roller and Kurt Shuster  for R2C2 pointers;
S.Y.~is grateful to Jaehoon Lee for various tips -- both theoretical and practical -- on scaling strategies, to Dan Roberts for writing tips, and to Ross Girshick, Mike Lewis, Eric Mintun, and Dan Roberts for unpublished collaboration on the effects of $C_Q=C_K=C_V=C_W=1$ and $C_U=C_X=0$;
and S.Z.~is grateful to FAIR for abundant compute resources and freedom to collaborate across the Labs.

On a technical note, to bring to the fore the future direction buried in footnote~\ref{foot:nvocab-stem}, we collectively acknowledge that -- given our time and energy constraints -- we didn't properly take into account the interplay between the batch size $\vert\mathcal{A}_t\vert$, sequence length $T$, vocabulary size $\nvocab$, and token distribution, but nonetheless hope that our work will be a helpful anecdatapoint for future work.
\newpage
\appendix

%\section{Supersymmetry}\label{app:self-attention}
%\section{More Attention on Statistics of Self-Attention}\label{app:self-attention}
\section{Attention on Statistics of Self-Attention}\label{app:self-attention}

\setlength{\epigraphwidth}{0.42\textwidth}
\epigraph{\textit{To know others you must know yourself first.}}{Bumblebee}%{Goldbug}

In this Appendix, we'll determine the statistics of the query--key dot product~\eqref{eq:matrix-product-1},
\be
\wst_{\sa;tt'}^{h}\equiv\frac{1}{\sqrt{C}}\sum_{c=1}^{C}q_{\sa;t;c}^{h}k_{\sa;t';c}^{h}=\frac{1}{\sqrt{C}}\sum_{c=1}^{C}\sum_{i_1,i_2=1}^{\hid}Q_{ci_1}^{h}K_{ci_2}^{h}s_{\sa;t;i_1}s_{\sa;t';i_2}\, ,
\ee
which in turn dictates the statistics of the self-attention matrices. In the main text, we've seen that the mean of this product~\eqref{eq:first-dot} vanishes as
\begin{align}\label{eq:first-dot-review}
\E{\wst_{\sa;tt'}^{h}}=&\E{\frac{1}{\sqrt{C}}\sum_{c=1}^{C}\sum_{i,j=1}^{\hid}Q_{ci}^{h}K_{cj}^{h}s_{\sa;t;i}s_{\sa;t';j}}\, \\
=&\frac{1}{\sqrt{C}}\sum_{c=1}^{C}\sum_{i,j=1}^{\hid}\E{Q_{ci}^{h}}\E{K_{cj}^{h}}\E{s_{\sa;t;i}s_{\sa;t';j}}=0\, ,\nonumber
\end{align}
while its covariance~\eqref{eq:second-dot} is given by
\begin{align}\label{eq:second-dot-review}
&\E{\wst_{\sa_1;t_1t'_1}^{h_1}\wst_{\sa_2;t_2t'_2}^{h_2}}\, \\
=&\frac{1}{C}\sum_{c_1,c_2=1}^{C}\sum_{i_1,i_2,j_1,j_2=1}^{\hid}\E{Q_{c_1i_1}^{h_1}K_{c_1j_1}^{h_1}s_{\sa_1;t_1;i_1}s_{\sa_1;t'_1;j_1}Q_{c_2i_2}^{h_2}K_{c_2j_2}^{h_2}s_{\sa_2;t_2;i_2}s_{\sa_2;t'_2;j_2}}\, \nonumber \\
=&\frac{1}{C}\sum_{c_1,c_2=1}^{C}\sum_{i_1,i_2,j_1,j_2=1}^{\hid}\frac{C_Q}{\hid}\frac{C_K}{\hid}\delta_{c_1c_2}\delta_{i_1i_2}\delta_{j_1j_2}\delta^{h_1h_2}\E{s_{\sa_1;t_1;i_1}s_{\sa_1;t'_1;j_1}s_{\sa_2;t_2;i_2}s_{\sa_2;t'_2;j_2}}\, \nonumber\\
=&\delta^{h_1h_2}C_Q C_K\E{\le(\frac{1}{\hid}\sum_{i=1}^{\hid}s_{\sa_1;t_1;i}s_{\sa_2;t_2;i}\ri)\le(\frac{1}{\hid}\sum_{j=1}^{\hid}s_{\sa_1;t'_1;j}s_{\sa_2;t'_2;j}\ri)}\, \nonumber\\
=&\delta^{h_1h_2}C_Q C_K \kernelLN_{(\sa_1;t_1)(\sa_2;t_2)}\kernelLN_{(\sa_1;t'_1)(\sa_2;t'_2)}+O\le(\frac{1}{n}\ri)\, \nonumber\\
\equiv&\delta^{h_1h_2}\kernelA_{(\sa_1;t_1t'_1)(\sa_2;t_2t'_2)}\, ,
\end{align}
where in the last line we've introduced the kernel $\kernelA_{(\sa_1;t_1t'_1)(\sa_2;t_2t'_2)}$ for the query--key dot products.

Assuming that the query and key weights are distributed symmetrically around zero -- as is usually the case -- we can clearly see that all the odd-point correlators vanish as
\be
\E{\wst_{\sa_1;t_1t'_1}^{h_1}\cdots \wst_{\sa_{2m-1};t_{2m-1}t'_{2m-1}}^{h_{2m-1}}}=0\, .
\ee
Thus we'll focus on the even-point correlators. Going ahead and rolling them out just as we did for the covariance (a step-by-step instruction will follow),  we get
\begin{align}\label{eq:many-times-dot}
&\E{\wst_{\sa_1;t_1t'_1}^{h_1}\cdots \wst_{\sa_{2m};t_{2m}t'_{2m}}^{h_{2m}}}\, \\
=&\frac{1}{C^m}\sum_{c_1,\ldots,c_{2m}=1}^{C}\sum_{i_1,\ldots,i_{2m}=1}^{\hid}\sum_{j_1,\ldots,j_{2m}=1}^{\hid}\E{Q_{c_1i_1}^{h_1}\cdots Q_{c_{2m}i_{2m}}^{h_{2m}}}\E{K_{c_1j_1}^{h_1}\cdots K_{c_{2m}j_{2m}}^{h_{2m}}}\, \nonumber\\
&\ \ \ \ \ \ \ \ \ \times\E{\Big(s_{\sa_1;t_1;i_1}\cdots s_{\sa_{2m};t_{2m};i_{2m}}\Big)\Big(s_{\sa_1;t'_1;j_1}\cdots s_{\sa_{2m};t'_{2m};j_{2m}}\Big)}\, \nonumber \\
=&\frac{1}{C^m}\sum_{c_1,\ldots,c_{2m}=1}^{C}\sum_{i_1,\ldots,i_{2m}=1}^{\hid}\sum_{j_1,\ldots,j_{2m}=1}^{\hid}\le(\frac{C_Q}{\hid}\frac{C_K}{\hid}\ri)^m\, \nonumber\\
&\ \ \ \ \ \ \ \ \ \times\le[\sum_{\text{all pairing}}\le(\delta_{c_{p_1}c_{p_2}}\cdots \delta_{c_{p_{2m-1}}c_{p_{2m}}}\ri)\le(\delta_{i_{p_1}i_{p_2}}\cdots \delta_{i_{p_{2m-1}}i_{p_{2m}}}\ri)\le(\delta^{h_{p_1}h_{p_2}}\cdots \delta^{h_{p_{2m-1}}h_{p_{2m}}}\ri)\ri]\, \nonumber\\
&\ \ \ \ \ \ \ \ \ \times\le[\sum_{\text{all pairing}}\le(\delta_{c_{q_1}c_{q_2}}\cdots \delta_{c_{q_{2m-1}}c_{q_{2m}}}\ri)\le(\delta_{j_{q_1}j_{q_2}}\cdots \delta_{j_{q_{2m-1}}j_{q_{2m}}}\ri)\le(\delta^{h_{q_1}h_{q_2}}\cdots \delta^{h_{q_{2m-1}}h_{q_{2m}}}\ri)\ri]\, \nonumber\\
&\ \ \ \ \ \ \ \ \ \times\E{\Big(s_{\sa_1;t_1;i_1}\cdots s_{\sa_{2m};t_{2m};i_{2m}}\Big)\Big(s_{\sa_1;t'_1;j_1}\cdots s_{\sa_{2m};t'_{2m};j_{2m}}\Big)}+O\le(\frac{1}{\hid}\ri)\, \nonumber \\
=&\frac{1}{C^m}C^m\sum_{i_1,\ldots,i_{2m}=1}^{\hid}\sum_{j_1,\ldots,j_{2m}=1}^{\hid}\le(\frac{C_Q}{\hid}\frac{C_K}{\hid}\ri)^m\, \nonumber\\
&\ \ \ \ \ \ \ \ \ \times\le[\sum_{\text{all pairing}}\le(\delta_{i_{p_1}i_{p_2}}\cdots \delta_{i_{p_{2m-1}}i_{p_{2m}}}\ri)\le(\delta^{h_{p_1}h_{p_2}}\cdots \delta^{h_{p_{2m-1}}h_{p_{2m}}}\ri)\le(\delta_{j_{p_1}j_{p_2}}\cdots \delta_{j_{p_{2m-1}}j_{p_{2m}}}\ri)\ri]\, \nonumber\\
&\ \ \ \ \ \ \ \ \ \times\E{\Big(s_{\sa_1;t_1;i_1}\cdots s_{\sa_{2m};t_{2m};i_{2m}}\Big)\Big(s_{\sa_1;t'_1;j_1}\cdots s_{\sa_{2m};t'_{2m};j_{2m}}\Big)}+O\le(\frac{1}{C}\ri)\, \nonumber \\
=&\le(C_Q C_K\ri)^m \sum_{\text{all pairing}}\le(\delta^{h_{p_1}h_{p_2}}\cdots \delta^{h_{p_{2m-1}}h_{p_{2m}}}\ri)\, \nonumber\\
&\ \ \ \ \ \ \ \ \ \times\mathbb{E}\Bigg[\le(\frac{1}{\hid}\sum_{i_1=1}^{\hid}s_{\sa_{p_1};t_{p_1};i_1}s_{\sa_{p_2};t_{p_2};i_1}\ri)\cdots \le(\frac{1}{\hid}\sum_{i_m=1}^{\hid}s_{\sa_{p_{2m-1}};t_{p_{2m-1}};i_{m}}s_{\sa_{p_{2m}};t_{p_{2m}};i_{m}}\ri)\, \nonumber\\
&\ \ \ \ \ \ \ \ \ \ \ \ \ \ \ \ \times \le(\frac{1}{\hid}\sum_{j_1=1}^{\hid}s_{\sa_{p_1};t'_{p_1};j_1}s_{\sa_{p_2};t'_{p_2};j_1}\ri)\cdots \le(\frac{1}{\hid}\sum_{j_m=1}^{\hid}s_{\sa_{p_{2m-1}};t'_{p_{2m-1}};j_{m}}s_{\sa_{p_{2m}};t'_{p_{2m}};j_{m}}\ri)\Bigg]+O\le(\frac{1}{C}\ri)\, \nonumber \\
=&\sum_{\text{all pairing}}\le(\delta^{h_{p_1}h_{p_2}}\cdots \delta^{h_{p_{2m-1}}h_{p_{2m}}}\ri)  \, \nonumber\\
&\ \ \ \ \ \ \ \ \ \times\kernelA_{(\sa_{p_{1}};t_{p_{1}}t'_{p_{1}})(\sa_{p_{2}};t_{p_{2}}t'_{p_{2}})}\cdots \kernelA_{(\sa_{p_{2m-1}};t_{p_{2m-1}}t'_{p_{2m-1}})(\sa_{p_{2m}};t_{p_{2m}}t'_{p_{2m}})}+O\le(\frac{1}{C}\ri)\, .\nonumber
\end{align}
Here, in the first equality, we explicitly wrote out the query--key dot products and separated the expectations for the query and key weights using their statistical independence; in the second equality, we used the Wick's theorem and expressed the expectation of query weights as the sum over all the $(2m-1)!!$ pairings of the auxiliary indices $1,\ldots, 2m$, and did the same for the key weights (this step is exact if the query and key weights are drawn from normal distributions while it involves the $1/\hid$ corrections for non-normal distributions); in the third equality we summed over the per-head-channel indices, noticing that the contributions would be $(1/C=H/n)$-suppressed when the pairings of those indices differ between that for the query weights and that for the key weights (if this explanation is cryptic, then we recommend performing this step explicitly for the case of $2m=4$); in the fourth step, we simply performed the summation over embedding indices; in the fifth step, we used our eightfold result~\eqref{eq:eightfold} to truncate away the $1/\hid$ corrections and used the definition of  the newly-introduced kernel $\kernelA_{(\sa_1;t_1t'_1)(\sa_2;t_2t'_2)}$.

All in all, at the leading order in $1/C=H/n$, the query--key dot product $\wst_{\sa;tt'}^{h}$ obeys Gaussian statistics with zero mean and order-one covariance $\delta^{h_1h_2}\kernelA_{(\sa_1;t_1t'_1)(\sa_2;t_2t'_2)}$. As such, any order-one function of the query--key dot product $\wst_{\sa;tt'}^{h}$, in particular the self-attention matrix $\ws_{\sa;tt'}^{h}$, has the expectation value of order one (unless the said function involves \textit{only} odd powers of the query--key dot products). We also emphasize that all the distinct heads are statistically independent -- as manifested by the Kronecker delta  $\delta^{h_1h_2}$ -- and, e.g.,  the expectation value $\E{\ws^{h}_{\sa_1;t_1t'_1}\ws^{h}_{\sa_2;t_2t'_2}}$ that appeared in footnote~\ref{foot:MHSA-stats-101} takes the same order-one value for all the heads $h$ and can be expressed as  a $\le(\vert\mathcal{D}\vert T^2\ri)$-dimensional Gaussian integral with the kernel $\kernelA_{(\sa_1;t_1t'_1)(\sa_2;t_2t'_2)}$.

Speaking of footnote~\ref{foot:MHSA-stats-101}, to show the factorization therein, we just have to note that the fivefold derivation~\eqref{eq:many-times-dot} of the Gaussian statistics above goes through verbatim to yield
\begin{align}
&\E{\wst_{\sa_1;t_1t'_1}^{h_1}\cdots \wst_{\sa_{2m};t_{2m}t'_{2m}}^{h_{2m}}\mathcal{F}(s)}\, \\
=&\le(C_Q C_K\ri)^m \sum_{\text{all pairing}}\le(\delta^{h_{p_1}h_{p_2}}\cdots \delta^{h_{p_{2m-1}}h_{p_{2m}}}\ri)\, \nonumber\\
&\ \ \ \times\mathbb{E}\Bigg[\le(\frac{1}{\hid}\sum_{i_1=1}^{\hid}s_{\sa_{p_1};t_{p_1};i_1}s_{\sa_{p_2};t_{p_2};i_1}\ri)\cdots \le(\frac{1}{\hid}\sum_{i_m=1}^{\hid}s_{\sa_{p_{2m-1}};t_{p_{2m-1}};i_{m}}s_{\sa_{p_{2m}};t_{p_{2m}};i_{m}}\ri)\, \nonumber\\
&\ \ \ \ \ \ \ \times \le(\frac{1}{\hid}\sum_{j_1=1}^{\hid}s_{\sa_{p_1};t'_{p_1};j_1}s_{\sa_{p_2};t'_{p_2};j_1}\ri)\!\cdots\! \le(\frac{1}{\hid}\sum_{j_m=1}^{\hid}s_{\sa_{p_{2m-1}};t'_{p_{2m-1}};j_{m}}s_{\sa_{p_{2m}};t'_{p_{2m}};j_{m}}\ri)\!\!\mathcal{F}(s)\Bigg]\!+O\le(\frac{1}{C}\ri)\, \nonumber
\end{align}
with an insertion of any function $\mathcal{F}(s)$ of the signals $s_{\sa;t;i}$,  and we can then use the eightfold result~\eqref{eq:eightfold}  once again to seal the deal~\eqref{eq:MHSA-stats-101}.

If you've been paying attention, then there is one more deal~\eqref{eq:MHSA-stats-102} to seal, that is, footnote~\ref{foot:MHSA-stats-102}. That sealing essentially boils down to realizing that
\begin{align}\label{eq:seal}
&\frac{1}{C^m}\sum_{c_1,\ldots,c_{2m}=1}^{C}\E{Q_{c_1i_1}^{h_1}\cdots Q_{c_{2m}i_{2m}}^{h_{2m}}}\E{K_{c_1j_1}^{h_1}\cdots K_{c_{2m}j_{2m}}^{h_{2m}}\le(K_{c_{2m+1} j_{2m+1}}^{h_{2m+1}}K_{c_{2m+2} j_{2m+2}}^{h_{2m+2}}\ri)}\, \\
=&\frac{1}{C^m}\sum_{c_1,\ldots,c_{2m}=1}^{C}\le(\frac{C_Q}{\hid}\frac{C_K}{\hid}\ri)^m\, \nonumber\\
&\ \ \ \times\le[\sum_{\text{all pairing}}\le(\delta_{c_{p_1}c_{p_2}}\cdots \delta_{c_{p_{2m-1}}c_{p_{2m}}}\ri)\le(\delta_{i_{p_1}i_{p_2}}\cdots \delta_{i_{p_{2m-1}}i_{p_{2m}}}\ri)\le(\delta^{h_{p_1}h_{p_2}}\cdots \delta^{h_{p_{2m-1}}h_{p_{2m}}}\ri)\ri]\, \nonumber\\
&\ \ \ \times\le[\sum_{\text{all pairing}}\le(\delta_{c_{q_1}c_{q_2}}\cdots \delta_{c_{q_{2m+1}}c_{q_{2m+2}}}\ri)\le(\delta_{j_{q_1}j_{q_2}}\cdots \delta_{j_{q_{2m+1}}j_{q_{2m+2}}}\ri)\le(\delta^{h_{q_1}h_{q_2}}\cdots \delta^{h_{q_{2m+1}}h_{q_{2m+2}}}\ri)\ri]\, \nonumber\\
=&\le(\frac{C_Q}{\hid}\frac{C_K}{\hid}\ri)^m\delta_{c_{2m+1}c_{2m+2}}\delta_{j_{2m+1}j_{2m+2}}\delta^{h_{2m+1}h_{2m+2}}\, \nonumber\\
&\times\le[\sum_{\text{all pairing}}\le(\delta_{i_{p_1}i_{p_2}}\cdots \delta_{i_{p_{2m-1}}i_{p_{2m}}}\ri)\!\le(\delta^{h_{p_1}h_{p_2}}\cdots \delta^{h_{p_{2m-1}}h_{p_{2m}}}\ri)\!\le(\delta_{j_{p_1}j_{p_2}}\cdots \delta_{j_{p_{2m-1}}j_{p_{2m}}}\ri)\!+O\le(\frac{1}{C}\ri)\ri]\, .\nonumber
\end{align}
For a neater derivation that generalizes better, search for interlayer correlations in Ref.~\cite{PDLT}.

\bibliographystyle{utphys}
\bibliography{ETT}{}
\end{document}